%% file: main.tex
\documentclass[10pt,twocolumn,letterpaper]{article}

\usepackage{cvpr}              

\usepackage{multirow}
\usepackage{float}
\usepackage{ifthen}
\usepackage{verbatim}
\usepackage{subcaption}

\input{preamble}

\definecolor{cvprblue}{rgb}{0.21,0.49,0.74}
\usepackage[pagebackref,breaklinks,colorlinks,citecolor=cvprblue]{hyperref}
\usepackage[accsupp]{axessibility}

\def\paperID{138}
\def\confName{3DV\xspace}
\def\confYear{2024\xspace}

\newcommand{\netname}{RAMDepth}

\newcommand{\mvsnet}{Yao et al. \cite{yao2018mvsnet}}
\newcommand{\ucsnet}{Cheng et al. \cite{cheng2020deep}}
\newcommand{\patchnet}{Wang et al. \cite{wang2021patchmatchnet}}
\newcommand{\casnet}{Gu et al. \cite{gu2020cascade}}
\newcommand{\smplrcn}{Sayed et al. \cite{sayed2022simplerecon}}
\newcommand{\vismvsnet}{Zhang et al. \cite{zhang2020vismvsnet}}
\newcommand{\dhcnet}{Yao et al. \cite{yao2019recurrent}}
\newcommand{\cermvs}{Ma et al. \cite{ma2022cermvs}}
\newcommand{\raftstereo}{Lipson et al $\dagger$ \cite{lipson2021raftstereo}}

\definecolor{somegray}{rgb}{0.5, 0.5, 0.5}
\newcommand{\darkgrayed}[1]{\textcolor{somegray}{#1}}
\makeatletter
\newcommand*\titleheader[1]{\gdef\@titleheader{#1}}
\AtBeginDocument{%
  \let\st@red@title\@title
  \def\@title{%
    \vskip-3em
    \bgroup\normalfont\large\centering\@titleheader\par\egroup
    \vskip1.5em\st@red@title}
}
\makeatother

\titleheader{\darkgrayed{This paper has been accepted for publication at the \\
IEEE International Conference on 3D Vision (3DV), Davos, 2024. \copyright IEEE}}

\title{Range-Agnostic Multi-View Depth Estimation with Keyframe Selection}

\author{Andrea Conti$^\dagger$ \quad\quad\quad\quad Matteo Poggi$^{\dagger,\ddagger}$ \quad\quad\quad\quad Valerio Cambareri$^*$ \quad\quad\quad\quad Stefano Mattoccia$^{\dagger,\ddagger}$ \\%
\small \quad\quad\quad $^\dagger$Department of Computer Science and Engineering \quad\quad\quad $^*$Sony Depthsensing Solutions \quad\quad\quad \\
\small $^\ddagger$Advanced Research Center on Electronic System (ARCES) \quad\quad\quad Brussels, Belgium \quad\quad\quad\quad\\
\small \quad\quad University of Bologna, Italy \quad\quad\quad \textcolor{white}{Brussels, Belgium} \quad\quad\quad\quad \\
}

\newcommand{\ignore}[1]{}

\begin{document}

\twocolumn[{
\maketitle
\vspace{-1.2cm}
\begin{center}
    \renewcommand{\tabcolsep}{2pt}
    \begin{tabular}{cc}
    \multicolumn{2}{c}{\scriptsize \textbf{Source Views}} \\
    \includegraphics[width=0.115\textwidth]{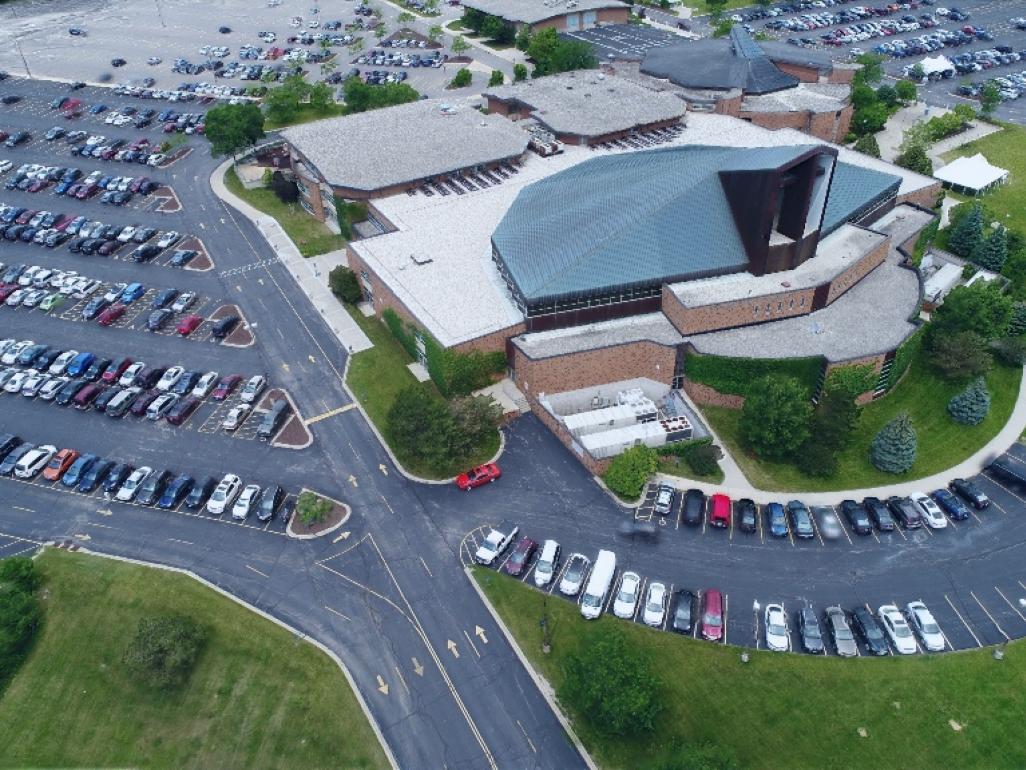} &
    \includegraphics[width=0.115\textwidth]{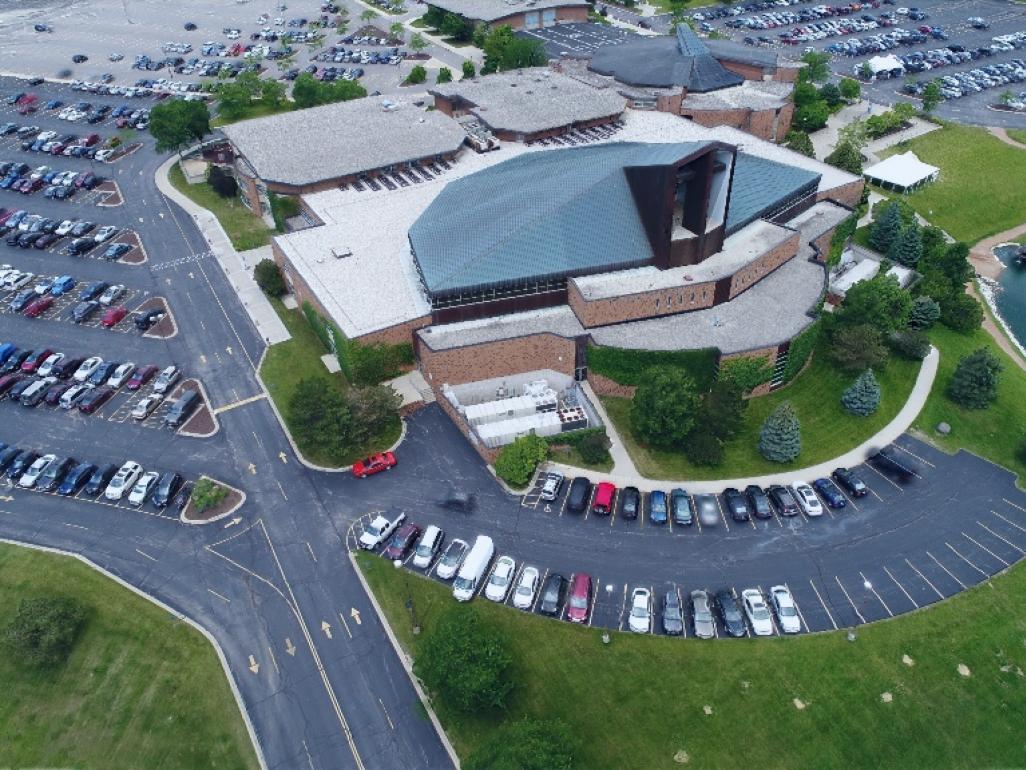} \\
    \includegraphics[width=0.115\textwidth]{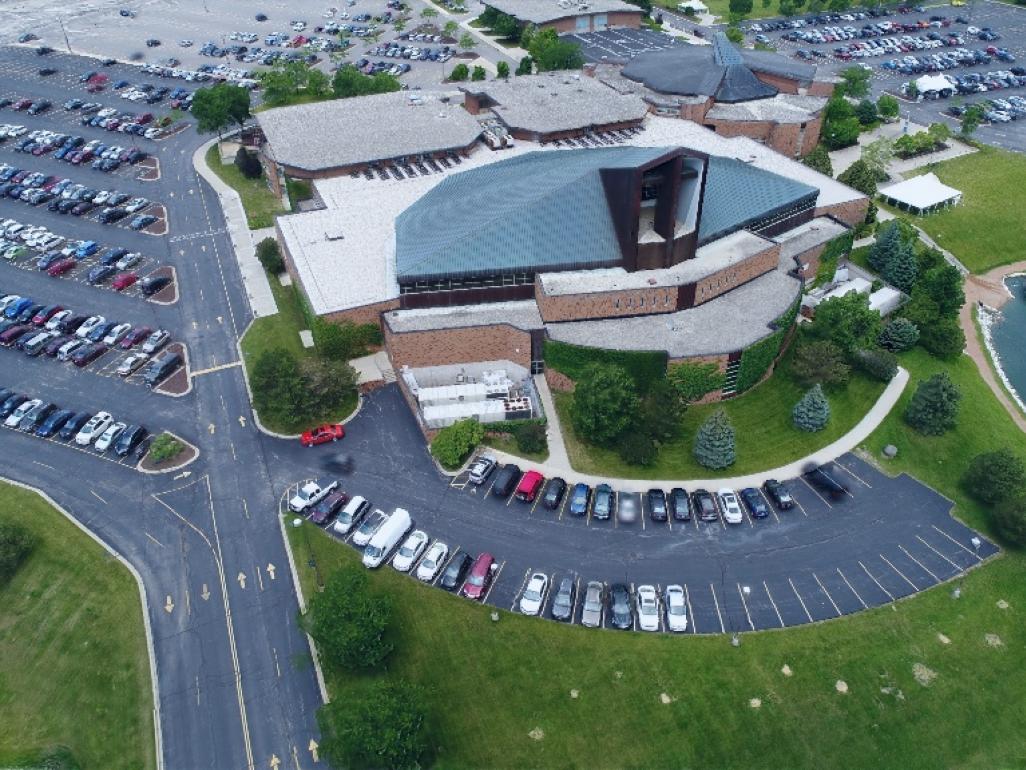} &
    \includegraphics[width=0.115\textwidth]{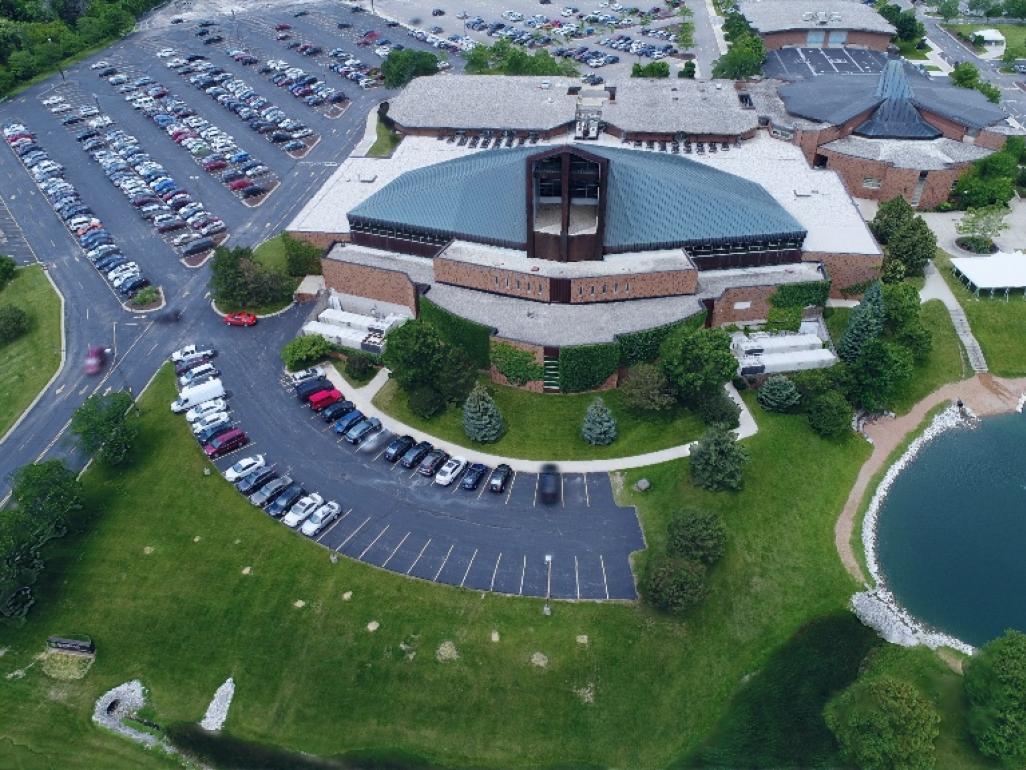} \\
    \end{tabular}
    \begin{tabular}{ccc}
    \multicolumn{1}{c}{\scriptsize \textbf{Reference View}} &
    \multicolumn{1}{c}{\scriptsize \textbf{Prediction}} &
    \multicolumn{1}{c}{\scriptsize \textbf{Ground Truth}} \\
    \includegraphics[width=0.24\textwidth]{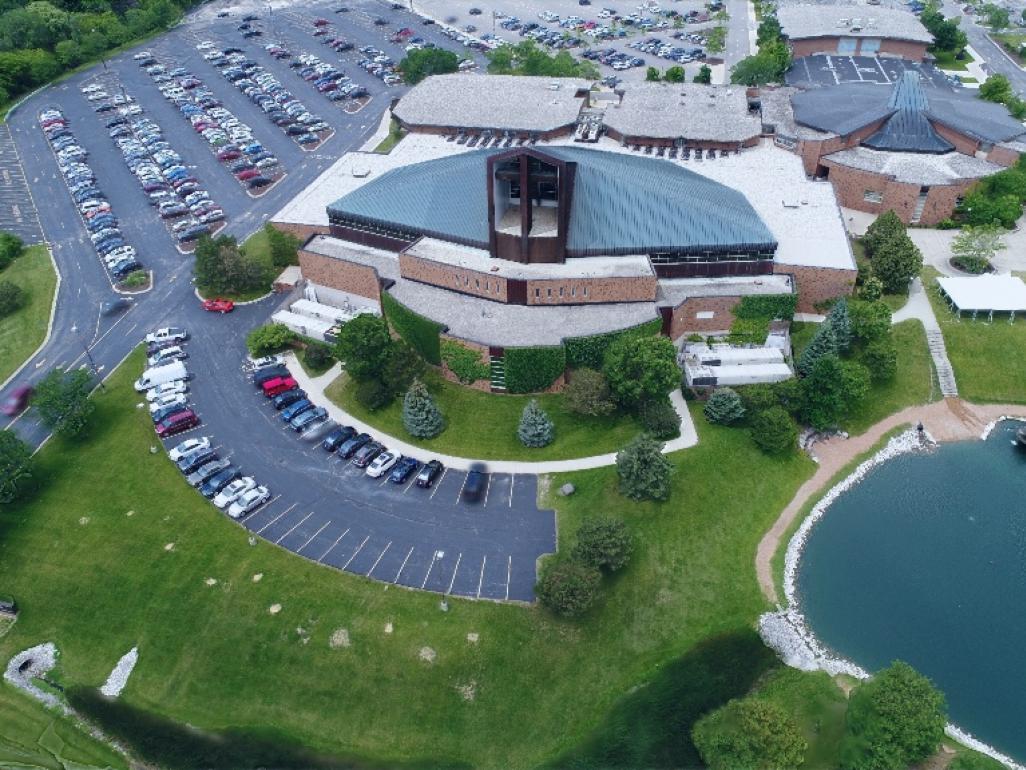} &
    \includegraphics[width=0.24\textwidth]{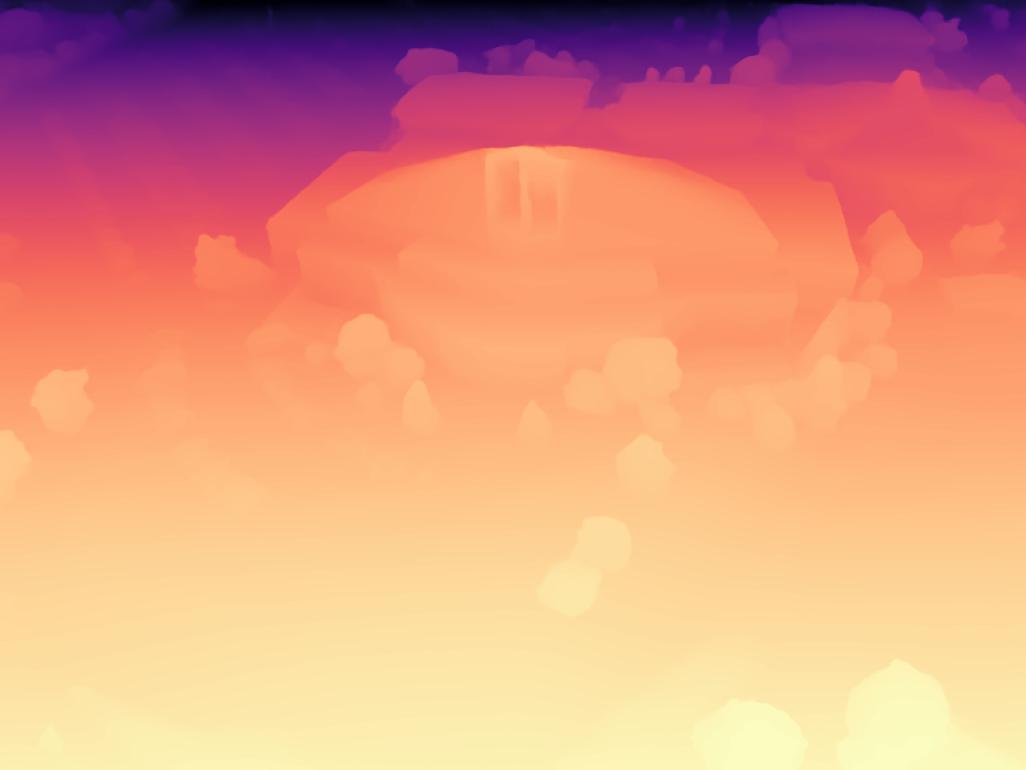} &
    \includegraphics[width=0.24\textwidth]{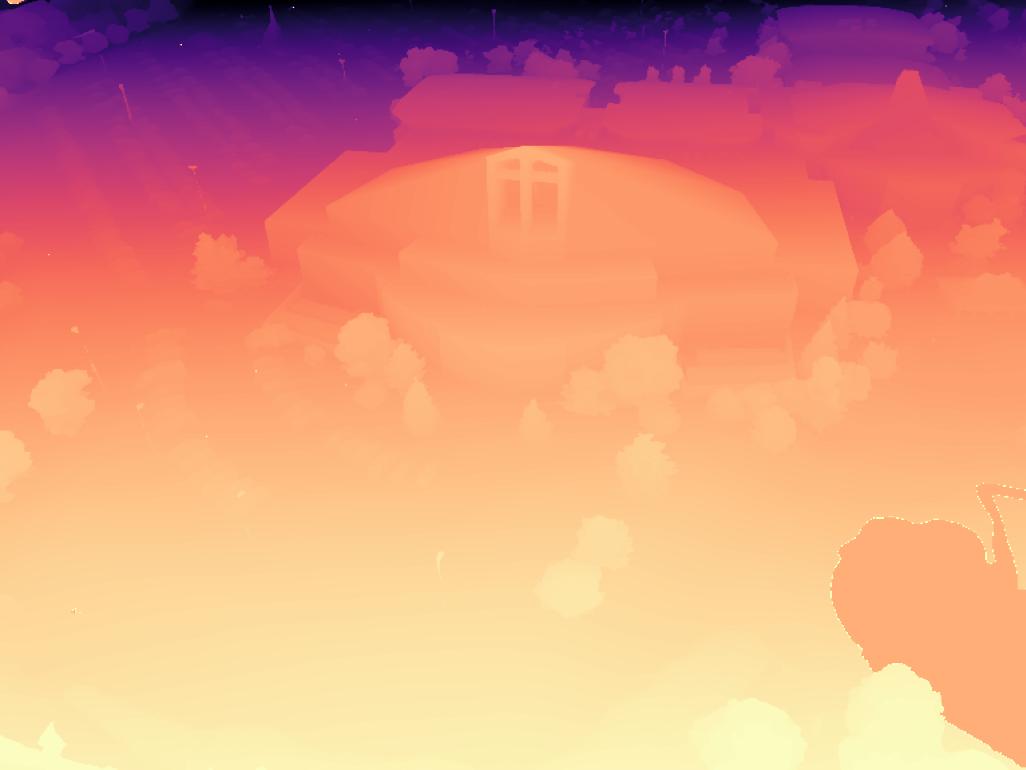} \\
    \end{tabular}
    \begin{tabular}{cc}
    \includegraphics[trim=0cm 4cm 0cm 0cm, clip, width=0.49\textwidth]{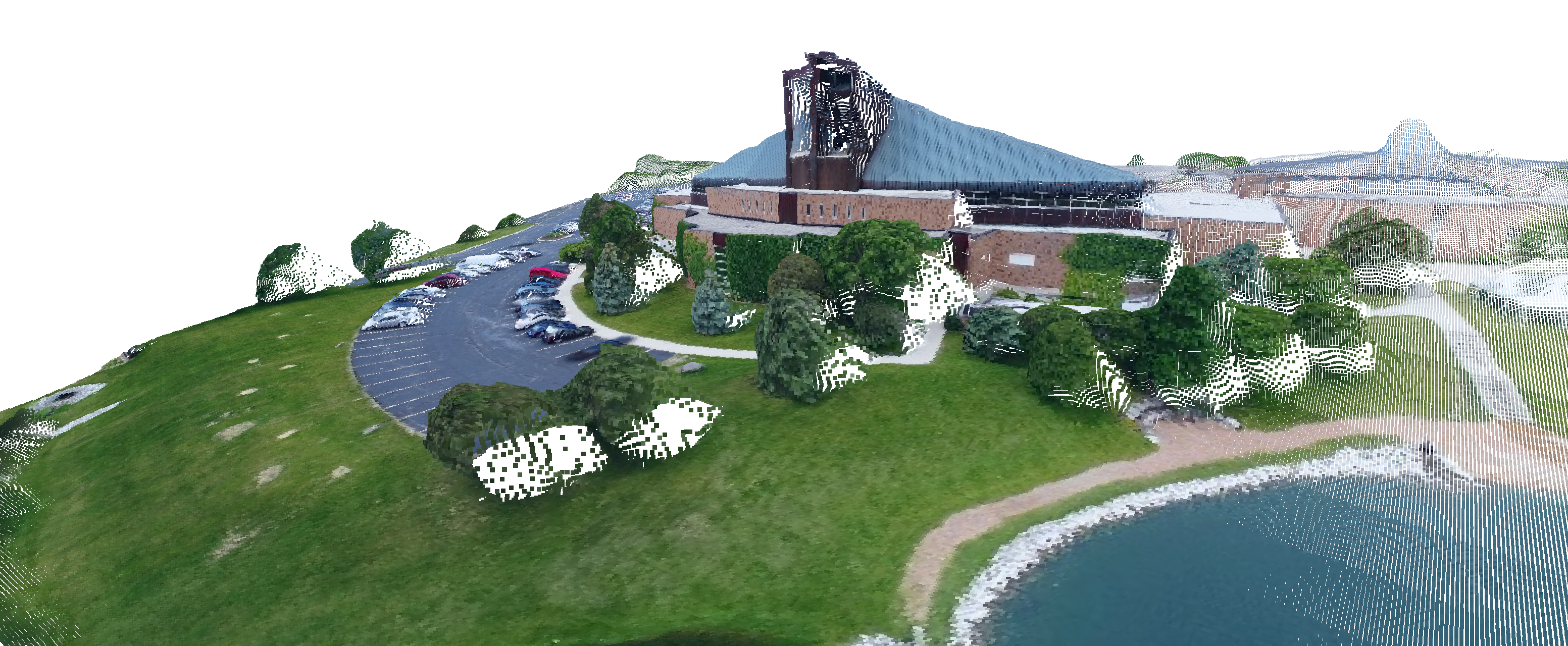} & 
    \includegraphics[trim=0cm 4cm 0cm 0cm, clip, width=0.49\textwidth]{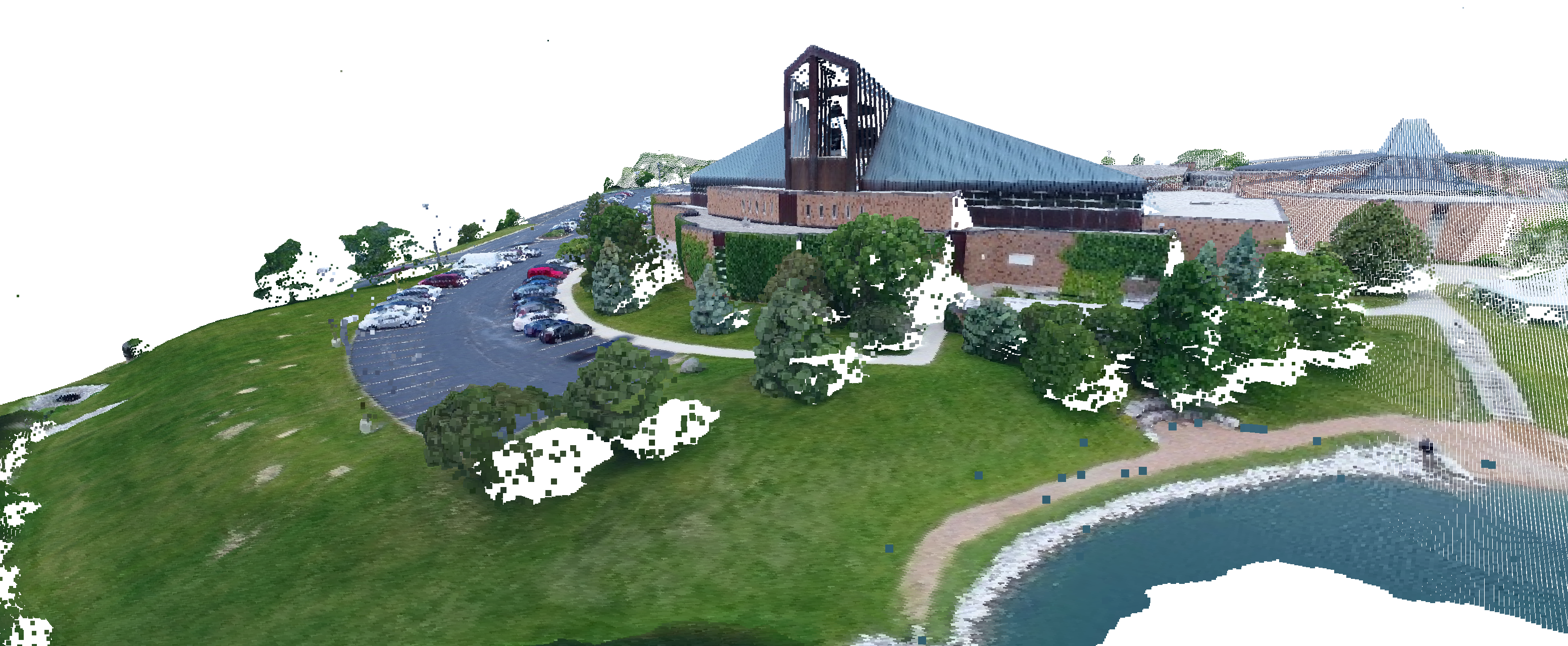} \\
    \multicolumn{1}{c}{\scriptsize \textbf{Prediction}} &
    \multicolumn{1}{c}{\scriptsize \textbf{Ground Truth}} \\
    \end{tabular}\vspace{-0.3cm}
    \captionof{figure}{
    \textbf{Depth Estimation and 3D reconstruction with \netname{} on Blended \cite{yao2020blendedmvs}.} On top: given five images of the same scene, our framework can estimate accurate depth maps through multi-view geometry without requiring any knowledge about the reference view depth range. At the bottom: the point cloud obtained from the prediction of the network and the respective ground-truth.
    }
    \label{fig:teaser}
\end{center}
\vspace{0.1cm}
}]

\thispagestyle{empty}


\begin{abstract}
    \vspace{-0.2cm}
    Methods for 3D reconstruction from posed frames require prior knowledge about the scene metric range, usually to recover matching cues along the epipolar lines and narrow the search range. However, such prior might not be directly available or estimated inaccurately in real scenarios -- e.g., outdoor 3D reconstruction from video sequences -- therefore heavily hampering performance.
    In this paper, we focus on multi-view depth estimation without requiring prior knowledge about the metric range of the scene by proposing \netname, an efficient and purely 2D framework that reverses the depth estimation and matching steps order. Moreover, we demonstrate the capability of our framework to provide rich insights about the quality of the views used for prediction.
    Additional material can be found on our \href{https://andreaconti.github.io/projects/range_agnostic_multi_view_depth}{project page}.
\end{abstract}

\vspace{-2em}

\section{Introduction}

Accurate 3D reconstruction is of profound interest in various fields: mixed reality and 3D content creation require highly detailed shape reconstruction to place digital objects in real environments, historical preservation models works of art digitally for further scientific analysis or public presentation, robotics and autonomous driving require depth estimation for navigation and planning. Active 3D sensing devices are typically preferred for high detail: LiDAR (Light Imaging Detection and Ranging) and ToF (Time of Flight) sensors can scan the scene actively with modulated laser illumination, while structured light scanners infer scene structure by projecting a known pattern and computing its deformation on surfaces.
Compared with such technologies, passive sensing from standard RGB cameras by triangulation has many advantages. Indeed, RGB cameras are energy efficient, compact in size, and may operate in various conditions. Among passive approaches, stereo vision leverages two calibrated cameras to restrict the matching problem to a 1D search space, yet requires two cameras in a constrained setting -- i.e., being nearly coplanar to allow for simpler calibration and rectification. On the other hand, a single monocular RGB camera in motion is the most flexible (as well as challenging) approach.

Traditionally, multi-view 3D reconstruction techniques can be classified in the following broad families: voxel, surface evolution, patch, or depth-based \cite{MVSGraphCuts, SemantincMVS, Li2015, galliani2015massively, yao2018mvsnet}. Despite being tackled with hand-crafted algorithms at first \cite{campbell2008using, furukawa2009accurate}, most state-of-the-art methods leverage depth-based deep learning architectures. These frameworks process a set of \textit{source} views and a \textit{reference} view and yield an estimated depth map for the latter.
Most deep architectures tackle this task by (i) extracting deep features from the images, (ii) building a cost volume sampled over the epipolar lines through a set of \textit{depth hypotheses} using differentiable homography, and (iii) predicting depth with a typically 3D convolutional module. The depth estimation pipeline sketched so far is effective but affected by some limitations.

First, according to step (ii), prior knowledge of the scene depth range is strictly required to sample depth hypotheses and build a meaningful cost volume \cite{schroeppel2022robustmvs}. Indeed, on the one hand, sampling hypotheses out of an underestimated range would make the network unable to predict depth values in out-of-range areas.
On the other hand, overestimating the depth range will result in sampling coarser hypotheses, thus reducing the fine-grained accuracy of estimated depth maps.  Unfortunately, such knowledge cannot be straightforwardly retrieved in real scenarios. When raw images are provided, camera poses can be obtained through traditional Structure-from-Motion (SfM) algorithms \cite{schoenberger2016colmap}, possibly estimating the depth range as well. However, such a range might be erroneously estimated due to a number of reasons -- e.g., untextured regions, visual occlusion, or poor field of view (FoV) overlap. We point out that many applications in which camera poses are known by other means exist (e.g., as often occurs in robotic applications \cite{jensen2014large}) and that modern mobile platforms provide pose information through dedicated inertial sensors.

Second, we argue that source frames must be carefully selected to allow proper depth estimation, with a set of requirements such as enough distance between optical centers to allow meaningful displacements, as well as sufficient cross-view overlap to allow matching. Moreover, the quality of the views must be considered as well: abrupt light or color changes, moving objects, or scene-specific occlusions must be taken into account to maximize matches. Unfortunately, all these aspects cannot be evaluated by simply considering pose similarity since many of them require an analysis of the images themselves. A better approach could be to apply SfM algorithms and analyze the distribution, quality, and amount of keypoint matches across different views, which would require additional offline processing. We argue that distinguishing meaningful matches from unreliable ones would ease the depth estimation task -- as highlighted by prior works \cite{zhang2020vismvsnet,ma2021epp} -- as well as possibly reduce the computational overhead by limiting the number of source views to those being strictly necessary to estimate accurate depth, although this latter aspect has never been explored. 

Prompted by the previous observations, we propose a novel framework that is (i) \textit{free} from prior knowledge of the depth range from which one samples hypotheses, and (ii) capable of distinguishing the most meaningful source frames among many. We will show that our \underline{R}ange \underline{A}gnostic \underline{M}ulti-View framework (\netname) enjoys the following properties:

\begin{itemize}
\item \textbf{Scene Depth Range Invariance.} Our approach is completely independent of any input depth range assumption and thus applicable everywhere a set of images along with their pose is provided. Instead of sampling features along epipolar lines according to a fixed set of depth hypotheses and then predicting depth, we reverse the mechanism: our framework iteratively updates a depth estimate dynamically moving along epipolar lines according to this latter to compute correlation scores. In this way, fixing an \textit{a priori} set of depth hypotheses is not required.

\item \textbf{Keyframes Ranking.} Our approach not only estimates depth, but also provides insights about the match quality of each source view and its contribution to the final prediction, allowing within a single inference step to rank input source views according to their actual matching against the reference view.

\end{itemize}

To assess the performance of \netname, we considered different challenging benchmarks with heterogeneous specifics.
On Blended \cite{yao2020blendedmvs} and TartanAir \cite{tartanair2020iros}, we demonstrate the capability of our framework to seamlessly estimate accurate depth in diverse scenes such as large-scale outdoor environments, top-view buildings, and indoor scenarios. Indeed, on the one hand, Blended \cite{yao2020blendedmvs} is characterized by significant pose changes, occlusions, and large FoV overlap. On the other hand, TartanAir \cite{tartanair2020iros} provides video streams characterized by small, unpredictable pose changes, where the depth range of each frame can change abruptly. 
Moreover, on UnrealStereo4K \cite{Tosi2021unrealstereo4k} we assess the generalization capability of \netname{} to video streams and the possibility of applying it to the stereo setup. To conclude, we validate our performance on DTU \cite{jensen2014large}, where the depth range is fixed. 
Along with this validation, we demonstrate the peculiar capabilities of our approach through specifically designed experiments. Fig. \ref{fig:teaser} shows the outcome of \netname{} on Blended \cite{yao2020blendedmvs}.

\section{Related Work}

We cover the most relevant research topics related to our proposal, by reviewing prior frameworks for estimating depth from multiple posed views. Depending on the settings they have been evaluated, we broadly classify them into two categories.

\textbf{Object-centric Reconstruction.} This computer vision task aims at reconstructing a 3D model -- often a point cloud -- of an arbitrarily large object by means of 2D images captured from multiple viewpoints. This task assumes a controlled environment where the object depth range is known and the viewpoints are object-centric, i.e., the object is often appearing centered in the images and fully covered by the viewpoints.
Traditional methods reconstruct the 3D structure through image points triangulation and manually engineered features. This formulation is essentially an optimization procedure based on photometric consistency across views, with shape and 3D structure priors being exploited as regularization \cite{campbell2008using, furukawa2009accurate, galliani2015massively, schonberger2016pixelwise}.
To date deep learning depth-based methods have taken the lead in this field, automating feature extraction and matching.
One of the first approaches in this direction is MVSNet \cite{yao2018mvsnet}, which builds a 3D cost volume by matching pixels along the epipolar lines. Such volume contains, for each pixel in the reference view, the variance between features sampled across the different source images employing differentiable homography. Then, a 3D convolutional network is applied as regularization, and finally, a (soft) arg-max operator is applied to extract depth, lately composed to build a global point cloud. Follow-up works mostly focused on cutting down memory requirements: \cite{yao2019recurrent} leverages 3D regularization with 2D recurrent networks \cite{cho2014gru}, while several improvements \cite{gu2020cascade, cheng2020deep, yang2020cost} follow a multi-scale approach with coarse-to-fine inference. Other extensions concern reasoning about pair-wise visibility \cite{zhang2020vismvsnet, ma2021epp}, deploying recurrent approaches \cite{ma2022cermvs, wang2022effimvs} or leveraging NeRF-inspired \cite{mildenhall2020nerf} optimization \cite{xi2022raymvsnet,chang2022rc}.
Despite all these methods being designed to predict depth, they often focus on the global 3D point cloud in a controlled environment. Our framework differs from such approaches in that \emph{a priori} depth hypotheses are not assumed at all when computing matching scores and epipolar geometry is exploited to iteratively refine estimated depth. Moreover, we do not pursue a global 3D point cloud reconstruction but focus more on fine-grained high-quality depth perception.

\textbf{Environment Reconstruction.} We categorize as environment reconstruction all those methods which seek to perform 3D reconstruction on navigable environments, such as indoor scenes. 
In this context, volumetric-based and depth-based approaches have been deployed. Volumetric-based methods seek to directly predict a global volumetric representation of the scene at once, usually a Truncated Signed Distance Function (TSDF). \cite{murez2020atlas} backprojects rays of deep features in a global voxel grid and then leverages a 3D convolutional architecture to directly regress the TSDF volume. \cite{sun2021neuralrecon} improves such approach by means of 3D recurrent layers and a coarse-to-fine approach. Further improvements by means of transformers have been proposed \cite{bozic2021transformerfusion, stier2021vortx}. Other approaches combine volumetric reasoning with depth-based reconstruction iteratively \cite{rich20213dvnet, choe2021volumefusion}. Overall, volumetric-based methods require high computational and memory resources due to the intensive usage of 3D convolutions, reconstructing the scene at once and requiring proper selection of the frames to be integrated into the TSDF volume. Moreover, they all require the scene metric range to initialize the voxel grid. On the other hand, depth-based methods solve this task by composing multiple depth maps predicted from a subset of source views of the whole scene \cite{newcombe2011kinectfusion}. Notably, \cite{sayed2022simplerecon} proposed meta-data integration in the cost volume and a 2D depth estimation module. 

However, all these approaches work in an extremely controlled (usually indoor) environment, where the scene depth range can be roughly estimated -- usually, up to a few meters. Such an assumption prevents a na\"ive extension to less constrained environments, either indoor (e.g., large, industrial factories) or outdoor. In contrast, we
design a lightweight 2D convolutional framework applicable to a wide range of scenarios. We do not aim at recovering the whole 3D reconstruction at once, but instead, we focus on accurate depth estimation since this latter covers a wider range of applications on its own as well -- and from which a 3D reconstruction can be obtained if required \cite{newcombe2011kinectfusion,sayed2022simplerecon}.

\begin{figure*}
    \centering
    \includegraphics[trim=0cm 7.2cm 3cm 0cm,clip,width=0.9\linewidth]{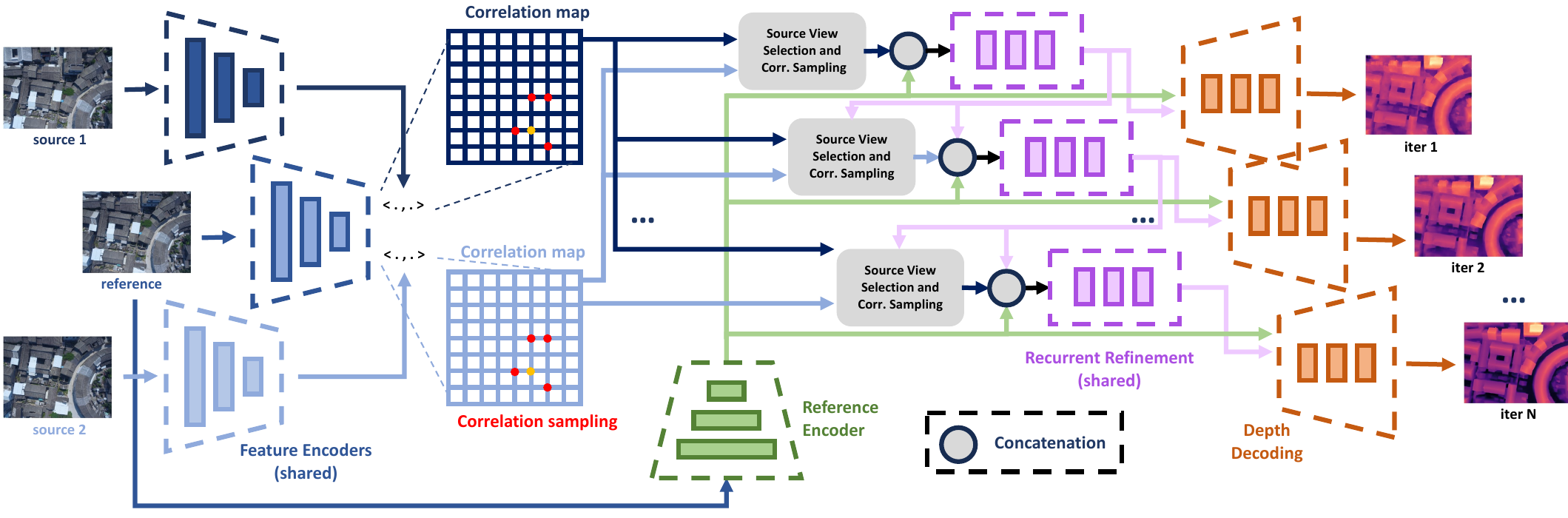}
    \vspace{-0.3cm}
    \caption{\textbf{\netname{} Architecture Description.} Our model instantiates an initial depth map and builds a pair-wise correlation table between the target view and each source image (in dark and light blue). Then, deformable sampling is iteratively performed over it, and the depth state is updated accordingly. Final depth prediction is upsampled through convex upsampling.}
    \label{fig:model}
\end{figure*}

\section{Proposed Framework}
\label{sec:method}

This paper proposes \netname, a deep framework to tackle 3D reconstruction from multiple posed views leveraging 2D convolutional layers only, and an iterative optimization procedure aimed at refining an internal depth map. Our design builds upon the following principle: given the reference view, matches over an arbitrary source view can be found given their relative pose and enough visual overlap. Thus, provided an initial depth map, dense matching costs can be computed between the reference and source views. Such information is then fed to a 2D learned module to properly refine the predicted depth map. This way, unlike any other framework that builds a cost volume relying on a set of \emph{a priori} depth hypotheses, \netname{} can dynamically navigate the matching space, while storing best matches as depth values into an inner state. Epipolar geometry comes into play since updating the stored depth values means moving over the epipolar lines defined by pose information. This approach can be thought of as reverting the common pipeline composed of (i) cost volume building and (ii) depth estimation. Moreover, we point out that the dense matching costs computed by our framework, each expressing the relationship between a specific source view and the reference one, can be regarded as a hint of the overall matchability between the views, that takes into account both FoV overlap and overall image quality. We will provide quantitative and qualitative evidence of this respectively in our experiments and supplementary material.

\textbf{Framework Overview.} Our architecture, sketched in Fig. \ref{fig:model}, can be decomposed into the following modules: (i) image features encoding, (ii) correlation sampling (iii) depth optimization, and (iv) output depth decoding. Steps (ii) and (iii) are performed multiple times for a fixed number of iterations. Thus, our model outputs a sequence of depth maps $(D^s)_{s \in \mathbb{N}}$ getting progressively more accurate.

\textbf{Features Encoding.}
Given a set of views $I^i, \ i \in [0, N]$ we refer to $I^0$ as the reference view -- i.e., the one for which we predict a depth map -- and $I^i, \ i \in [1, N]$ as the source ones. We forward each view $I^i$ to a deep convolutional encoder to extract latent features $\mathcal{F}^i \in \mathbb{R}^{\frac{W}{8} \times \frac{H}{8} \times F}$, that will be used to compute correlation scores in the next step. These are depicted in shades of blue in Fig. \ref{fig:model} and share the same weights.
Moreover, exclusively for $I^0$, we also extract a disentangled set of feature maps to provide monocular contextual information $\hat{\mathcal{F}} \in \mathbb{R}^{\frac{W}{8} \times \frac{H}{8} \times F}$, depicted in green in Fig. \ref{fig:model}. Despite the iterative nature of \netname, features are extracted only once, at bootstrap.

\textbf{Correlation Sampling.}
Once the reference and source views have been encoded into deep latent features, at any iteration the current depth estimate $D^s_{u_0v_0}$ for pixel $q^0 = [u_0, v_0, 1]^T$ -- in homogeneous coordinates -- can be used to index a specific pixel $q^i = [u_i, v_i, 1]^T$ of a source view $I^i$ as described in Eq. \ref{eq:projection}, according to camera intrinsic and extrinsic parameters $K_0,K_i$ and $E_0,E_i$.

\begin{equation}
    q^i = K_iE_iE_0^{-1}D_{u_0v_0}K_0^{-1}q^0 \label{eq:projection}
\end{equation}

This procedure leverages epipolar geometry since changing $D^s_{u_0v_0}$ means moving over the corresponding epipolar line while not being bound to \emph{a priori} depth hypotheses. Then, source view features are sampled accordingly to compute a pixel-wise correlation map $C_{u_0v_0u_iv_i}$ -- shown in Fig. \ref{fig:model} in shades of blue according to the selected source view

\begin{equation}
    \mathcal{C}_{u_0v_0u_iv_i} = \sum_{f=1}^F \mathcal{F}^0_{u_0v_0f} \mathcal{F}^i_{u_iv_if}
\end{equation}

However, this correlation map does not provide useful information on the direction in which better matches can be found. Thus, to better guide the optimization process we compute correlation scores in a neighborhood $\mathcal{N}(u_i, v_i)$ of $q^i$. Specifically, such a neighborhood is predicted by a 2D convolutional module $\Theta$, predicting $Z$ index offsets conditioned by the reference features $\hat{\mathcal{F}}$ and each iteration hidden state $\mathcal{H}^s$. The $Z$ output channels are summed to the $u_i,v_i$ coordinates to obtain the sampling locations.

\begin{equation}
    \mathcal{N}(u_i,v_i) = \big[(u_i,v_i) + \Theta( \hat{\mathcal{F}}_{u_0v_0},\mathcal{H}^s)_z, \hspace{0.1cm} z \in Z \big]
    \label{eq:sampling}
\end{equation}

This mechanism resembles deformable convolutions \cite{Dai_2017_ICCV} in that it samples from a dynamic neighborhood, yet it differs since it does not accomplish a proper convolution with the sampled features but instead performs correlation with the features sampled from another view. It is worth observing that since $\Theta$ is conditioned with a state that changes at each iteration, these offsets may change at each iteration accordingly. The reference view context potentially allows to adaptively sample correlation scores in a narrower or wider region depending on the ambiguity of the reference image itself, like in the presence of object boundaries or low-textured regions.

The correlation sampling mechanism described so far works on a single source view at a time. This is a problem when multiple source views are available. Following existing approaches, correlation features could be extracted from each source view and then fused together. However, this approach would require developing a merging mechanism independent of the number of source views -- e.g., simple concatenation would be unsuitable as it fixes the number of input channels. Many existing models compute feature-level variance to combine the volumes \cite{yao2018mvsnet}. Instead, we propose to use a different source view for each update step in our framework, following a simple round-robin approach. This methodology is simple and elegant since it exploits the iterative nature of our architecture, it does not require hand-crafted fusing modules, and can be extended to any variable number of source views. While different scheduling strategies can be employed, in this paper we limit to the simplest one and leave their in-depth study to future developments. We delve into further analysis on this in the supplementary material, where we show that such an approach is also invariant to the source views order.

\textbf{Keyframes Ranking.} Since \netname{} exploits a single source view at each iteration, $\mathcal{C}$ is related to a single specific source view, as it contains correlation scores between deep features of the source and reference views. Such correlation grows as the source view features are correctly projected over the reference view, and thus can be regarded as a score about matching quality \cite{lindenberger2023lightglue}. It is worth mentioning that such a score is susceptible to the FoV overlap but also moving objects, blurring, or any other factor violating the multi-view geometry assumptions or that the encoding procedure is not robust against. Thus, it is directly linked with the capability of the network to exploit such source views to improve its prediction. Accordingly, we can rank each view by taking the last correlation map computed for each source view and averaging it over the spatial dimensions. Since the network learns to perform good matches directly from depth supervision, there is no need to directly supervise this output which is a byproduct of our approach.

\textbf{Depth Optimization.}
With the components defined so far, \netname{} estimates a depth map for the reference view iteratively. At any stage $s$, a shallow recurrent network -- in purple in Fig. \ref{fig:model} -- made of a Gated Recurrent Unit processes the sampled correlation scores $\mathcal{C}$ and reference features $\hat{\mathcal{F}}$ together with the current hidden state $\mathcal{H}^{s}$ and depth map $D^{s}$ (i.e., coming from the previous optimization stage) to output an updated hidden state $\mathcal{H}^{s+1}$. Then, two convolutional layers predict a depth update $\Delta D^s$ yielding a refined depth map $D^{s+1} := D^{s} + \Delta D^s$.
At bootstrap, $D^0$ is initialized to zero and then the aforementioned iterative process allows for rapidly updating the depth map state towards a final, accurate prediction. 
At the first iteration, the correlation scores $\mathcal{C}$ will not be meaningful for depth, thus the network learns to provide a monocular initialization for $D^1$. Other approaches could consist of either randomly initializing $D^0$ or inserting a further module to learn an initialization. The former would be inaccurate if no information about the depth range is assumed, the latter is equivalent to zero initialization yet requires an extra component.

\textbf{Depth Decoding.}
Since \netname{} iterates at a lower resolution, a final upsampling of the depth maps to the original input resolution is required. Many approaches leverage either bilinear upsampling \cite{yao2018mvsnet, yao2019recurrent, gu2020cascade, cheng2020deep, wang2021patchmatchnet} or a deep convolutional decoder. Instead, we compute a weighting mask with an upsampling module -- in orange in Fig. \ref{fig:memory-time} -- fed with the latest hidden state $\mathcal{H}^{s+1}$ and the reference view features $\hat{\mathcal{F}}$, then we perform convex upsampling \cite{Teed2020raft}. This approach is faster than employing a decoder and yields much better results compared to using hand-crafted upsampling approaches.

\textbf{Loss Function.} \netname{} is supervised by computing a simple L1 loss between the ground-truth depth $D_\text{gt}$ and each estimated depth map, with a weight decay $\gamma$ set to 0.8

\begin{equation}
\mathcal{L} = \sum_{s=1}^S \gamma^{S-s} || D_\text{gt} - D^s  ||_1
\end{equation}

\begin{table*}[t]
    \resizebox{\textwidth}{!}{
    \begin{tabular}{@{}c@{}c@{}c@{}}
    \begin{tabular}{l|} \hline \hline
        \multirow{2}*{Method}                       \\
                                                    \\
        \hline                                       
        \ignore{Y. }Yao et al. \cite{yao2018mvsnet}             \\  
        \ignore{Y. }Yao et al. \cite{yao2019recurrent}          \\  
        \ignore{S. }Cheng et al. \cite{cheng2020deep}           \\  
        \ignore{F. }Wang et al. \cite{wang2021patchmatchnet}    \\  
        \ignore{X. }Gu et al. \cite{gu2020cascade}              \\  
        \ignore{J. }Zhang et al. \cite{zhang2020vismvsnet}      \\  
        \ignore{M. }Sayed et al. \cite{sayed2022simplerecon}    \\ 
        \ignore{Z. }Ma et al. \cite{ma2022cermvs}               \\  
        \netname{} (ours)                                           \\ 
        \hline \hline
    \end{tabular}
    &
    \begin{tabular}{ccccccc|}
        \hline \hline
                                \multicolumn{7}{c|}{Ground Truth Depth Range}                           \\
        MAE         & RMSE        & $>$1 m      & $>$2 m      & $>$3 m      & $>$4 m      & $>$8 m      \\
        \hline
        0.6168      & 1.5943      & 0.1392      & 0.0731      & 0.0457      & 0.0309      & 0.0103      \\
        0.7815      & 1.7397      & 0.1864      & 0.1007      & 0.0637      & 0.0433      & 0.0141      \\
        0.3590      & 1.3589      & 0.0704      & 0.0378      & 0.0244      & 0.0171      & 0.0064      \\
        0.3849      & 1.3581      & 0.0749      & 0.0386      & 0.0247      & 0.0175      & 0.0067      \\
        0.3684      & 1.3449      & 0.0714      & 0.0365      & 0.0234      & 0.0165      & 0.0062      \\
        0.3318      & 1.2396      & 0.0662      & 0.0323      & 0.0197      & 0.0133      & 0.0044      \\
        0.5921      & 1.4340      & 0.1404      & 0.0584      & 0.0308      & 0.0191      & 0.0057      \\
        2.1666      & 26.934      & 0.0752      & 0.0441      & 0.0316      & 0.0247      & 0.0138      \\
        \bt{0.2982} & \bt{1.1724} & \bt{0.0645} & \bt{0.0285} & \bt{0.0159} & \bt{0.0102} & \bt{0.0033} \\
        \hline \hline
    \end{tabular}
    &
    \begin{tabular}{ccccccc}
        \hline \hline
                                    \multicolumn{7}{c}{Unique Depth Range}                              \\
        MAE         & RMSE        & $>$1 m      & $>$2 m      & $>$3 m      & $>$4 m      & $>$8 m      \\
        \hline     
        2.1115      & 5.3122      & 0.3021      & 0.1637      & 0.1194      & 0.0964      & 0.0526      \\
        1.2568      & 2.6033      & 0.2918      & 0.1464      & 0.0933      & 0.0676      & 0.0286      \\
        1.6489      & 4.1094      & 0.1844      & 0.1235      & 0.1046      & 0.0932      & 0.0602      \\
        22.420      & 25.026      & 0.6721      & 0.5067      & 0.4989      & 0.4956      & 0.4761      \\
        1.8978      & 4.2927      & 0.2341      & 0.1427      & 0.1101      & 0.0921      & 0.0597      \\
        1.0536      & 2.8939      & 0.1682      & 0.0913      & 0.0643      & 0.0508      & 0.0285      \\
        0.5921      & 1.4340      & 0.1404      & 0.0584      & 0.0308      & 0.0191      & 0.0057      \\
        8.2120      & 55.710      & 0.5780      & 0.5400      & 0.4960      & 0.3540      & 1.1150      \\
        \bt{0.2982} & \bt{1.1724} & \bt{0.0645} & \bt{0.2849} & \bt{0.0159} & \bt{0.0102} & \bt{0.0033} \\
        \hline \hline
    \end{tabular}
    \\
    & (a) & (b)
    \end{tabular}
    }
    \centering
    \vspace{-1.2em}
    \caption{\textbf{Blended Benchmark}. We report comparisons with existing methods under two settings: (a) by providing full knowledge about the scene depth to each method, (b)  
    by assuming a unique depth range to cover the whole test set. Since \netname{} does not exploit any knowledge about such range, its accuracy is not affected by the setup, unlike others.}
    \label{tab:blended-benchmark}
\end{table*}

\section{Experimental Results}

To assess the effectiveness of our approach in the most challenging environments available, we perform experiments on Blended \cite{yao2020blendedmvs}, TartanAir \cite{tartanair2020iros}, UnrealStereo4K \cite{Tosi2021unrealstereo4k} and DTU \cite{jensen2014large}.
These datasets cover a wide range of applications of interest -- e.g. outdoor multi-view settings, monocular video sequences, stereo perception, and object-centric indoor setups. Specifically, Blended \cite{yao2020blendedmvs} provides large complex aerial views of buildings characterized by high inter-view pose displacements, while TartanAir provides outdoor and indoor monocular video sequences with small but unpredictable pose changes. In both, it is difficult to decide the depth range \emph{a priori} as it is not usually constant within the same scene as well between scenes. On UnrealStereo4K \cite{Tosi2021unrealstereo4k}, we assess the generalization capability of \netname{} and the possibility to perform stereo depth perception seamlessly -- to further support its strong matching effectiveness. Finally, DTU \cite{jensen2014large} provides interesting cues about the performance in a controlled environment, where the depth range can be accurately known \emph{a priori}. Our framework consists of 5.9M parameters, the detailed architecture, training setting and evaluation parameters are reported in the supplementary material. In any experiment, we compute the mean absolute error (MAE), root mean squared error (RMSE), and the percentage of pixels having depth error larger than a given threshold ($>\tau$).

\begin{figure*}[t]
    \centering
    \resizebox{\textwidth}{!}{
    \begin{tabular}{@{}ccccc@{}}
    \scriptsize \textbf{Reference View} &
    \scriptsize \textbf{\mvsnet} &
    \scriptsize \textbf{\patchnet} &
    \scriptsize \textbf{Ours} &
    \scriptsize \textbf{Ground Truth} \\
    \includegraphics[width=0.18\textwidth]{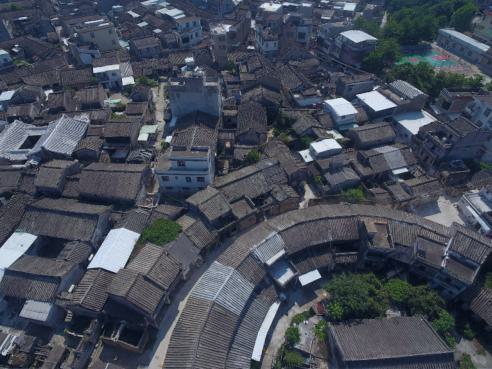}           &
    \includegraphics[width=0.18\textwidth]{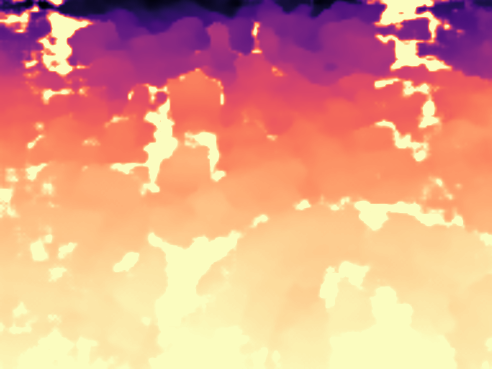}        &
    \includegraphics[width=0.18\textwidth]{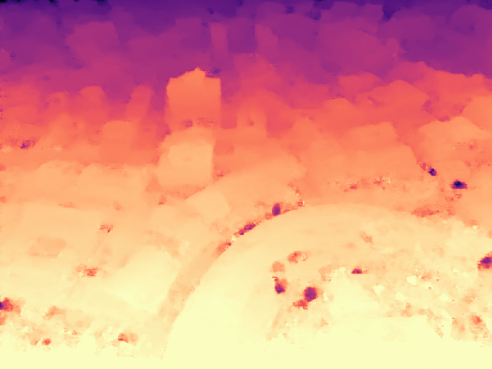} &
    \includegraphics[width=0.18\textwidth]{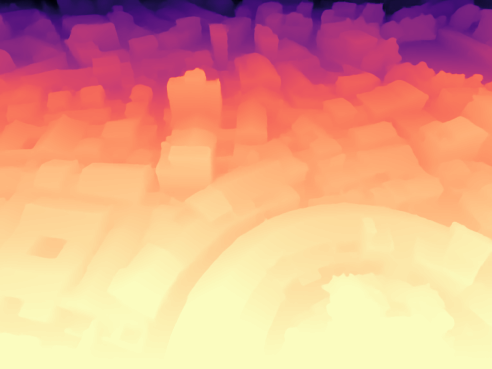}          &
    \includegraphics[width=0.18\textwidth]{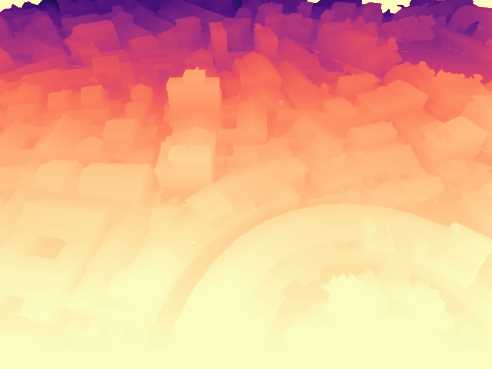}            \\
    \includegraphics[width=0.18\textwidth]{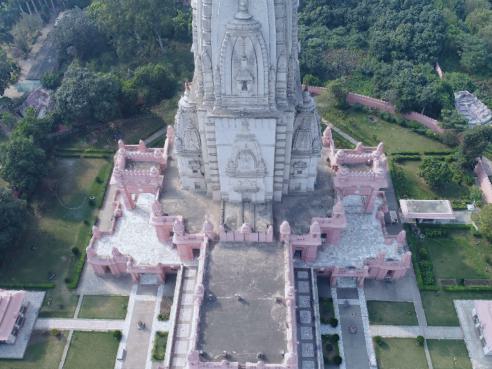}           &
    \includegraphics[width=0.18\textwidth]{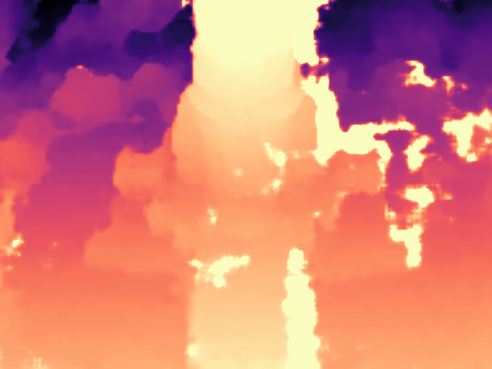}        &
    \includegraphics[width=0.18\textwidth]{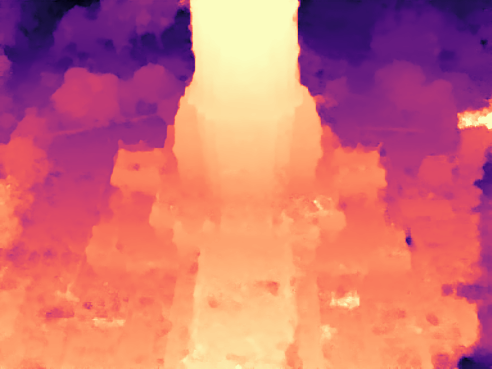} &
    \includegraphics[width=0.18\textwidth]{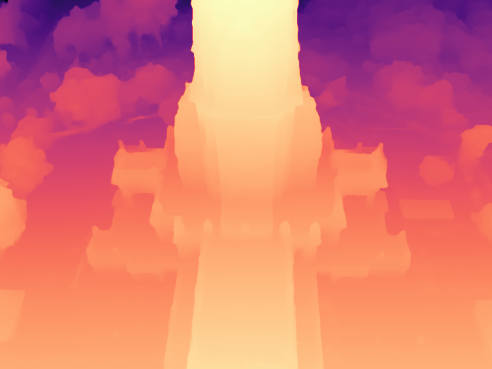}          &
    \includegraphics[width=0.18\textwidth]{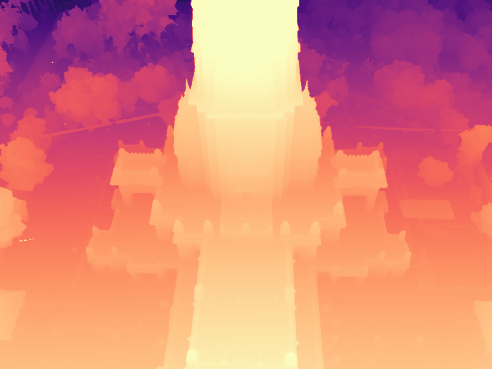}
    \end{tabular}
    }
    \vspace{-1em}
    \caption{\textbf{Qualitative results on Blended.}
    Our approach extracts consistent and visually pleasant depth maps, not showing any visible outliers as can be observed in competitor methods.
    }
    \label{fig:blended-qualitatives}
    \vspace{-1em}
\end{figure*}

\textbf{Blended Benchmark.}
The Blended dataset \cite{yao2020blendedmvs} collects 110K images from about 500 scenes, rendered from meshes obtained through 3D reconstruction pipelines. It features large overhead views where the scene depth range would be hard to be properly recovered in a real use case, but also several objects closeups. Following \cite{poggi2022guided}, we test any method with five input images on the standard test set, composed of 7 heterogeneous scenes. We first evaluate \netname{} following the protocol and metrics by \cite{poggi2022guided} to assess the accuracy of predicted depth maps. In this experiment, each method except ours exploits the reference view ground-truth depth range. Results are collected in Table~\ref{tab:blended-benchmark}~(a). Our framework consistently produces more accurate depth maps, despite not exploiting any knowledge about the depth range of any scene. We also point out how \netname{} produces much better depth maps than other methods, which show frequent artifacts as shown in Fig.~\ref{fig:blended-qualitatives}.

\textbf{Depth Range Analysis.} We now focus on the importance of not depending on prior knowledge about the scene depth range. Purposely, we design a benchmark tailored to study this specific aspect on the Blended test set, given the wide set of heterogeneous scenes with depth ranges varying from a few meters up to hundreds.
In Table \ref{tab:blended-benchmark} (b) each competitor relying on the depth range is fed with a global unique depth range, computed to cover the whole dataset one. To ease the task for competitor methods we perform the following steps: (i) we normalize the extrinsic translations between the reference and the source views to have a mean value equal to 1 and compute the corresponding depth scaling factor, then (ii) we compute the mean depth on the test set using the rescaled ground-truth depth and estimate an appropriate set of depth hypotheses equal for every sample to cover the whole dataset depth range, finally (iii) the depth predictions by models processing depth hypotheses are scaled back to the original metrical range. This procedure acts on the scene scale only, not affecting the performance of trained networks, except for changing the set of depth hypotheses used to build cost volumes. This procedure is a precaution adopted to have closer values for minimum and maximum depth in large-scale scenes, to ease numerical stability. Nonetheless, results reported in Table~\ref{tab:blended-benchmark}~(b) emphasize how the lack of precise knowledge about the depth range of the scene heavily penalizes existing methods, whereas \netname{} remains unaffected.

\begin{figure}[t]
    \centering
    \includegraphics[width=0.9\linewidth]{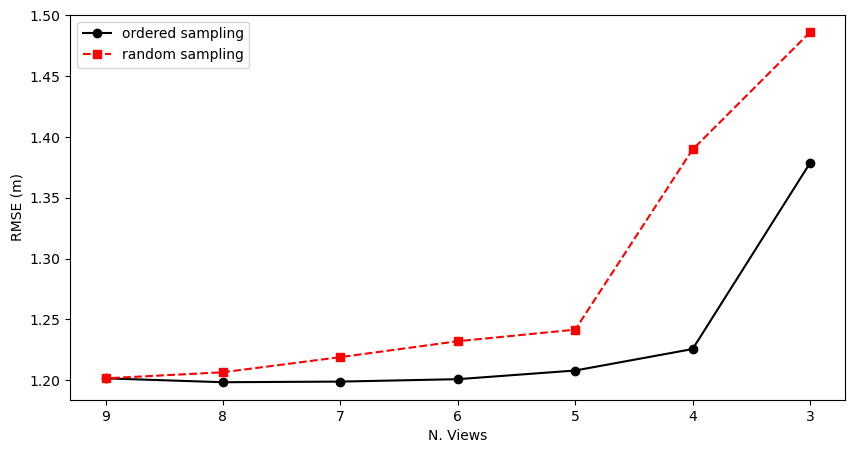}
    \vspace{-1em}
    \caption{\textbf{Keyframes Ranking.} We plot RMSE achieved by dropping input views in random order (red) or according with the ranking information provided by \netname{} (black).
    }
    \label{fig:view-pruning}
    \vspace{-1em}
\end{figure}

\begin{table*}[t]
    \resizebox{\textwidth}{!}{
    \begin{tabular}{@{}c@{}c@{}c@{}}
    \begin{tabular}{l|} \hline \hline
        \multirow{2}*{Method}                       \\
                                                    \\
        \hline                                       
        \mvsnet \\
        \dhcnet \\
        \ucsnet \\
        \patchnet \\
        \casnet \\
        \vismvsnet \\
        \smplrcn \\
        \cermvs \\
        \netname{} (ours)                                                    \\ 
        \hline \hline
        \raftstereo \\
        \hline \hline
    \end{tabular}
    &
    \begin{tabular}{ccccccc|}
        \hline \hline
                                \multicolumn{7}{c|}{ Monocular Video Benchmark} \\
       MAE (m) & RMSE (m) &     $>$1 m &     $>$2 m &     $>$3 m &     $>$4 m &     $>$8 m \\
       \hline
       5.330 &      8.638 &      0.590 &      0.418 &      0.329 &      0.273 &      0.166 \\
       4.077 &      7.106 &      0.550 &      0.376 &      0.278 &      0.216 &      0.118 \\
       6.511 &      9.935 &      0.635 &      0.468 &      0.375 &      0.314 &      0.196 \\
       7.883 &      10.84 &      0.637 &      0.495 &      0.413 &      0.359 &      0.244 \\
       6.364 &      9.521 &      0.630 &      0.469 &      0.373 &      0.314 &      0.197 \\
       6.287 &      8.949 &      0.602 &      0.454 &      0.373 &      0.319 &      0.208 \\
       5.460 &      7.951 &      0.743 &      0.566 &      0.439 &      0.350 &      0.168 \\
       7.344 &      13.74 &      0.645 &      0.474 &      0.372 &      0.306 &      0.187 \\
       \bt{3.773} & \bt{6.876} & \bt{0.514} & \bt{0.353} & \bt{0.264} & \bt{0.201} & \bt{0.101} \\
       \hline \hline
       -     &      -     &      -     &      -     &      -     &      -     &      -      \\
       \hline \hline
    \end{tabular}
    &
    \begin{tabular}{cccccc}
        \hline \hline
                                    \multicolumn{6}{c}{Stereo Benchmark}  \\
        MAE (px)    & RMSE (px)   & $>$1 px      & $>$2 px      & $>$3 px      & $>$4 px \\
        \hline     
        9.142       & 16.142      & 0.685       & 0.456      & 0.352       & 0.304       \\
        8.963       & 16.057      & 0.663       & 0.425      & 0.320       & 0.270       \\
        7.539       & 16.096      & 0.357       & 0.261      & 0.230       & 0.211       \\
        3.485       & 10.462      & 0.240       & 0.160      & 0.129       & 0.112       \\
        18.408      & 68.167      & 0.424       & 0.342      & 0.304       & 0.278       \\
         9.899      & 26.357      & 0.319       & 0.256      & 0.226       & 0.206       \\
        22.323      & 27.022      & 0.979       & 0.959      & 0.944       & 0.924       \\
        3.832       & 10.156      & 0.268       & 0.196      & 0.161       & 0.137       \\
        \bt{1.837}  & \bt{5.7930}  & \bt{0.157}  & \bt{0.099} & \bt{0.076}  & \bt{0.063}  \\
       \hline \hline
        1.646       & 4.8090       & 0.139       & 0.089      & 0.069       & 0.057       \\
        \hline \hline
    \end{tabular}
    \\
    & (a) & (b)
    \end{tabular}
    }
    \centering
    \vspace{-1.2em}
    \caption{\textbf{UnrealStereo4k Benchmark}. Application of our and competitor frameworks to UnrealStereo4k either selecting source views from monocular video sequences (a) or using rectified left and right stereo couples as target and source views (b). We process images at $960 \times 544$ resolution. 
    }
    \label{tab:unrealstereo4k-benchmark}
    \vspace{-0.8em}
\end{table*}

\textbf{Keyframes Ranking.}
To assess the quality of the source views ranking produced by \netname, we perform a peculiar experiment: for each sample in the Blended test set, we rank its source frames according to the method described in Section \ref{sec:method}. Then, we progressively decrease the number of source frames provided to our framework by selecting them either randomly or according to our ranking. Fig. \ref{fig:view-pruning} shows the results of this experiment. Selecting frames according to our ranking approach yields an overall error that diverges much more slowly. Despite not being the direct goal of this paper, this experiment lays the groundwork for interesting potential applications like automatically removing blurred, out-of-view, or non-static frames from the set of source views, as may happen on video sequences.

\begin{table}[t]
\centering
\resizebox{\linewidth}{!}{
\begin{tabular}{@{}l|ccccccc@{}}
    \hline \hline
    Method                                               & MAE        & RMSE       & $>$1 m     & $>$2 m     & $>$3 m     & $>$4 m     & $>$8 m     \\
    \hline       
    \mvsnet                                              & 1.887      & 4.457      & 0.278      & 0.183      & 0.138      & 0.110      & 0.056      \\
    \dhcnet                                              & 2.191      & 4.729      & 0.346      & 0.228      & 0.170      & 0.134      & 0.066      \\
    \ucsnet                                              & 1.461      & 3.860      & 0.216      & 0.141      & 0.106      & 0.084      & 0.043      \\
    \patchnet                                            & 2.351      & 4.980      & 0.331      & 0.228      & 0.176      & 0.144      & 0.078      \\
    \casnet                                              & 1.582      & 4.017      & 0.230      & 0.150      & 0.113      & 0.090      & 0.047      \\
    \smplrcn                                             & 1.561      & 3.303      & 0.316      & 0.167      & 0.112      & 0.083      & 0.036      \\ 
    \cermvs                                              & 3.405      & 10.20      & 0.322      & 0.211      & 0.163      & 0.134      & 0.077      \\
    \netname{} (ours)                                    & \bt{1.258} & \bt{3.289} & \bt{0.203} & \bt{0.125} & \bt{0.090} & \bt{0.070} & \bt{0.034} \\
    \hline \hline
\end{tabular}
}
\vspace{-0.8em}
\caption{\textbf{TartanAir Benchmark.} Results achieved by existing multi-view frameworks and ours on TartanAir \cite{tartanair2020iros}. Our method consistently demonstrates better performance.}
\label{tab:tartanair-benchmark}
\vspace{-1.3em}
\end{table}

\textbf{UnrealStereo4k Benchmark.} The UnrealStereo4K dataset \cite{Tosi2021unrealstereo4k} provides synthetic stereo videos in different challenging scenarios. On this dataset, we seek to assess the generalization capabilities of our architecture on monocular video sequences and, peculiarly, at dealing with the rectified stereo use case. Thus, we use the Blended pre-trained models without any kind of fine-tuning. Concerning the stereo perception application, we use the right view as the reference view and the left as the source one. Even in this case, we provide the ground-truth depth range in input to all the methods requiring \emph{a priori} depth hypotheses. However, it is worth mentioning that from a practical point of view, this is an unrealistic assumption when dealing with left-right stereo pairs, yet necessary to deploy multi-view networks relying on depth hypotheses in this setting -- except for ours. In Table \ref{tab:unrealstereo4k-benchmark} we leverage five consecutive frames (a) or a single stereo pair (b). In both cases, we achieve substantial improvements over existing models, highlighting a dramatic margin by \netname{} over other solutions. As a reference, we also report the performance achieved by \cite{lipson2021raftstereo}, a state-of-the-art stereo network trained on a variety of stereo datasets, to highlight how close our solution gets to it, despite not being trained explicitly to deal with this specific setting -- 
since Blended is not even a stereo dataset. This evidence further supports the great flexibility of our approach.

\begin{figure}[t]
    \renewcommand{\tabcolsep}{2pt}
    \centering
    \begin{tabular}{@{}ccc@{}}
    \multicolumn{1}{c}{\scriptsize \textbf{Reference}}    & 
    \multicolumn{1}{c}{\scriptsize \textbf{Prediction}}   & 
    \multicolumn{1}{c}{\scriptsize \textbf{Ground Truth}} \\
    \includegraphics[width=0.32\linewidth]{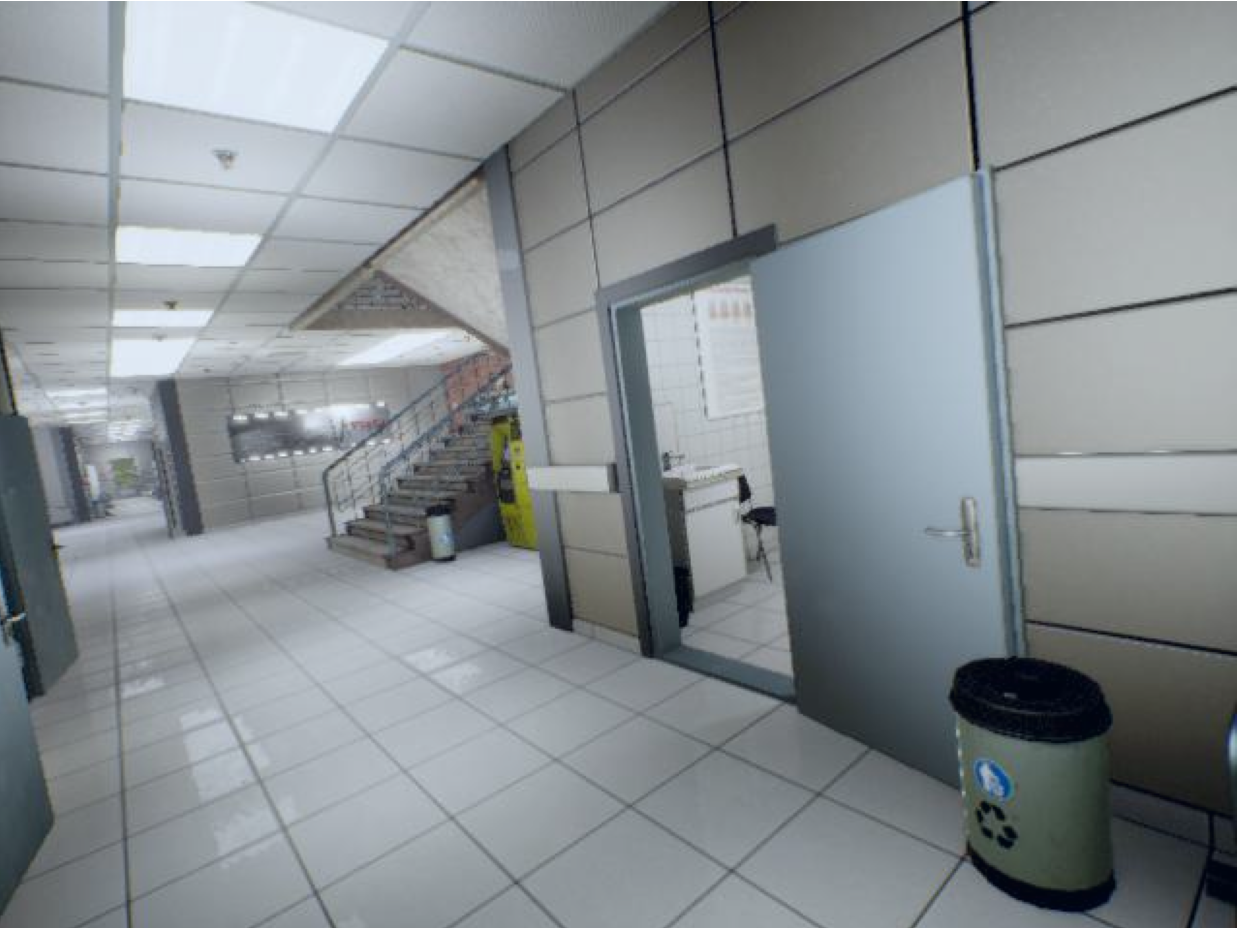}  &
    \includegraphics[width=0.32\linewidth]{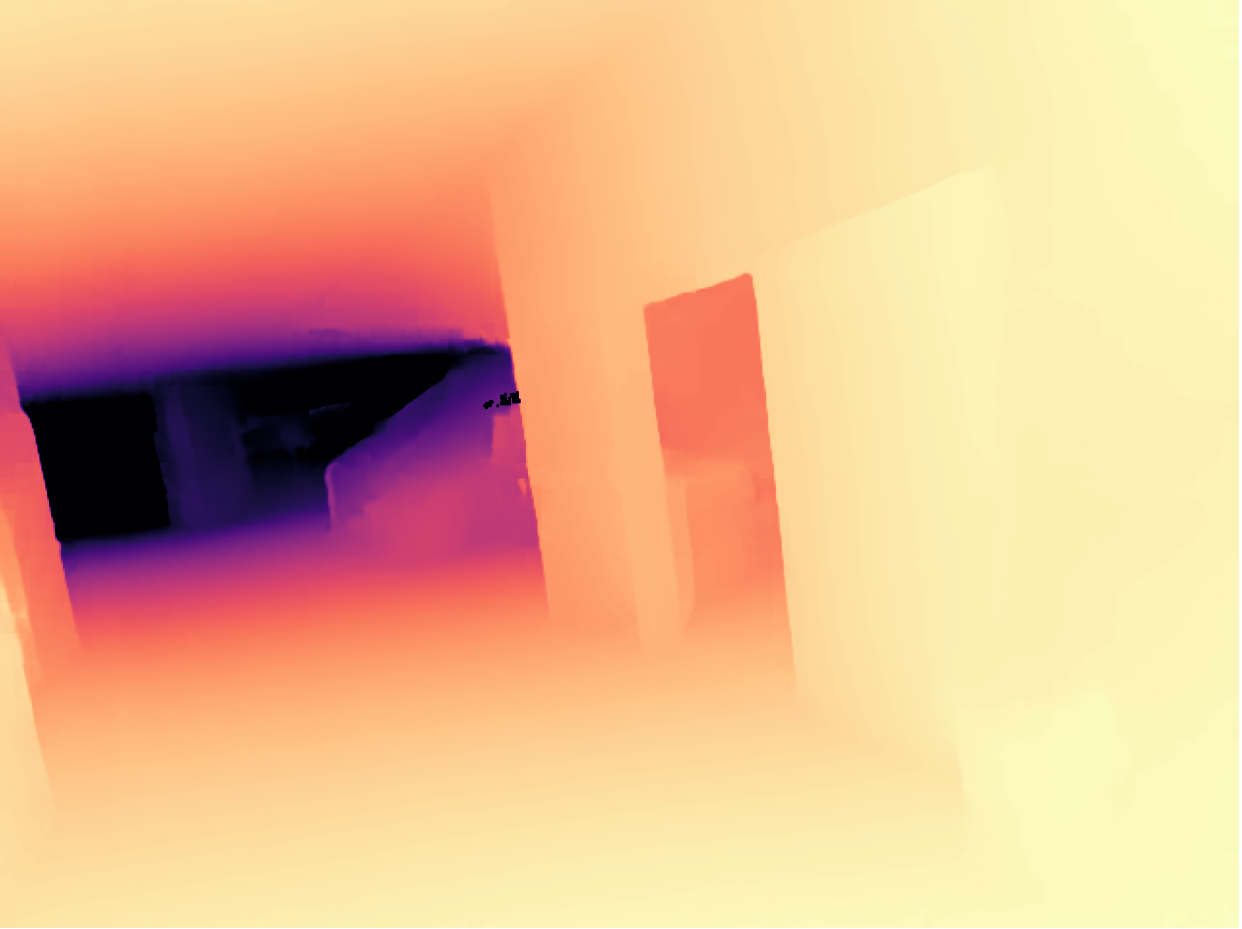} &
    \includegraphics[width=0.32\linewidth]{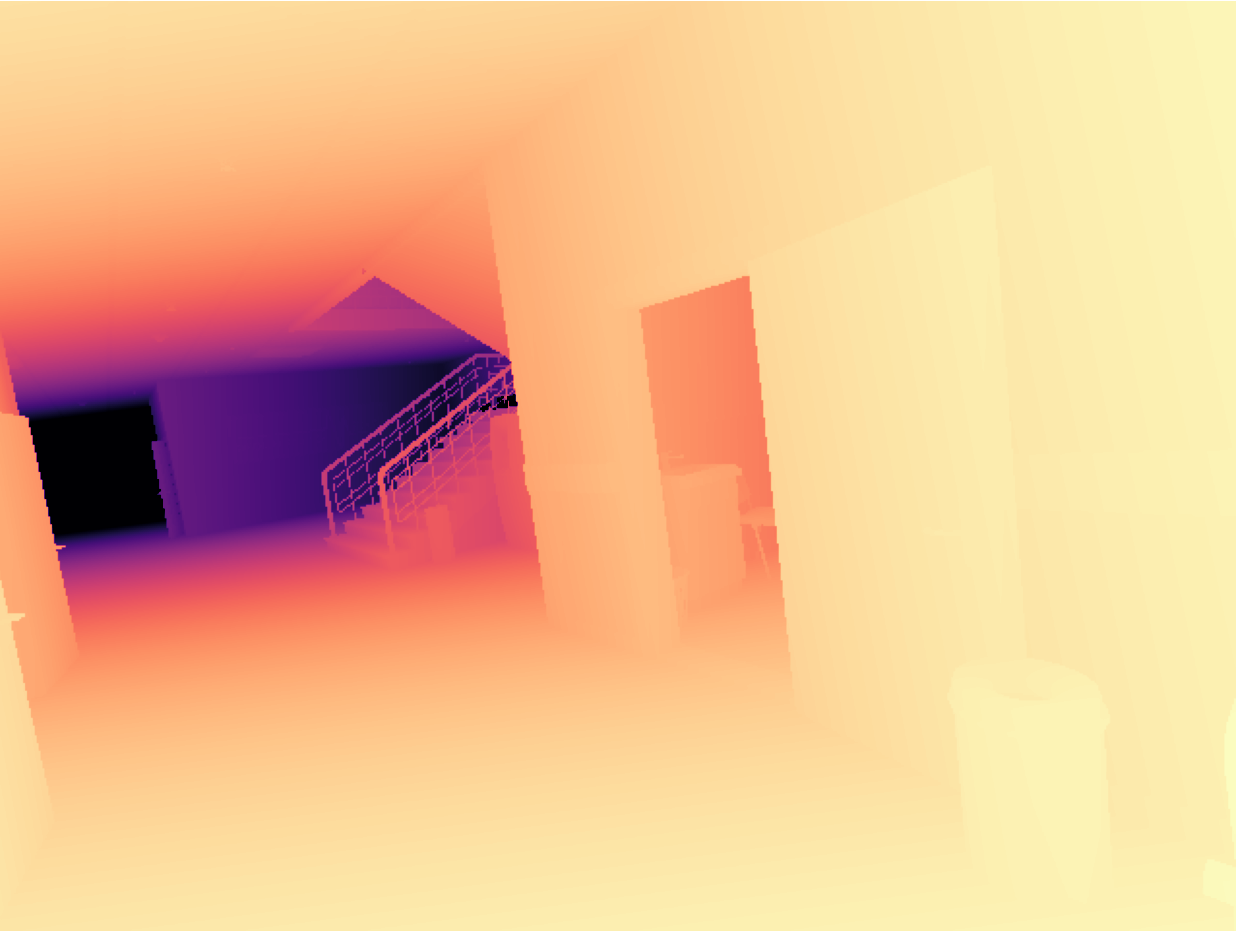}         \\
    \includegraphics[width=0.32\linewidth]{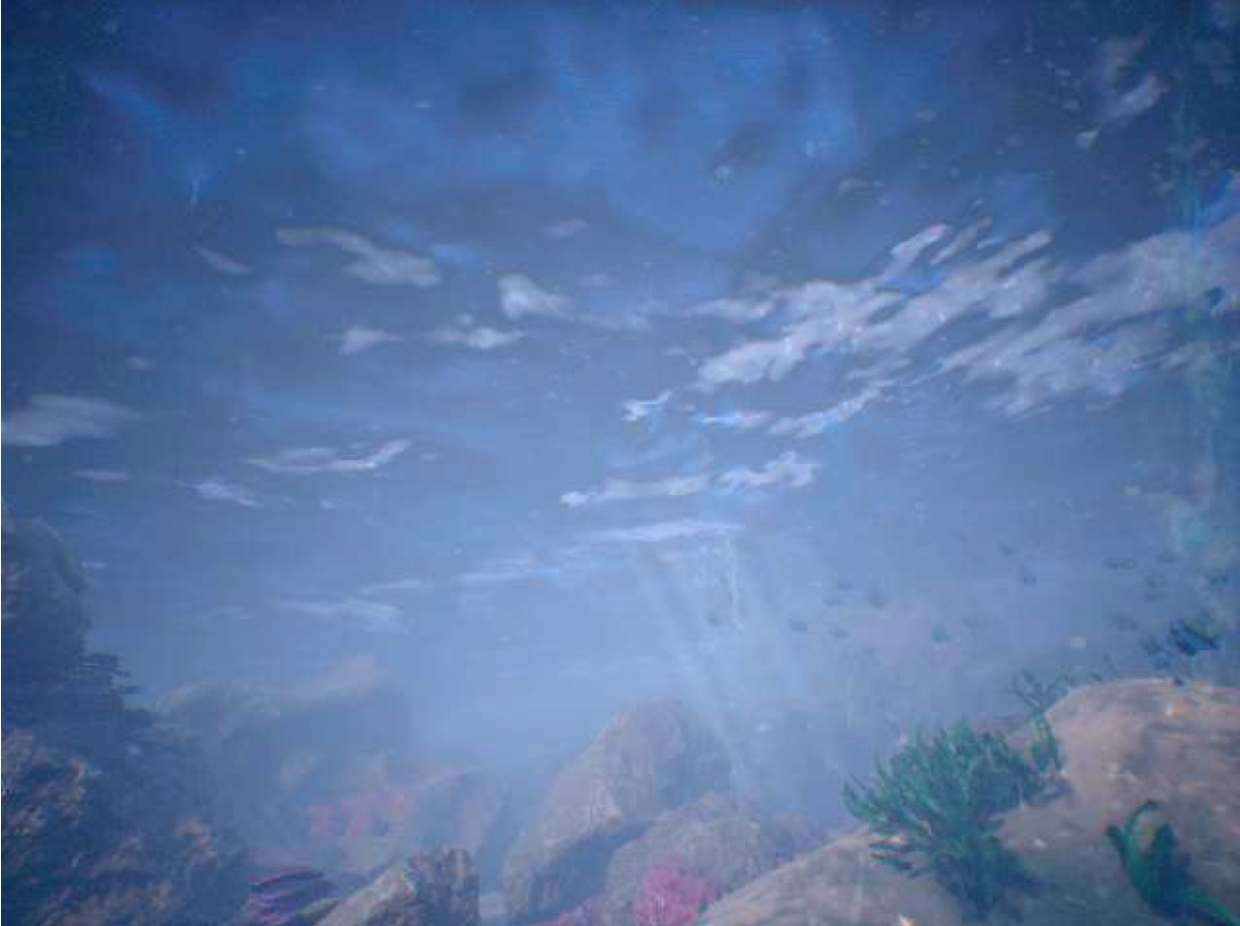}  &
    \includegraphics[width=0.32\linewidth]{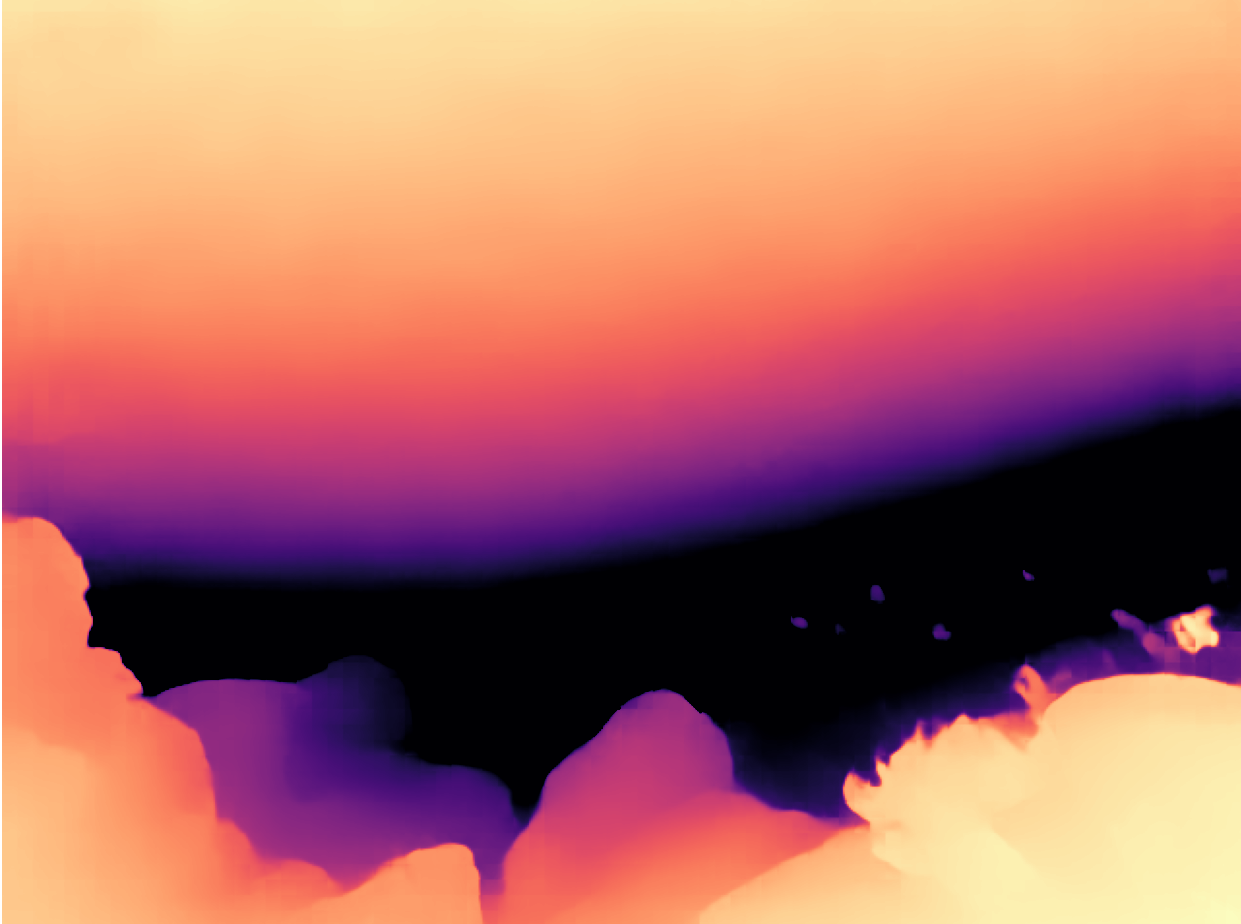} &
    \includegraphics[width=0.32\linewidth]{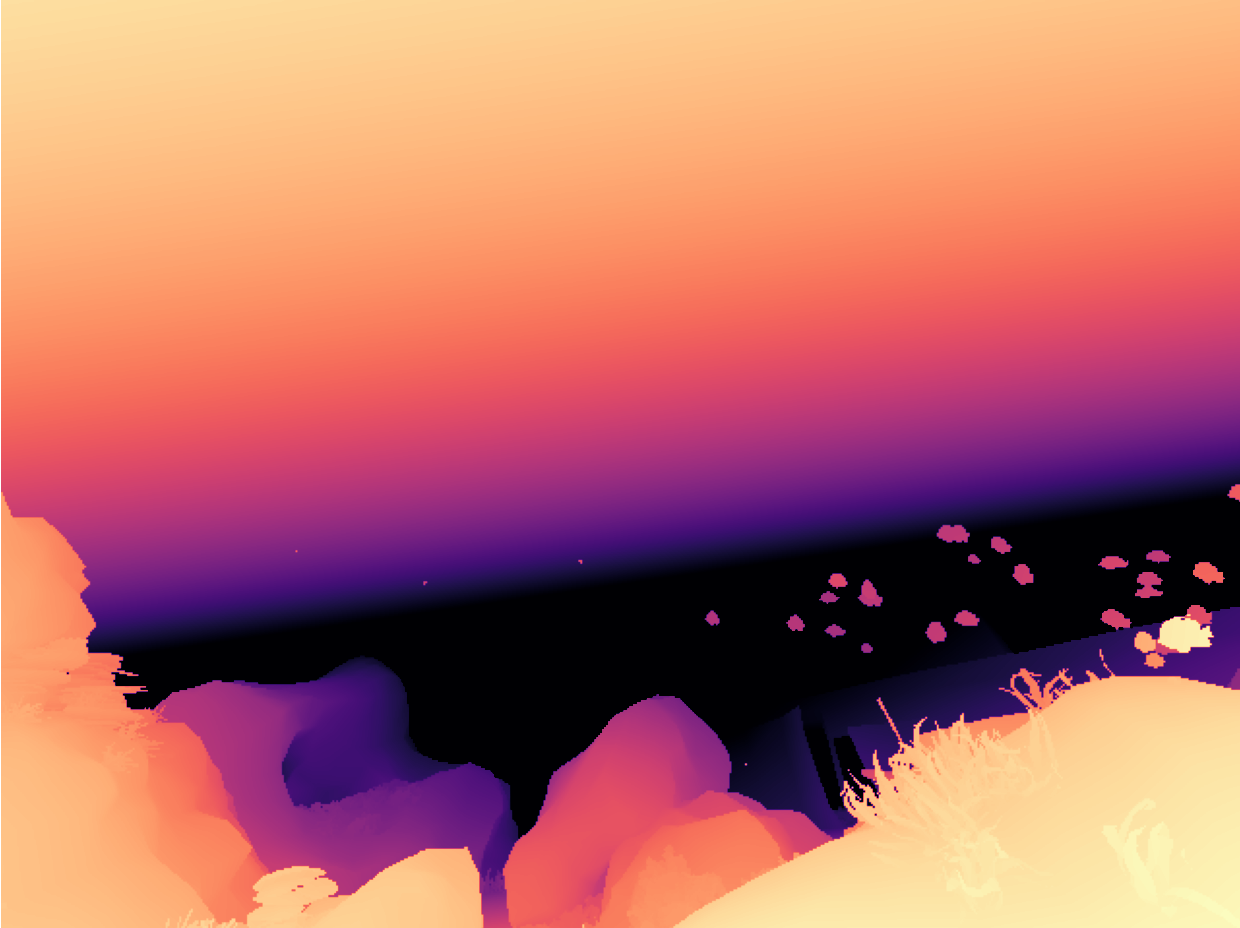}         \\
    \includegraphics[width=0.32\linewidth]{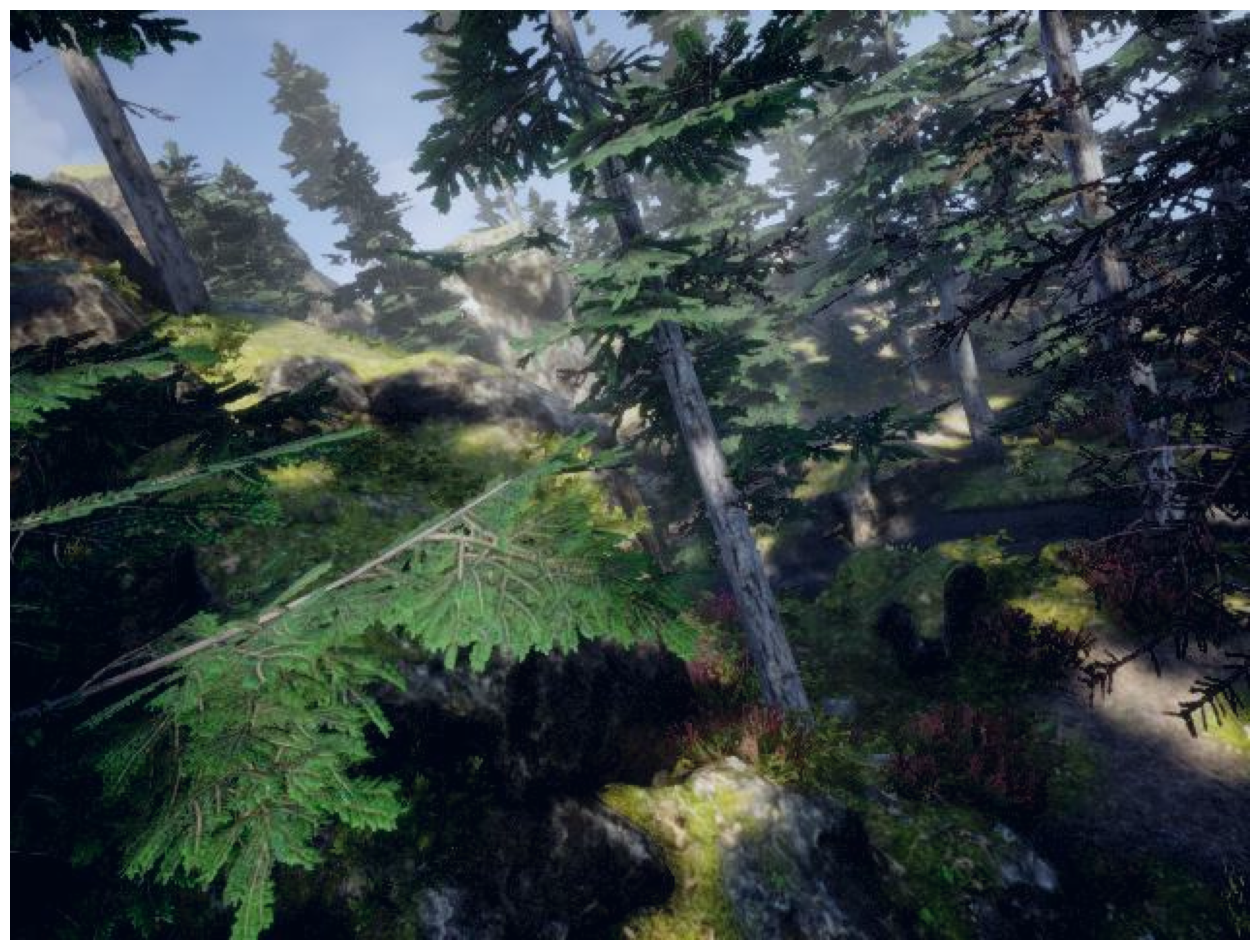}  &
    \includegraphics[width=0.32\linewidth]{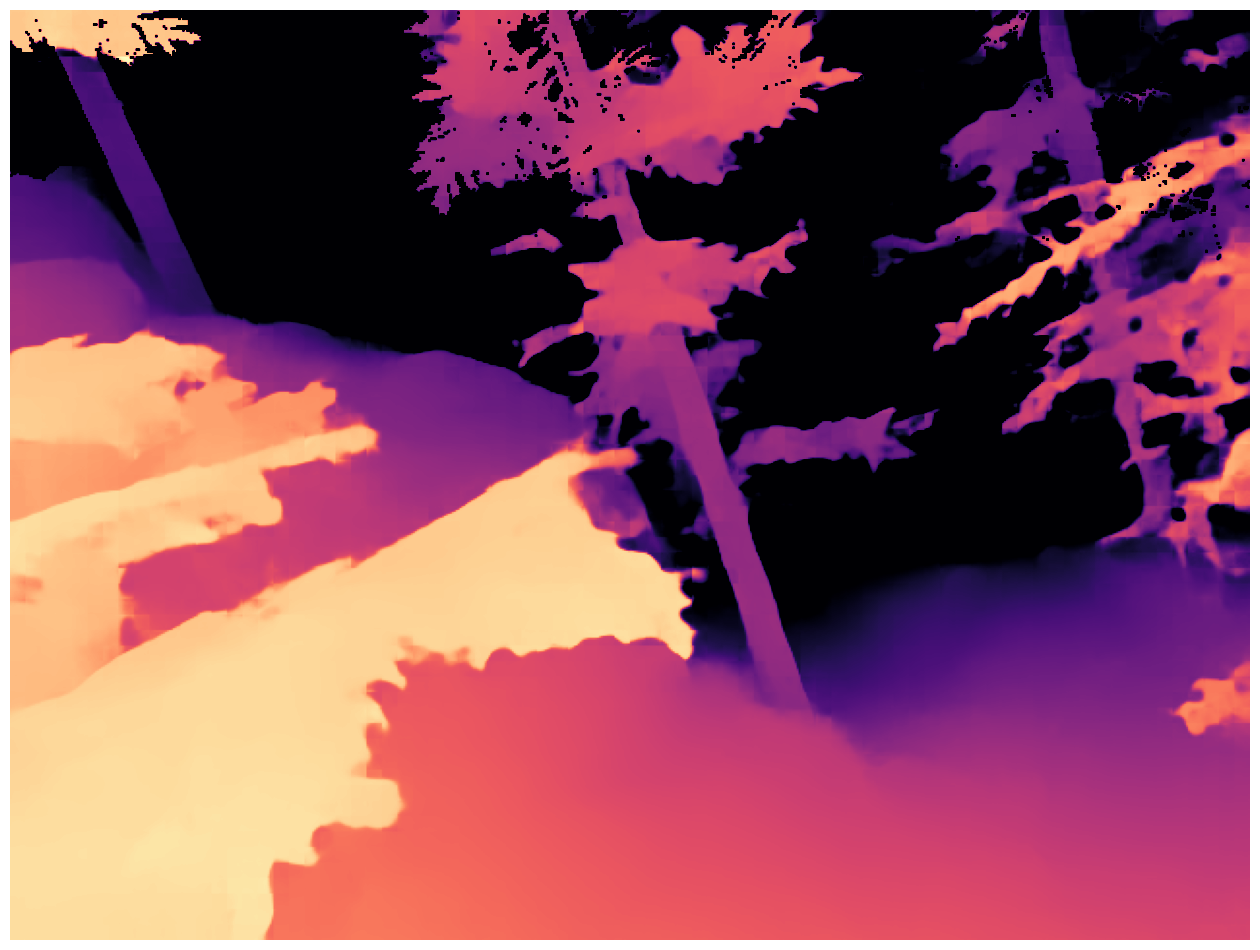} &
    \includegraphics[width=0.32\linewidth]{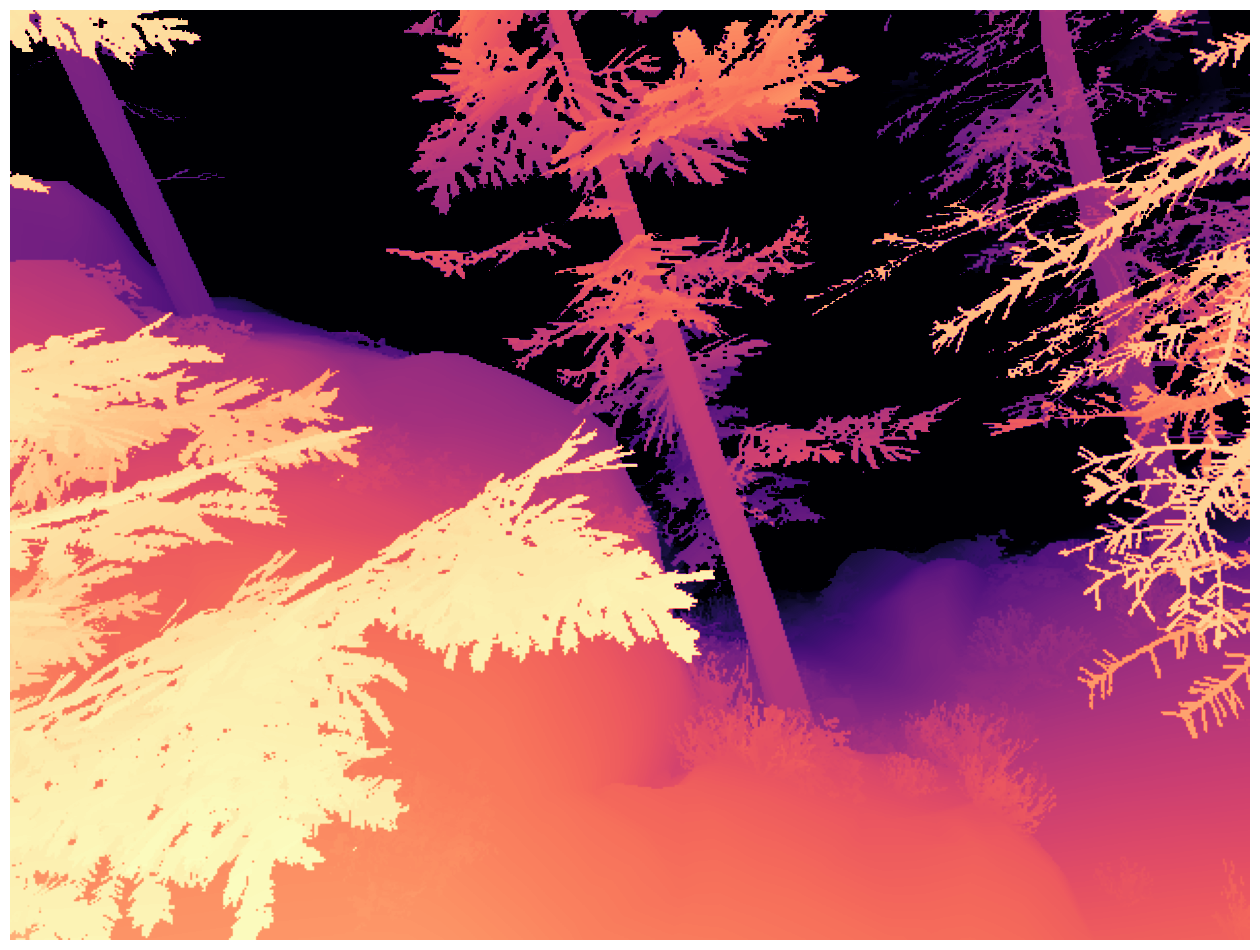}         \\
    \end{tabular}
    \vspace{-0.8em}
    \caption{\textbf{TartanAir Qualitatives.} TartanAir provides a wide range of complex environments, we provide a few examples along with the predictions by \netname.}
    \label{fig:tartanair-qualitatives}
    \vspace{-1em}
\end{figure}

\begin{figure*}[t]
    \centering
    \begin{tabular}{@{}c@{}c@{}}
    \includegraphics[width=0.44\textwidth]{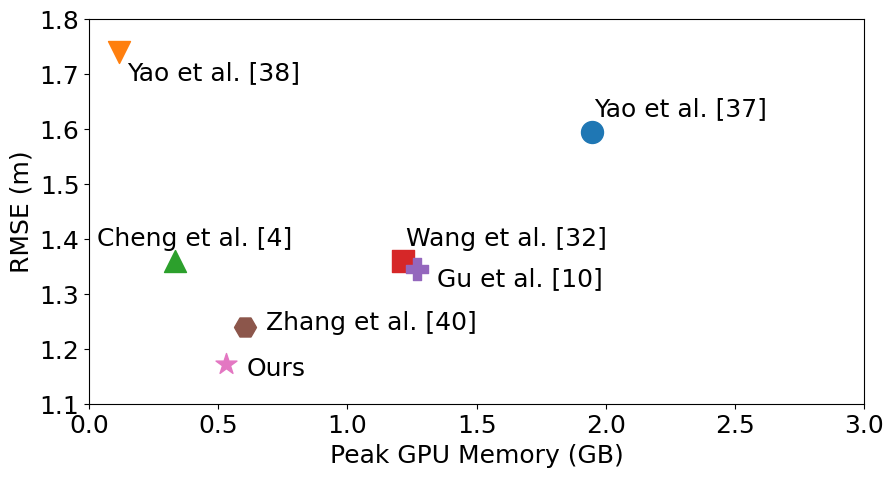}
    \includegraphics[width=0.43\textwidth]{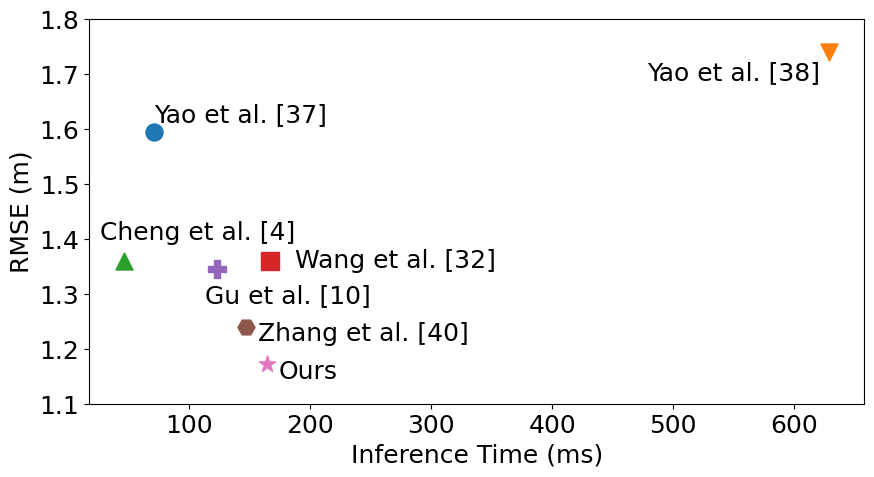}
    \end{tabular}
    \vspace{-0.8em}
    \caption{\textbf{Benchmark on Memory and Time Requirements.} We test each model in evaluation mode on a single NVIDIA RTX 3090 in 32FP precision, with input size $768 \times 576$ and 5 input views. We measure peak memory as the minimum memory needed to run a model in evaluation, time in milliseconds and RMSE on Blended \cite{yao2020blendedmvs}.} 
    \label{fig:memory-time}
    \vspace{-1em}
\end{figure*}

\textbf{TartanAir Benchmark.}
The TartanAir dataset \cite{tartanair2020iros} is a large synthetic dataset composed of a wide spectrum of indoor, outdoor, aerial, and underwater scenarios recorded by a monocular camera, with different moving patterns of variable toughness. It also contains a few moving objects like fishes, steam, and industrial machines as well as high-frequency details like tree leaves. In this scenario, the depth range of each single view is hard to define since it can embrace hundreds of meters in a landscape view or a few meters when the camera moves around a wall, and this can happen within the same scene as well. Thus, this environment is a perfect benchmark for \netname. In Table~\ref{tab:tartanair-benchmark} we show the performance of our approach and existing multi-view methods, where each competitor is fed with the depth range from the ground-truth depth. Even though this is unfair to our approach, not knowing anything about the prediction range, we still exhibit the best performance. We show a few qualitative examples in Fig. \ref{fig:tartanair-qualitatives}.

\begin{table}[t]
\centering
\resizebox{\linewidth}{!}{
\begin{tabular}{@{}l|cccc|ccc@{}}
    \hline \hline
    \multirow{2}*{Method}                       & \multicolumn{4}{|c|}{2D Metrics}                                                         & \multicolumn{3}{c}{3D Metrics}             \\
                                                \ignore{& mae (mm)   & rmse (mm)}& $>$1 mm     & $>$2 mm     & $>$3 mm      & $>$4 mm      & acc.        & compl.       & avg           \\
    \hline
    \mvsnet                                              \ignore{& 24.07      & 52.08     }& 0.5550      & 0.3400      & 0.2680       & 0.2370       & 0.6350      & 0.3040       & 0.4695        \\
    \dhcnet                                              \ignore{& 16.90      & \bt{32.13}}& 0.6300      & 0.4230      & 0.3290       & 0.2830       & 0.6620      & 0.3420       & 0.5020        \\
    \ucsnet                                              \ignore{& 25.70      & 52.45     }& 0.5060      & 0.3320      & 0.2770       & 0.2540       & 0.5510      & 0.2720       & 0.4115        \\
    \patchnet                                            \ignore{& 18.29      & 41.38     }& 0.4750      & 0.3100      & 0.2600       & 0.2360       & 0.4610      & 0.2980       & 0.3795        \\
    \casnet                                              \ignore{& 25.53      & 56.23     }& 0.4800      & 0.3070      & 0.2570       & 0.2330       & 0.5280      & 0.2620       & 0.3950        \\
    \cermvs                                              \ignore{& \bt{9.834} & 131.26    }& 0.4126      & 0.2556      & \bt{0.2029}  & \bt{0.1770} & 0.4966      & \bt{0.2581}  & 0.3773        \\
    \netname{} (ours)                                    \ignore{& 14.78      & 33.48     }& \bt{0.3683} & \bt{0.2439} & 0.2063       & 0.1884       & \bt{0.4466} & 0.2775       & \bt{0.3620}   \\
    \hline \hline
\end{tabular}
}
\vspace{-0.8em}
\caption{\textbf{DTU Benchmark.} Results achieved by other multi-view frameworks and ours on DTU. Even though other methods are advantaged by the fixed depth range of this dataset our method is still comparable in performance.}
\label{tab:dtu-benchmark}
\vspace{-1.3em}
\end{table}

\textbf{DTU Benchmark.}
DTU \cite{jensen2014large} is a dataset composed of small objects whose 3D structure is captured by means of a robotic arm and a structured light sensor. Due to these specifics, it exhibits a really small and fixed depth range. In this context, methods relying on the scene depth range are advantaged since they can make use of robust and precise information which limits outliers, especially in texture-less areas. We pretrain on \cite{yao2020blendedmvs} following \cite{poggi2022guided}. In Table \ref{tab:dtu-benchmark} we show both 2D depth metrics and standard 3D metrics obtained with the same reconstruction pipeline from \cite{poggi2022guided} on \cite{jensen2014large}. Our approach is still competitive in both 3D point cloud reconstruction and depth estimation, despite being disadvantaged in this context. We provide examples of reconstructed point clouds in the supplementary material.

\textbf{Ablation study.} We provide a simple ablation study about neighborhood sampling and depth decoding components, shown in Table \ref{tab:ablations}. We perform such an experiment with a slightly smaller number of training steps on Blended \cite{yao2020blendedmvs} with respect to our final tuned model, thus we report also the results of our final model for a better comparison. \netname{} greatly benefits from both of these modules.

\textbf{Memory and Time Analysis.}
Finally, we provide an analysis of the time and memory requirements of our method, compared with existing approaches in Fig. \ref{fig:memory-time}. We measure peak memory usage, runtime, and RMSE error using $5$ input views of size $768 \times 576$, on a single NVIDIA RTX 3090. The choice to measure peak memory is justified by the fact that this latter is the minimum memory required when deploying these models in a real application and thus we believe it is the most significant metric in this sense. In Fig. \ref{fig:memory-time} we can clearly observe that despite being neither the fastest nor the lighter approach, our proposal provides a good balance in memory usage and inference time, while still being the best one in performance.

\begin{table}[t]
    \centering
    \resizebox{\linewidth}{!}{
    \begin{tabular}{@{}l|cc|cc@{}}
     \hline \hline
                         & Convex            & Deformable      & \multirow{2}*{MAE} & \multirow{2}*{$>$1 m}\\
                         & Upsampling        & Sampling        &                    &                    \\
     \hline               
     Baseline            &                   &                  & 0.4673            &  0.1085            \\
     Baseline + Deform.  &                   & \checkmark       & 0.4525            &  0.1046            \\
     Baseline + Convex   & \checkmark        &                  & 0.3406            &  0.0756            \\
     Full                & \checkmark        & \checkmark       & \bt{0.3197}       &  \bt{0.0695}        \\
     \hline
     Full (Tuned)        & \checkmark        & \checkmark       & 0.2982            &  0.0645            \\
     \hline \hline
    \end{tabular}
    }
    \vspace{-0.8em}
    \caption{\textbf{Ablation study on \netname.} We assess the impact of convex upsampling and deformable sampling modules on Blended \cite{yao2020blendedmvs}. Each ablation has been performed with the same number of training steps, smaller than the total used to train our final model (Tuned). When convex upsampling is not applied we use bilinear upsampling instead.}
    \label{tab:ablations}
    \vspace{-1em}
\end{table}

\section{Conclusion}

In this paper, we have presented \netname, a novel framework for multi-view depth estimation completely independent from scene depth range assumptions. We have demonstrated its applicability to different environments like monocular posed videos characterized by multiple views with small baseline distances, stereo cameras, and multi-view cameras with large unconstrained baseline values. We have studied the implications of our approach, highlighting its capability to introspect on view importance in correlation matching. This latter feature softens the deploying issues of multi-view frameworks, allowing for identifying less meaningful views and reducing inference time and memory requirements. However, future research may identify significantly more effective approaches for this latter purpose.

\textbf{Acknowledgment.} We gratefully acknowledge Sony
Depthsensing Solutions SA/NV for funding this research.

{\small
\bibliographystyle{ieeenat_fullname}
\bibliography{egbib}
}

\end{document}


\maketitle

This manuscript provides additional insights about our paper ``Range-Agnostic Multi-View Depth Estimation with Keyframe Selection".
We collect here additional qualitative and experimental material about our multi-view depth estimation proposal. Moreover, we provide details about the network architecture and training procedure, as well as a qualitative study of our keyframe ranking approach.

\section{Qualitative Results on Blended}

\begin{figure*}[h]
    \centering
    \renewcommand{\tabcolsep}{2pt}
    \begin{tabular}{@{}cccccc@{}}
    \includegraphics[width=0.15\linewidth]{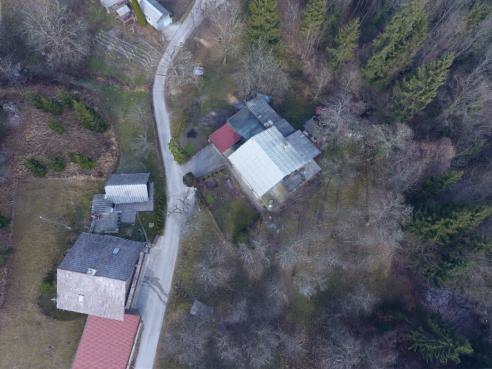} &
    \includegraphics[width=0.15\linewidth]{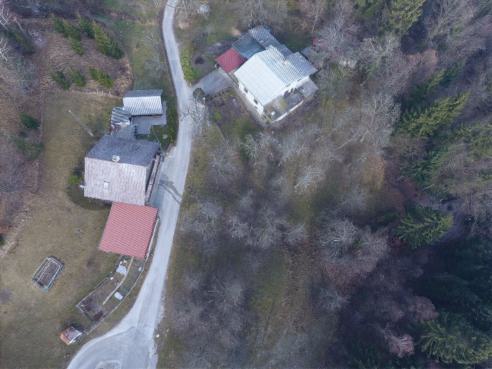} &
    \includegraphics[width=0.15\linewidth]{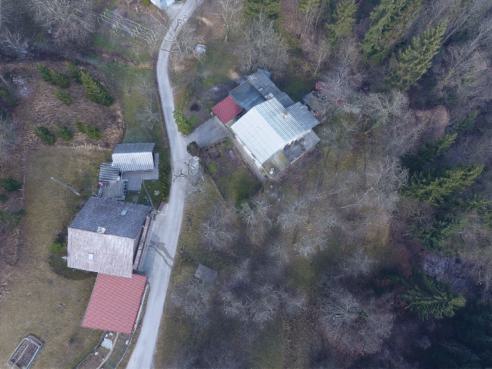} &
    \includegraphics[width=0.15\linewidth]{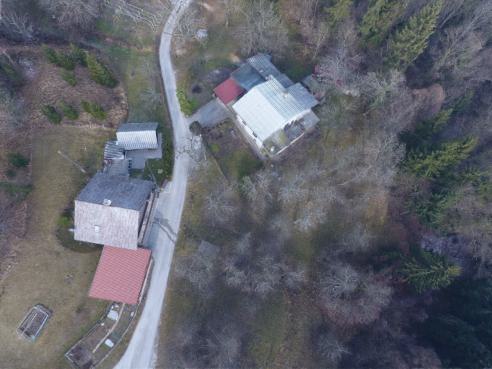} &
    \includegraphics[width=0.15\linewidth]{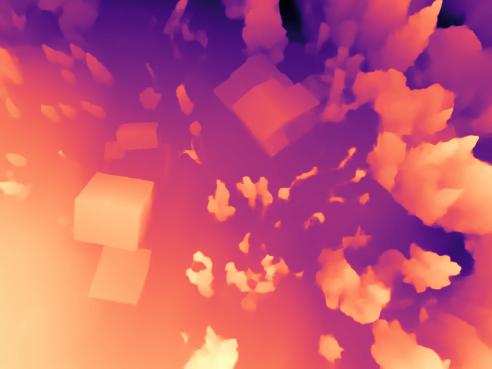}  &
    \includegraphics[width=0.15\linewidth]{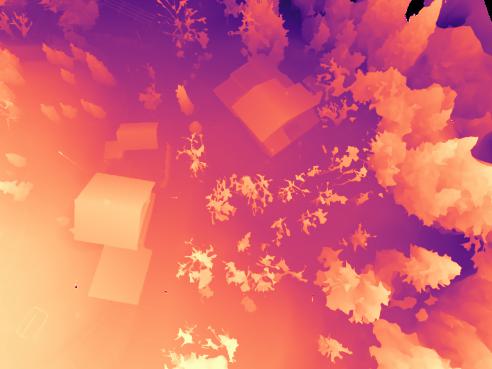}   \\
    \includegraphics[width=0.15\linewidth]{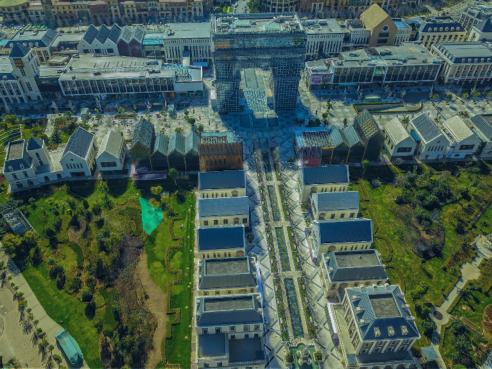} &
    \includegraphics[width=0.15\linewidth]{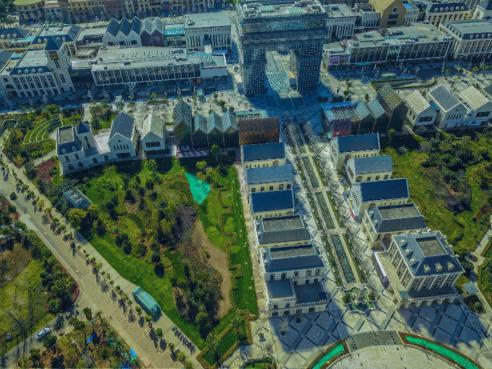} &
    \includegraphics[width=0.15\linewidth]{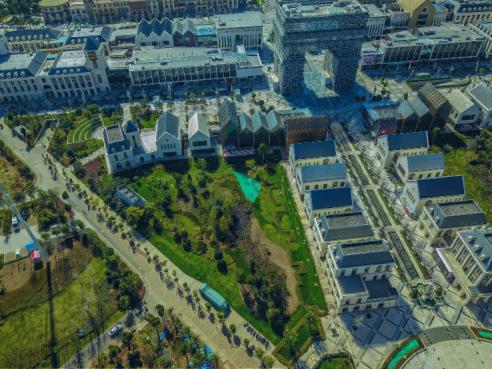} &
    \includegraphics[width=0.15\linewidth]{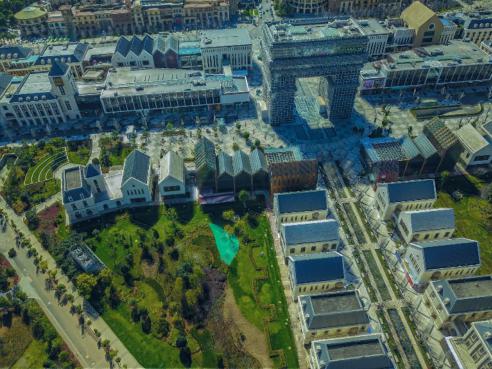} &
    \includegraphics[width=0.15\linewidth]{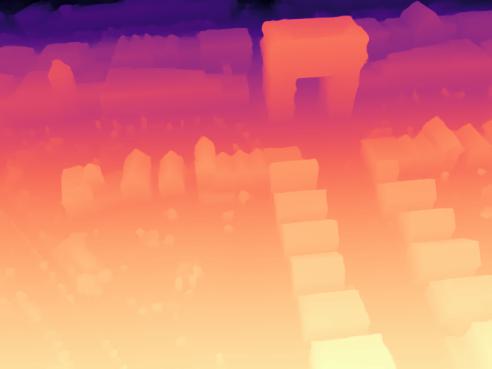}  &
    \includegraphics[width=0.15\linewidth]{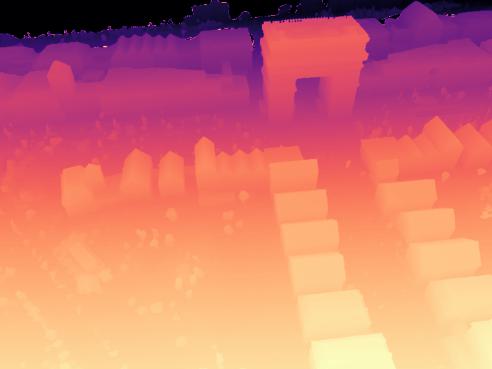}   \\
    \includegraphics[width=0.15\linewidth]{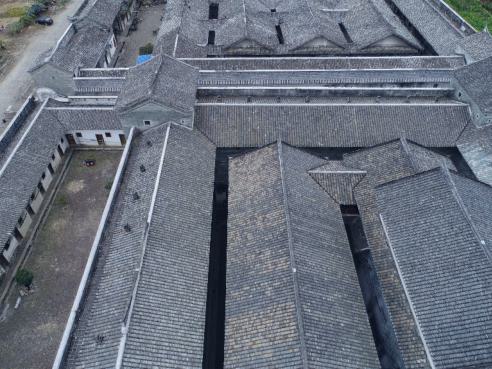} &
    \includegraphics[width=0.15\linewidth]{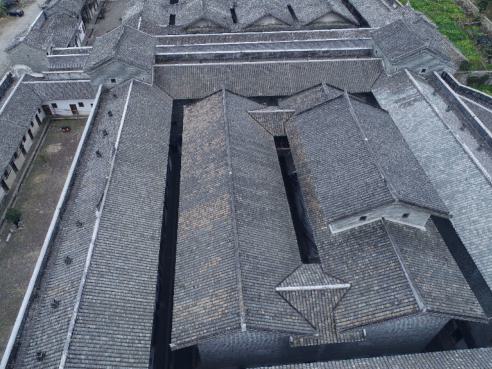} &
    \includegraphics[width=0.15\linewidth]{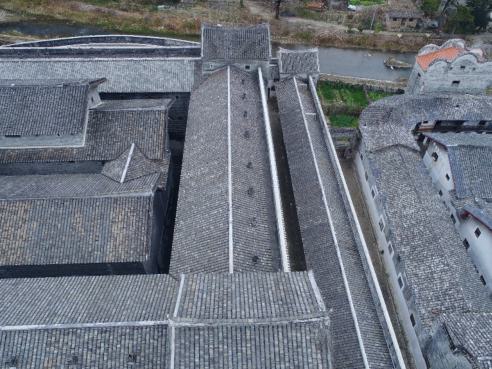} &
    \includegraphics[width=0.15\linewidth]{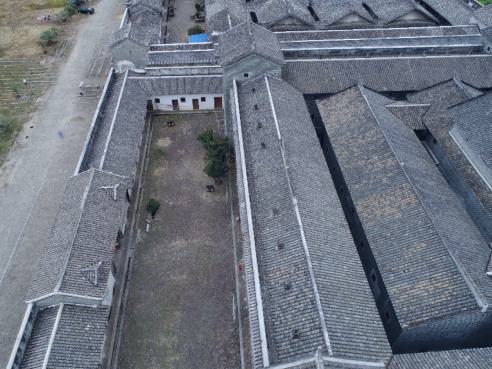} &
    \includegraphics[width=0.15\linewidth]{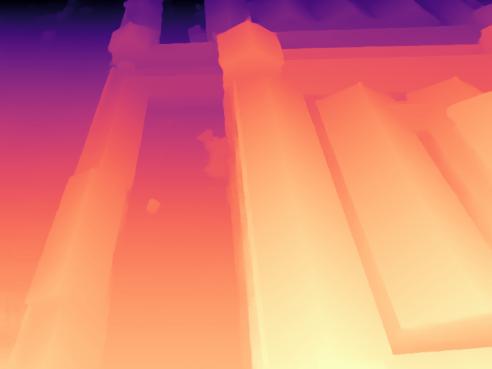}  &
    \includegraphics[width=0.15\linewidth]{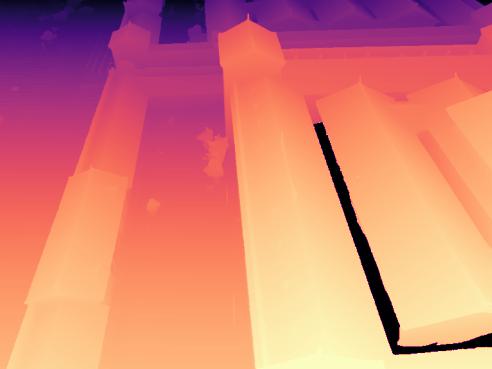}   \\
    \includegraphics[width=0.15\linewidth]{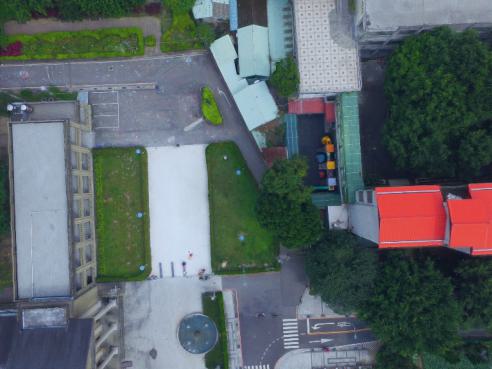} &
    \includegraphics[width=0.15\linewidth]{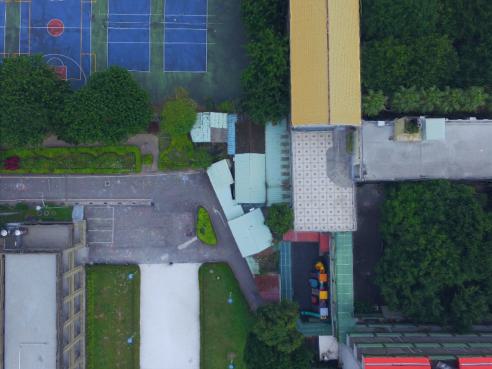} &
    \includegraphics[width=0.15\linewidth]{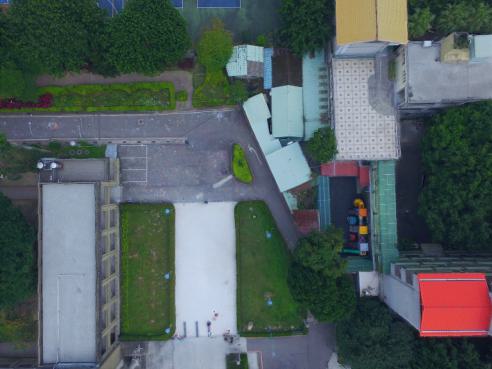} &
    \includegraphics[width=0.15\linewidth]{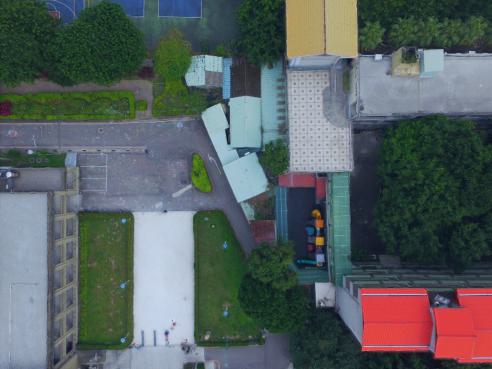} &
    \includegraphics[width=0.15\linewidth]{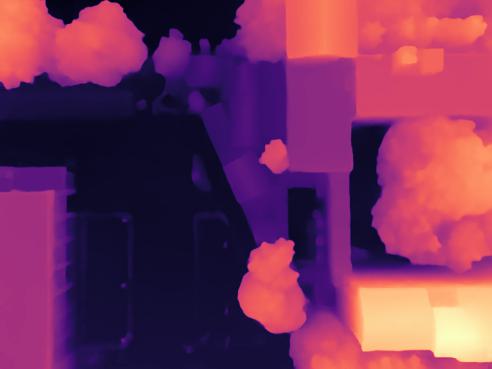}  &
    \includegraphics[width=0.15\linewidth]{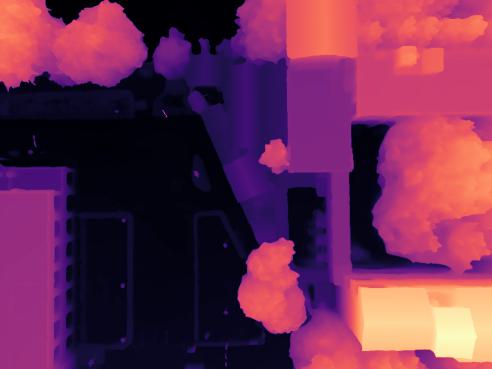}   \\
    \includegraphics[width=0.15\linewidth]{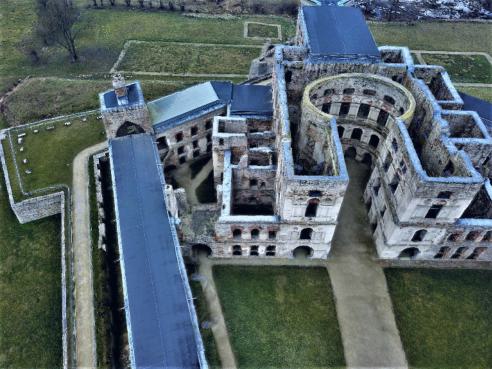} &
    \includegraphics[width=0.15\linewidth]{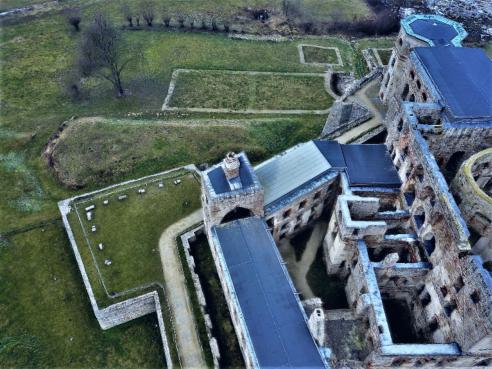} &
    \includegraphics[width=0.15\linewidth]{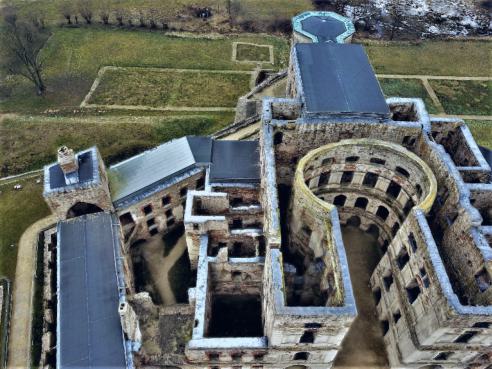} &
    \includegraphics[width=0.15\linewidth]{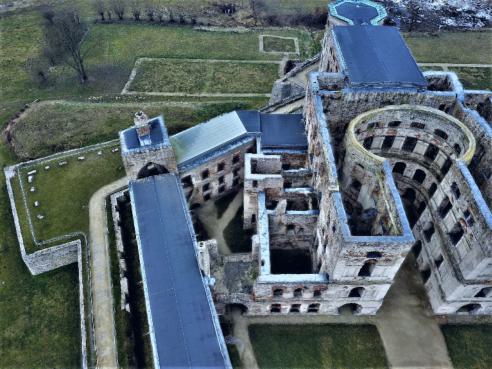} &
    \includegraphics[width=0.15\linewidth]{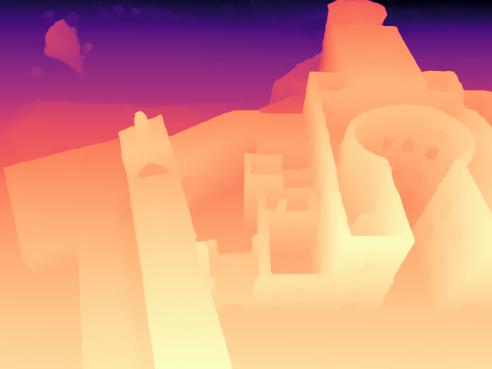}  &
    \includegraphics[width=0.15\linewidth]{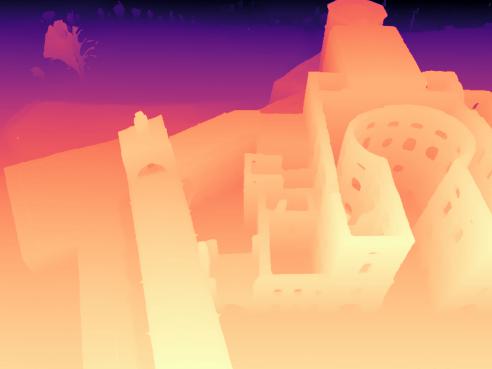}   \\
    \scriptsize \textbf{Source 1}       &
    \scriptsize \textbf{Source 2}       &
    \scriptsize \textbf{Source 3}       &
    \scriptsize \textbf{Reference}      &
    \scriptsize \textbf{Prediction}     &
    \scriptsize \textbf{Ground Truth}
    \end{tabular}
    \vspace{0.1em}
    \caption{\textbf{Qualitative results on Blended.} Predictions obtained by using 5 views as input (only 4 are showed for representative purpose).}
    \label{fig:blended-qualitatives}
\end{figure*}

We report a few sample scenes from Blended to show the network capability to extract fine details. In Figure \ref{fig:blended-qualitatives} we plot 4 out of 5 views provided to the network along with the prediction and the ground-truth (dark black represents missing values in the ground-truth).

\clearpage

\section{Qualitative Results on UnrealStereo4K}

\begin{figure*}[!h]
    \centering
    \renewcommand{\tabcolsep}{2pt}
    \begin{tabular}{@{}cccc@{}}
    \includegraphics[width=0.24\linewidth]{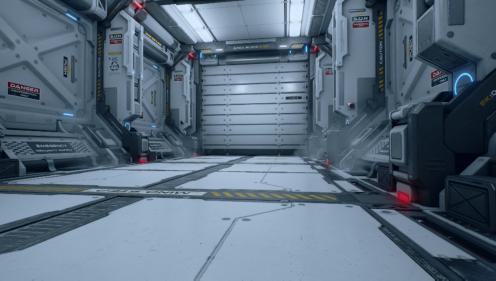}  &
    \includegraphics[width=0.24\linewidth]{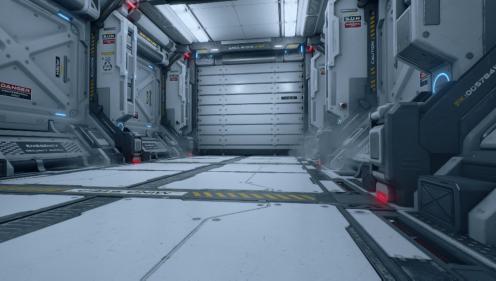} &
    \includegraphics[width=0.24\linewidth]{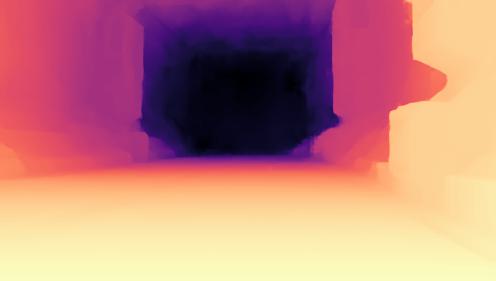}      &
    \includegraphics[width=0.24\linewidth]{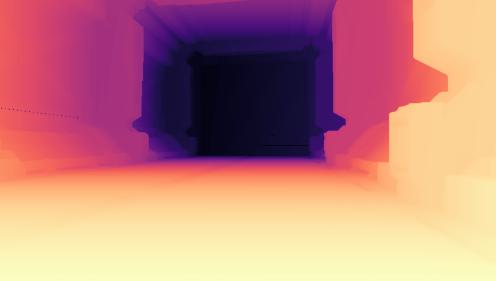}        \\
    \includegraphics[width=0.24\linewidth]{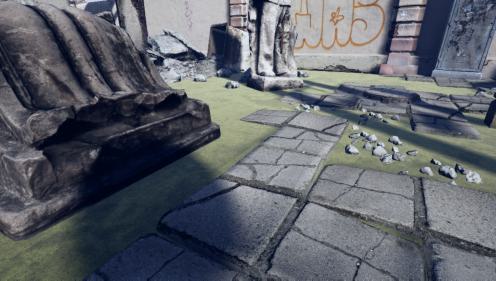}  &
    \includegraphics[width=0.24\linewidth]{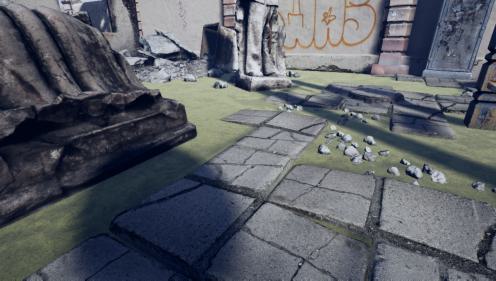} &
    \includegraphics[width=0.24\linewidth]{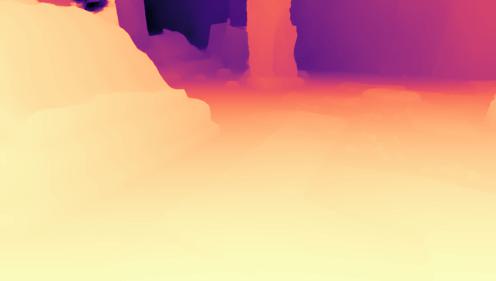}      &
    \includegraphics[width=0.24\linewidth]{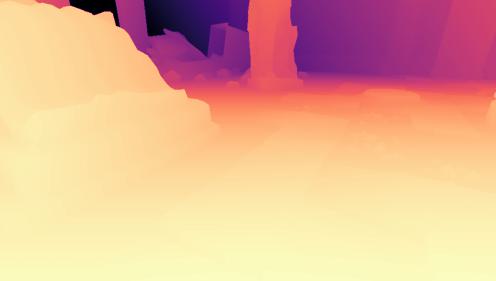}        \\
    \includegraphics[width=0.24\linewidth]{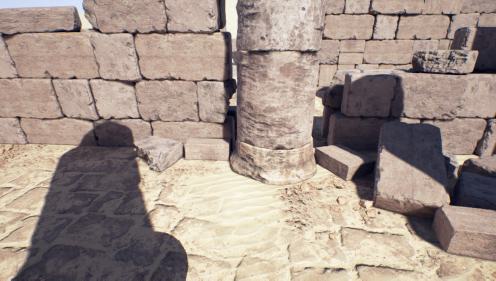}  &
    \includegraphics[width=0.24\linewidth]{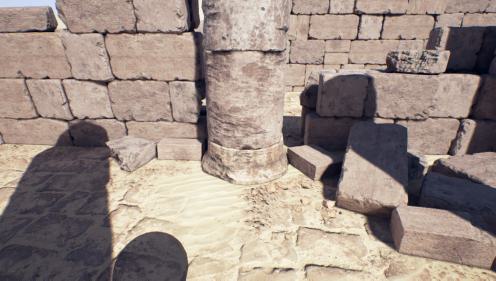} &
    \includegraphics[width=0.24\linewidth]{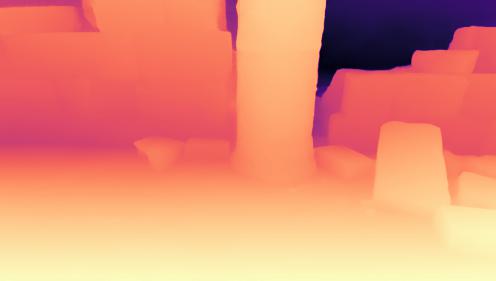}      &
    \includegraphics[width=0.24\linewidth]{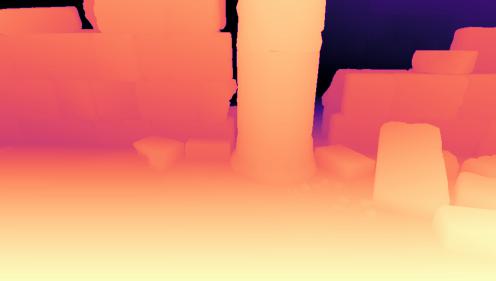}        \\
    \includegraphics[width=0.24\linewidth]{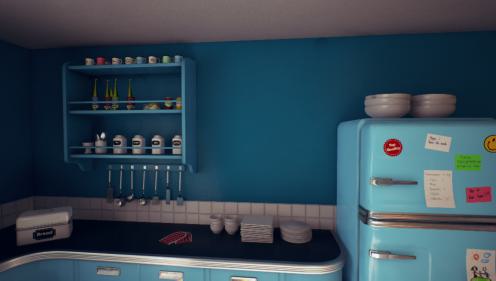}  &
    \includegraphics[width=0.24\linewidth]{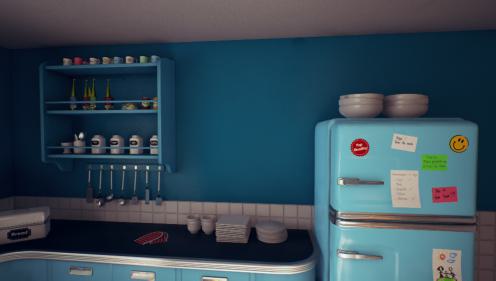} &
    \includegraphics[width=0.24\linewidth]{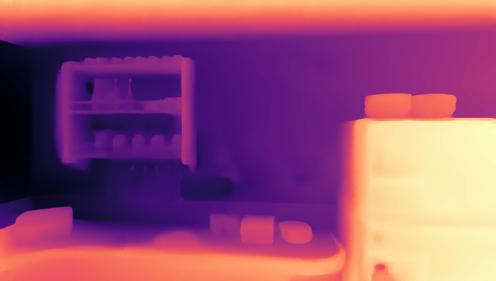}      &
    \includegraphics[width=0.24\linewidth]{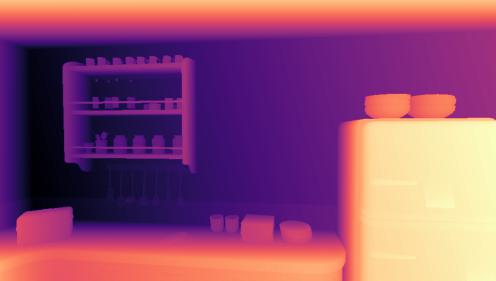}        \\
    \includegraphics[width=0.24\linewidth]{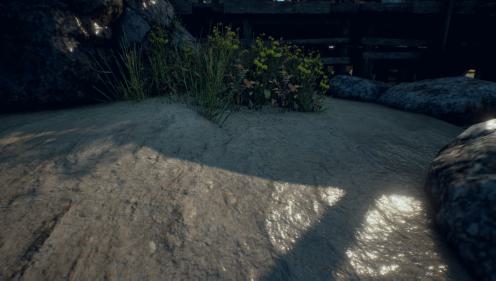}  &
    \includegraphics[width=0.24\linewidth]{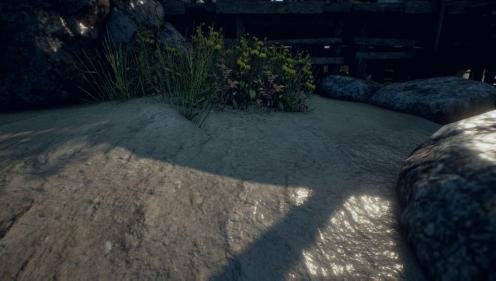} &
    \includegraphics[width=0.24\linewidth]{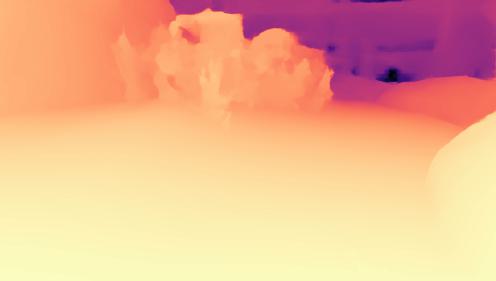}      &
    \includegraphics[width=0.24\linewidth]{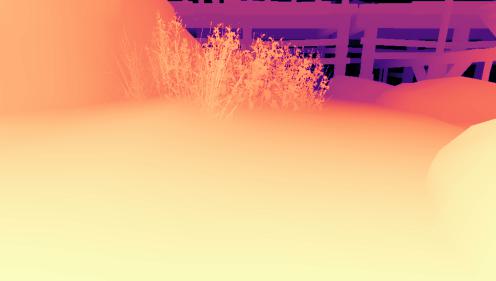}        \\
    \includegraphics[width=0.24\linewidth]{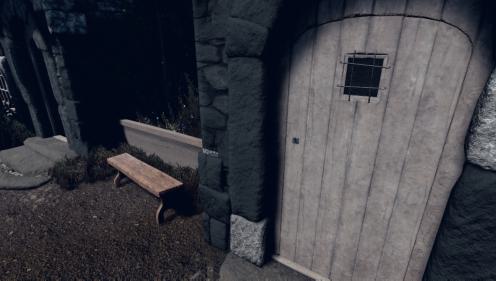}  &
    \includegraphics[width=0.24\linewidth]{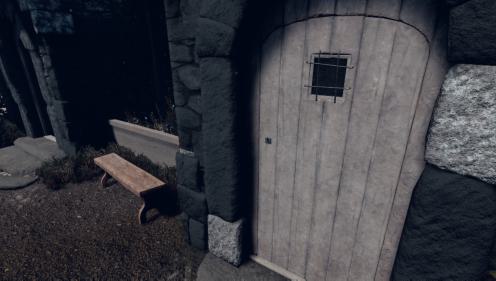} &
    \includegraphics[width=0.24\linewidth]{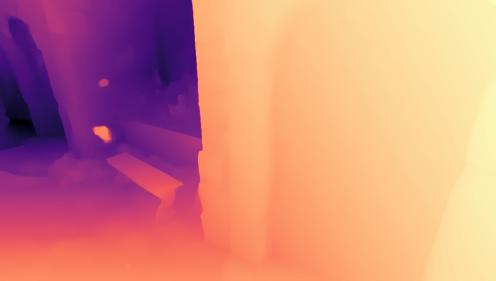}      &
    \includegraphics[width=0.24\linewidth]{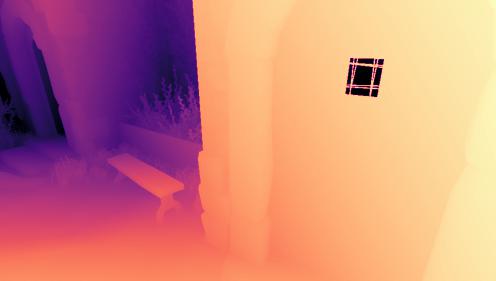}        \\
    \scriptsize \textbf{Left Image}   &
    \scriptsize \textbf{Right Image}  &
    \scriptsize \textbf{Prediction}   &
    \scriptsize \textbf{Ground Truth}
    \end{tabular}
    \vspace{0.1em}
    \caption{\textbf{Qualitative results on UnrealStereo4K Stereo.} We use the pre-trained model on Blended to show the capability of our method to generalize across datasets.}
    \label{fig:unrealstereo-stereo}
\end{figure*}

We evenly select a few samples from the available sequences of UnrealStereo4K and show the stereo pair, network prediction, and ground-truth in Figure \ref{fig:unrealstereo-stereo}. UnrealStereo4K is a very challenging dataset containing heterogeneous indoor and outdoor scenes. Our network is not fine-tuned on the dataset itself -- i.e., we use the model trained on Blended to assess the generalization capabilities of our approach.

\clearpage

\section{Qualitative Results on TartanAir}

\begin{figure*}[h]
    \centering
    \renewcommand{\tabcolsep}{2pt}
    \begin{tabular}{@{}cccccc@{}}
    \includegraphics[width=0.15\linewidth]{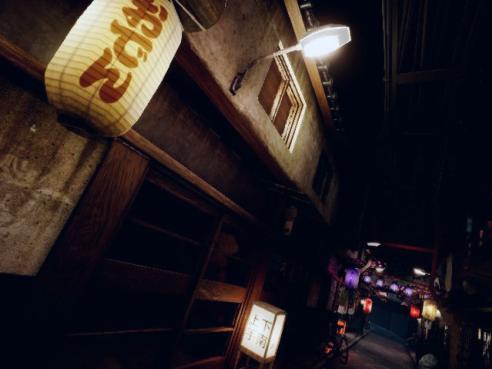} &
    \includegraphics[width=0.15\linewidth]{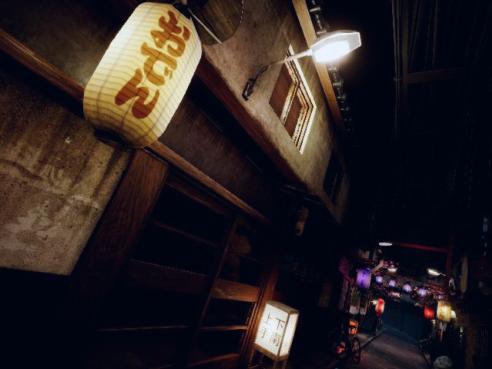} &
    \includegraphics[width=0.15\linewidth]{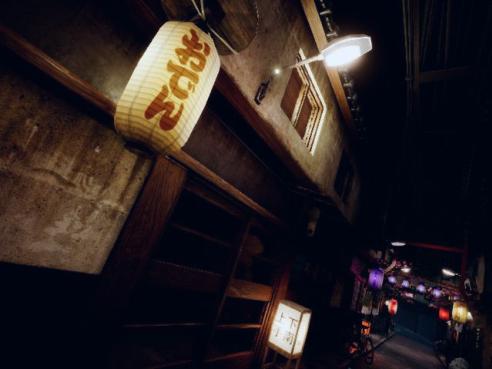} &
    \includegraphics[width=0.15\linewidth]{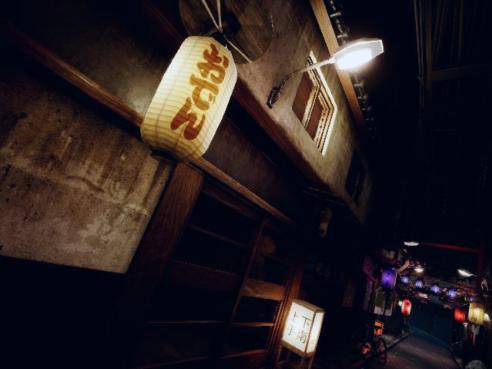} &
    \includegraphics[width=0.15\linewidth]{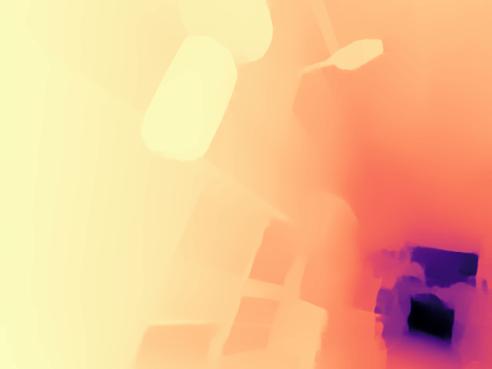}  &
    \includegraphics[width=0.15\linewidth]{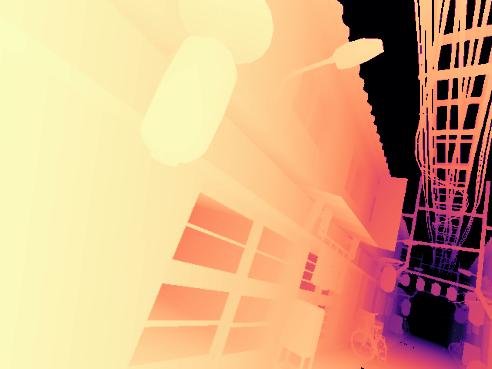}   \\
    \includegraphics[width=0.15\linewidth]{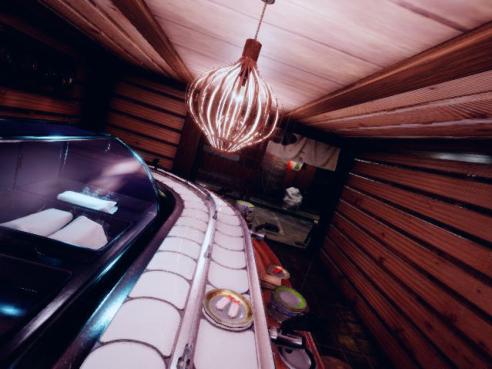} &
    \includegraphics[width=0.15\linewidth]{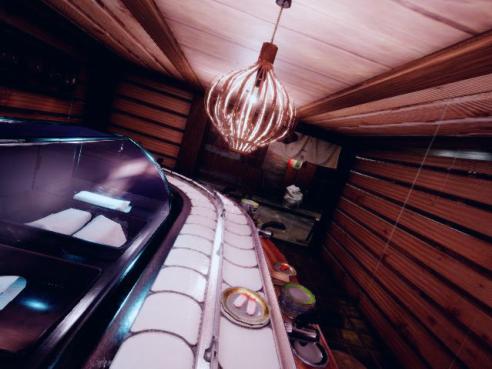} &
    \includegraphics[width=0.15\linewidth]{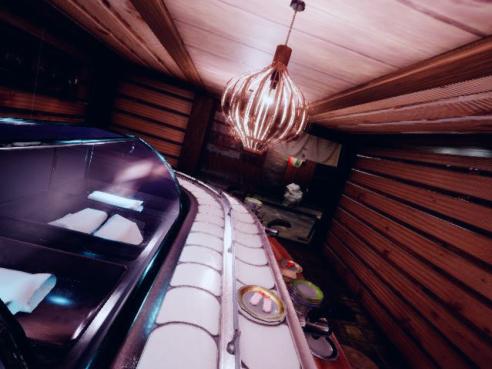} &
    \includegraphics[width=0.15\linewidth]{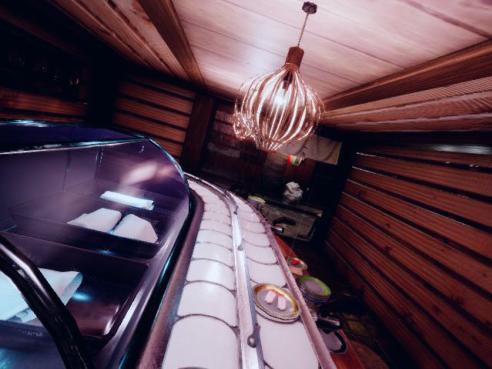} &
    \includegraphics[width=0.15\linewidth]{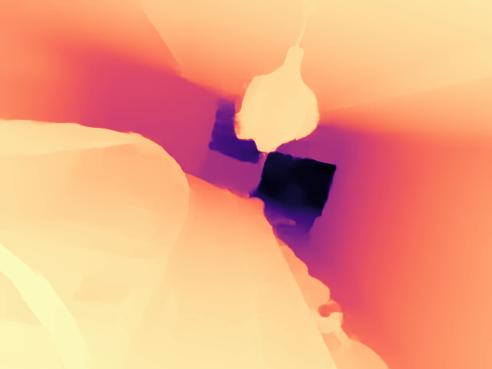}  &
    \includegraphics[width=0.15\linewidth]{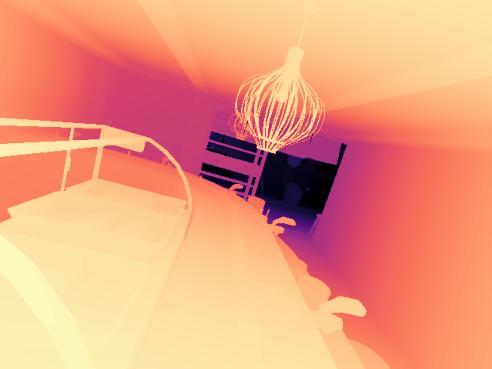}   \\
    \includegraphics[width=0.15\linewidth]{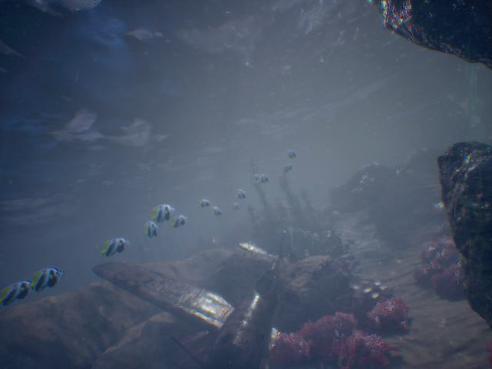} &
    \includegraphics[width=0.15\linewidth]{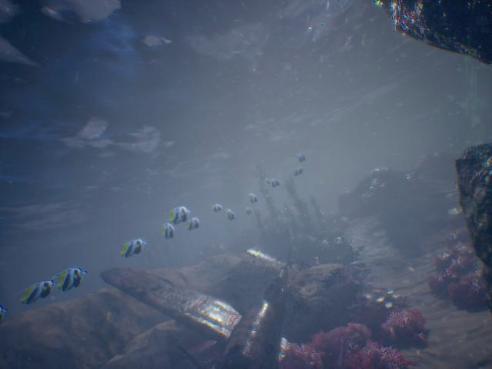} &
    \includegraphics[width=0.15\linewidth]{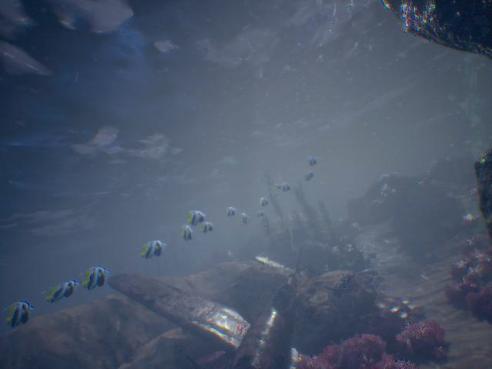} &
    \includegraphics[width=0.15\linewidth]{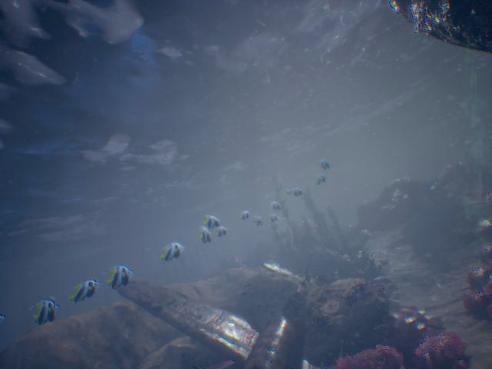} &
    \includegraphics[width=0.15\linewidth]{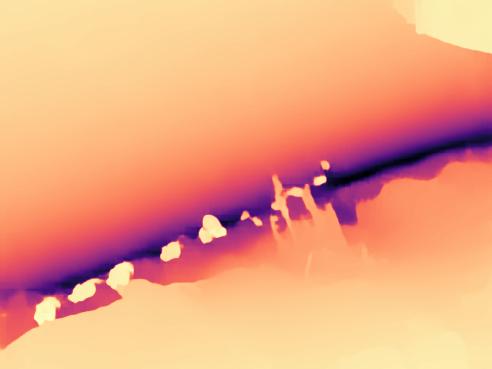}  &
    \includegraphics[width=0.15\linewidth]{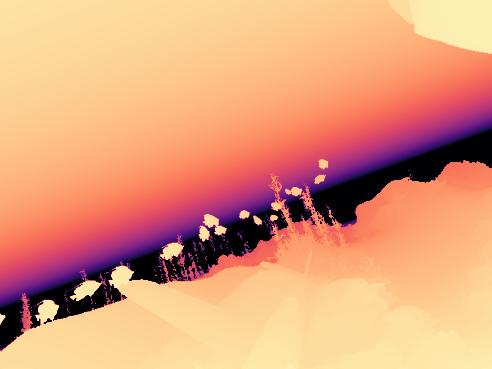}   \\
    \includegraphics[width=0.15\linewidth]{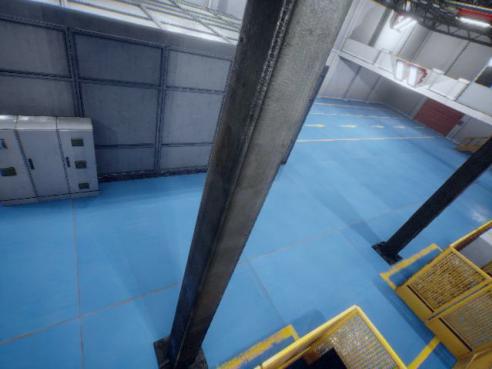} &
    \includegraphics[width=0.15\linewidth]{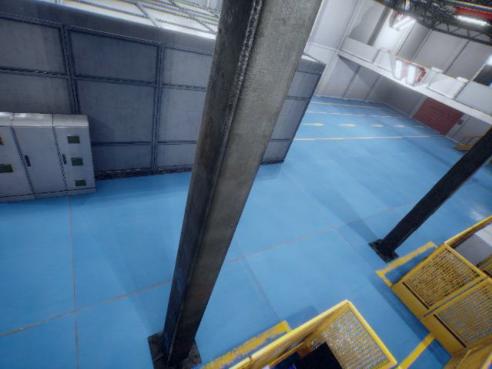} &
    \includegraphics[width=0.15\linewidth]{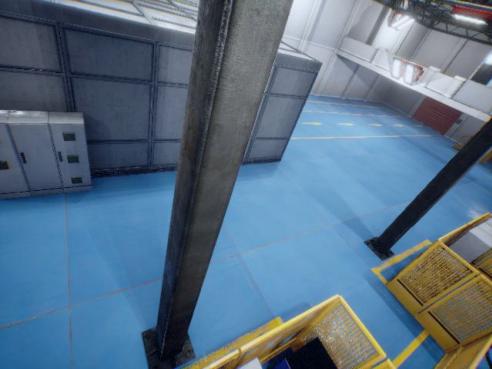} &
    \includegraphics[width=0.15\linewidth]{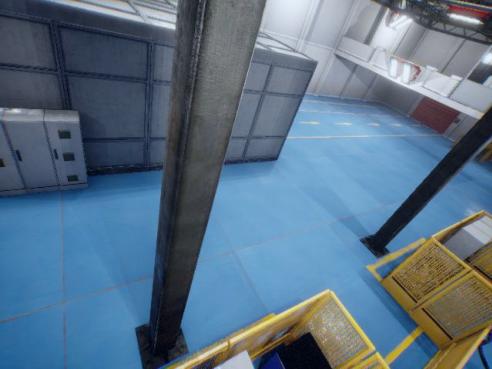} &
    \includegraphics[width=0.15\linewidth]{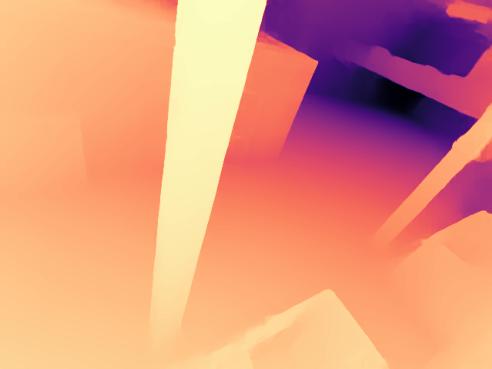}  &
    \includegraphics[width=0.15\linewidth]{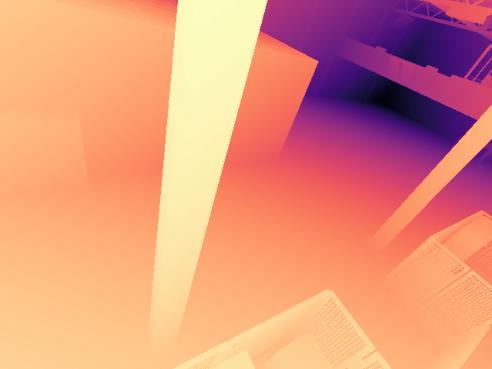}   \\
    \scriptsize \textbf{Source 1}       &
    \scriptsize \textbf{Source 2}       &
    \scriptsize \textbf{Source 3}       &
    \scriptsize \textbf{Reference}      &
    \scriptsize \textbf{Prediction}     &
    \scriptsize \textbf{Ground Truth}
    \end{tabular}
    \vspace{0.1em}
    \caption{\textbf{Qualitative results on TartanAir.} Predictions obtained by using 5 views as input (only 4 are showed for representative purpose).}
    \label{fig:tartanair-qualitatives}
\end{figure*}

We report a few sample scenes from TartanAir to show the network capability on this complex dataset. In Figure \ref{fig:tartanair-qualitatives} we plot 4 out of 5 views provided to the network along with the prediction and the ground-truth (dark black represents missing values in the ground-truth).

\section{Qualitative Results on DTU}

\begin{figure*}[h]
    \centering
    \begin{tabular}{@{}c@{}c@{}c@{}c@{}}
    \includegraphics[width=0.22\linewidth,trim={15cm 5cm 5cm  3cm},clip]{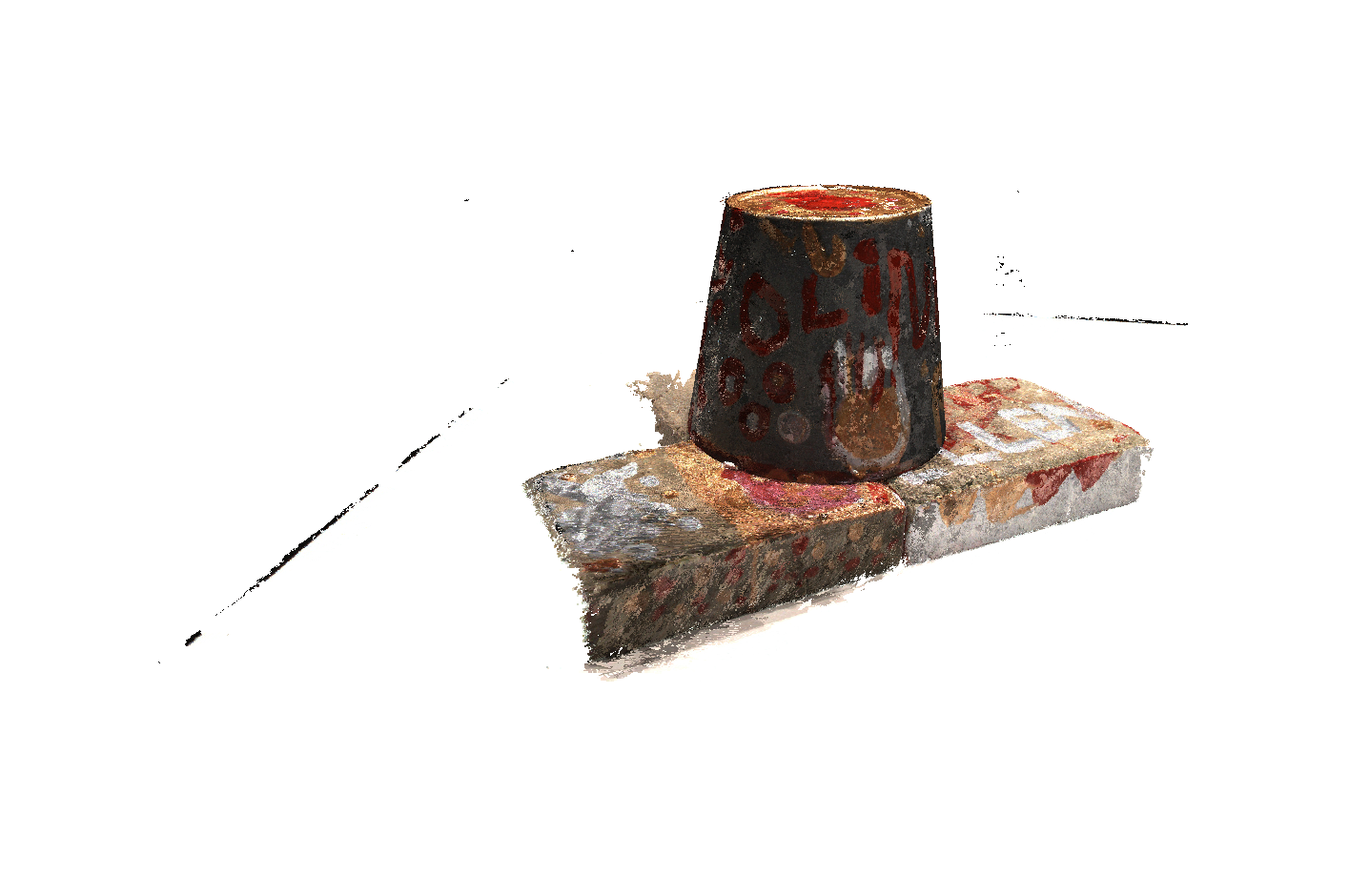}  &
    \includegraphics[width=0.22\linewidth,trim={15cm 5cm 5cm  3cm},clip]{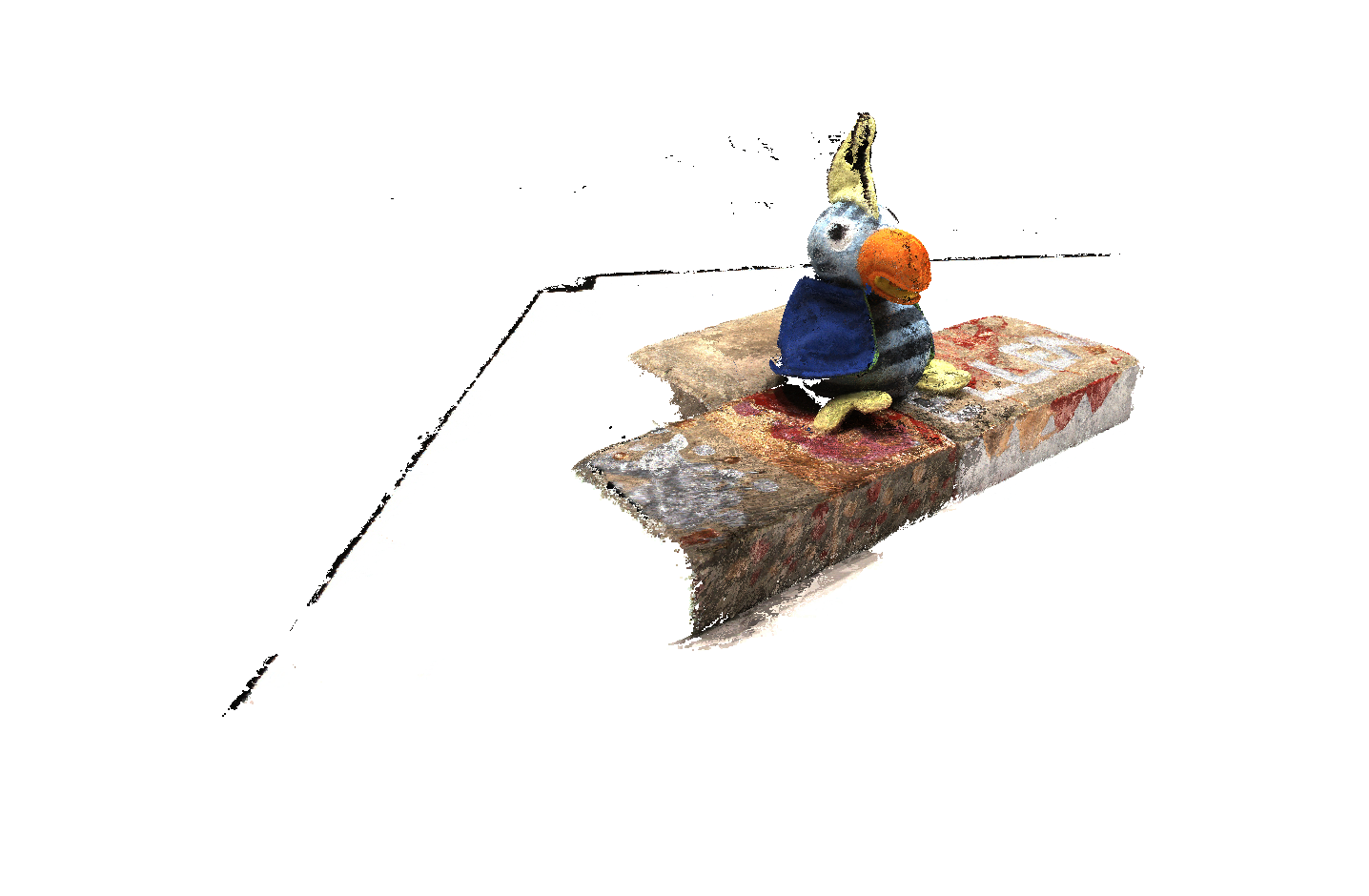}  &
    \includegraphics[width=0.22\linewidth,trim={13cm 5cm 7cm  3cm},clip]{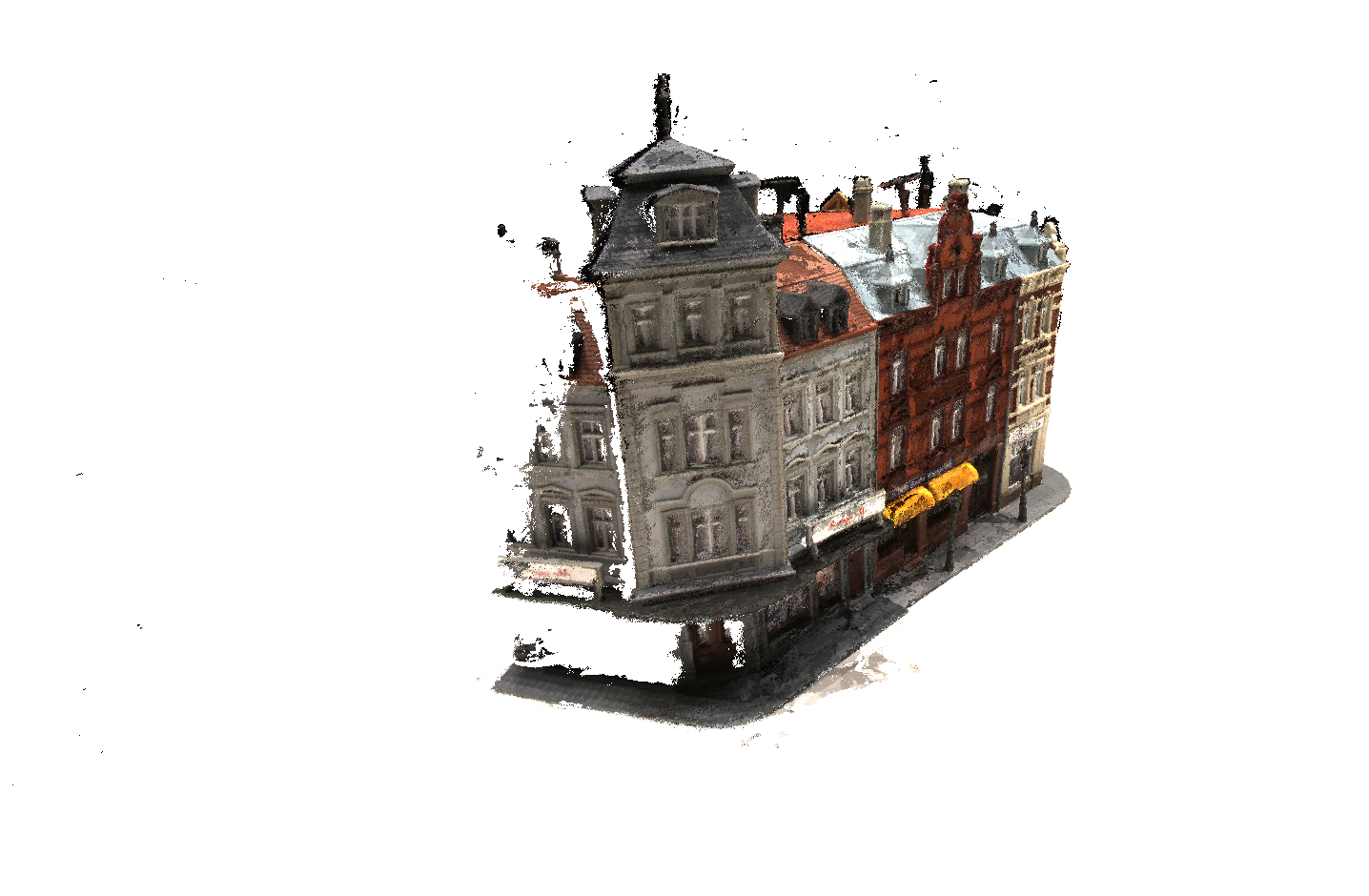}  &
    \includegraphics[width=0.22\linewidth,trim={10cm 5cm 10cm 3cm},clip]{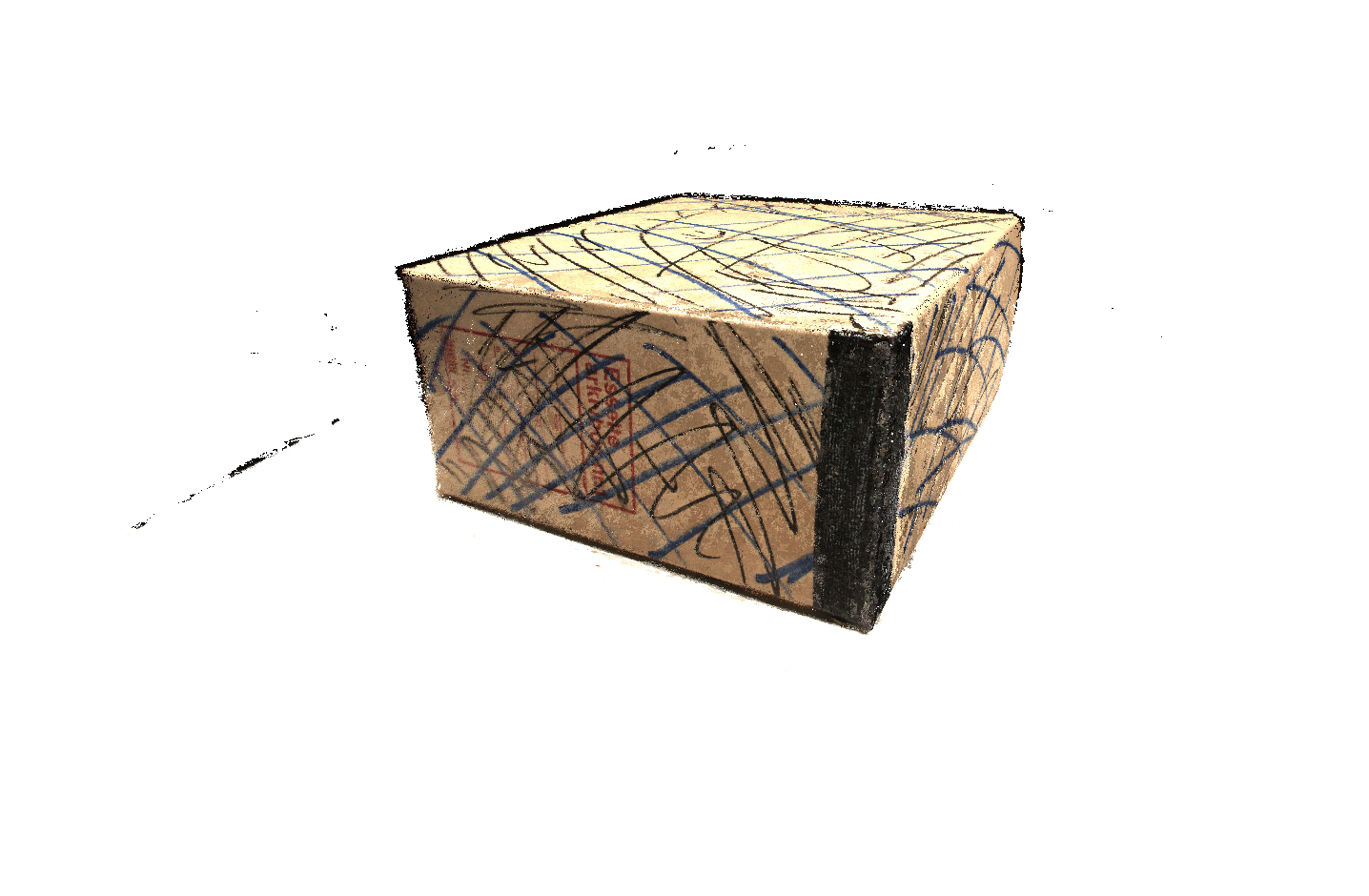} \\
    \includegraphics[width=0.22\linewidth,trim={12cm 5cm 8cm  3cm},clip]{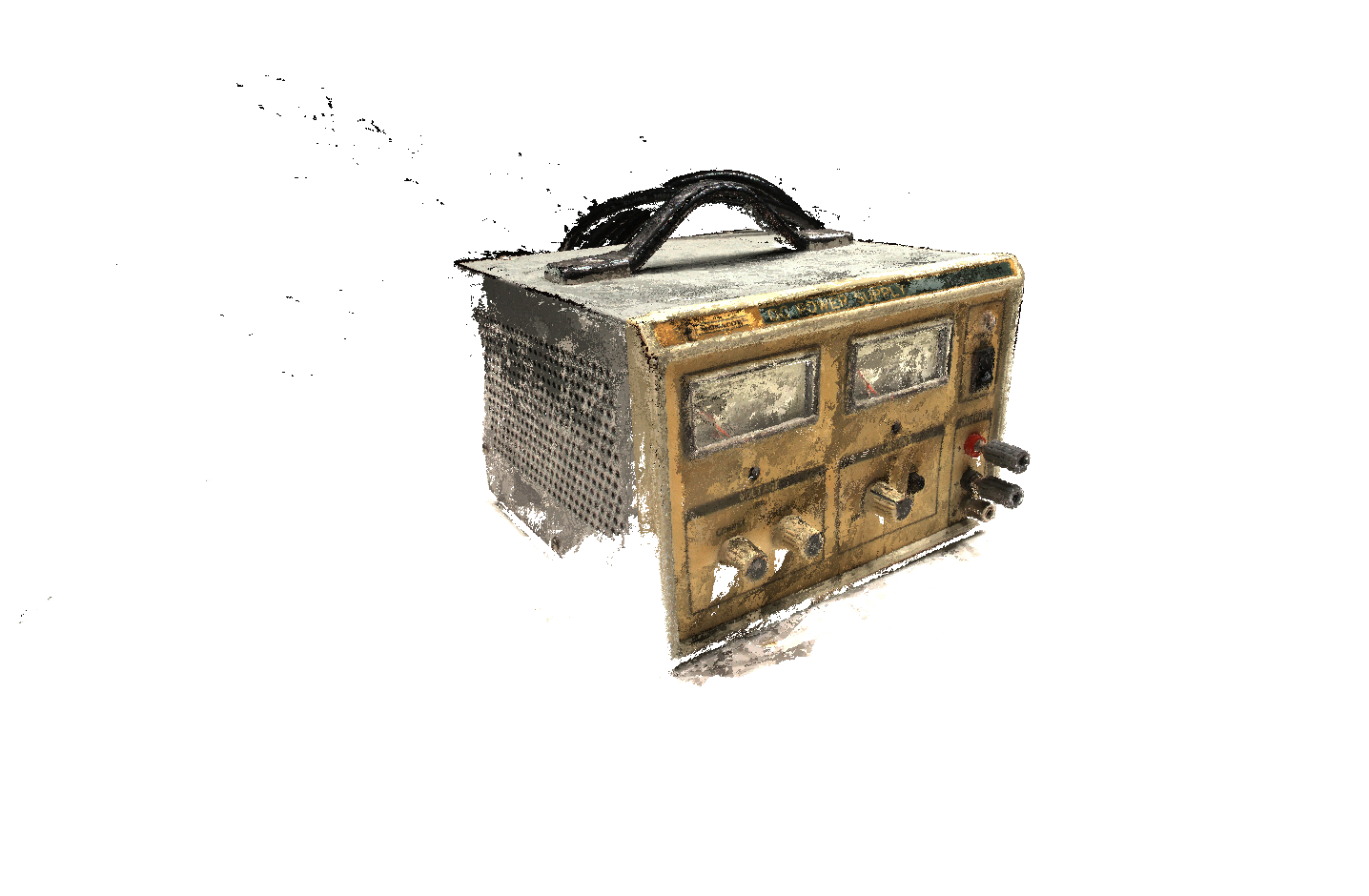}  &
    \includegraphics[width=0.22\linewidth,trim={10cm 5cm 8cm  3cm},clip]{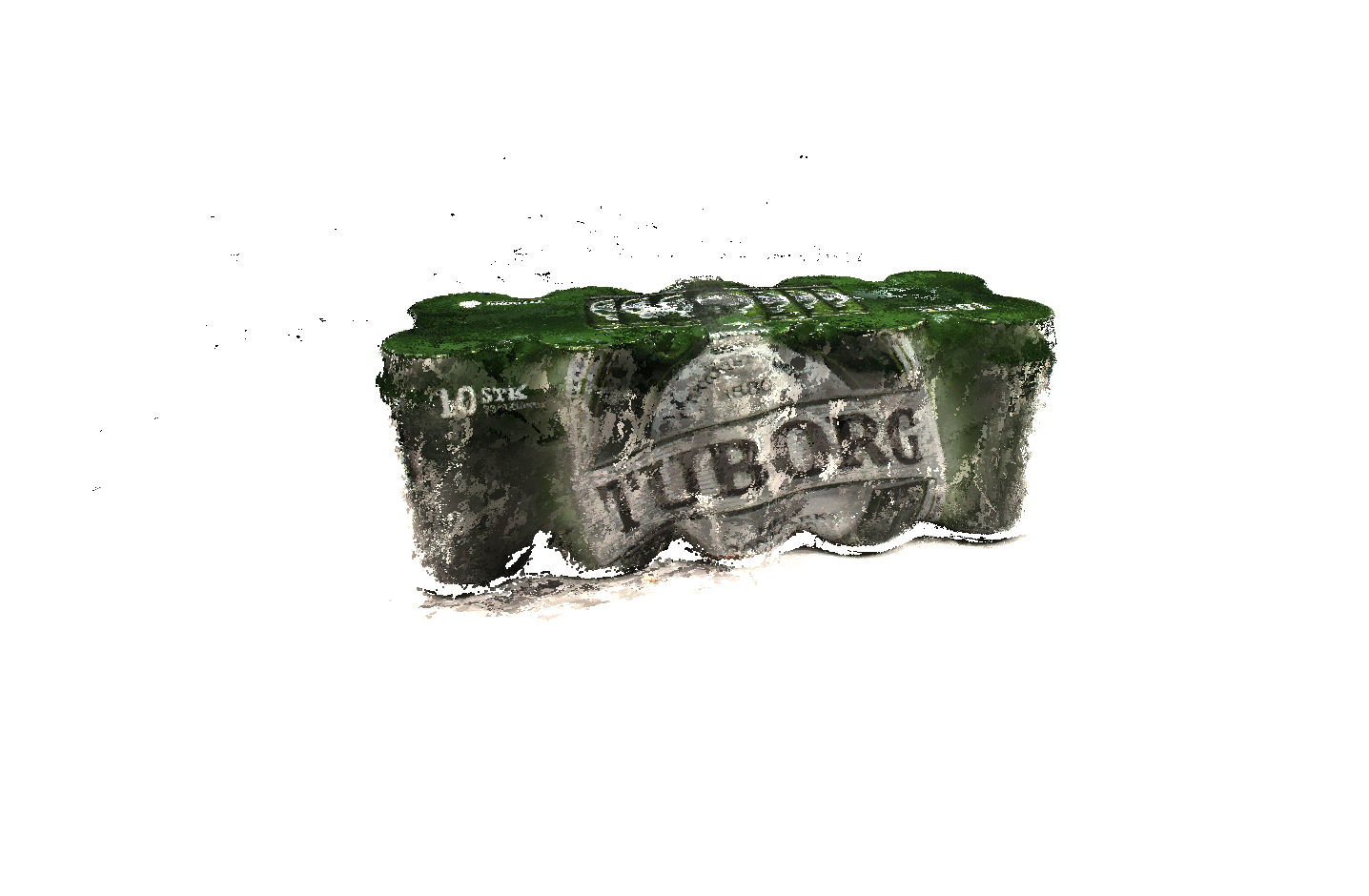}  &
    \includegraphics[width=0.22\linewidth,trim={10cm 5cm 6cm  3cm},clip]{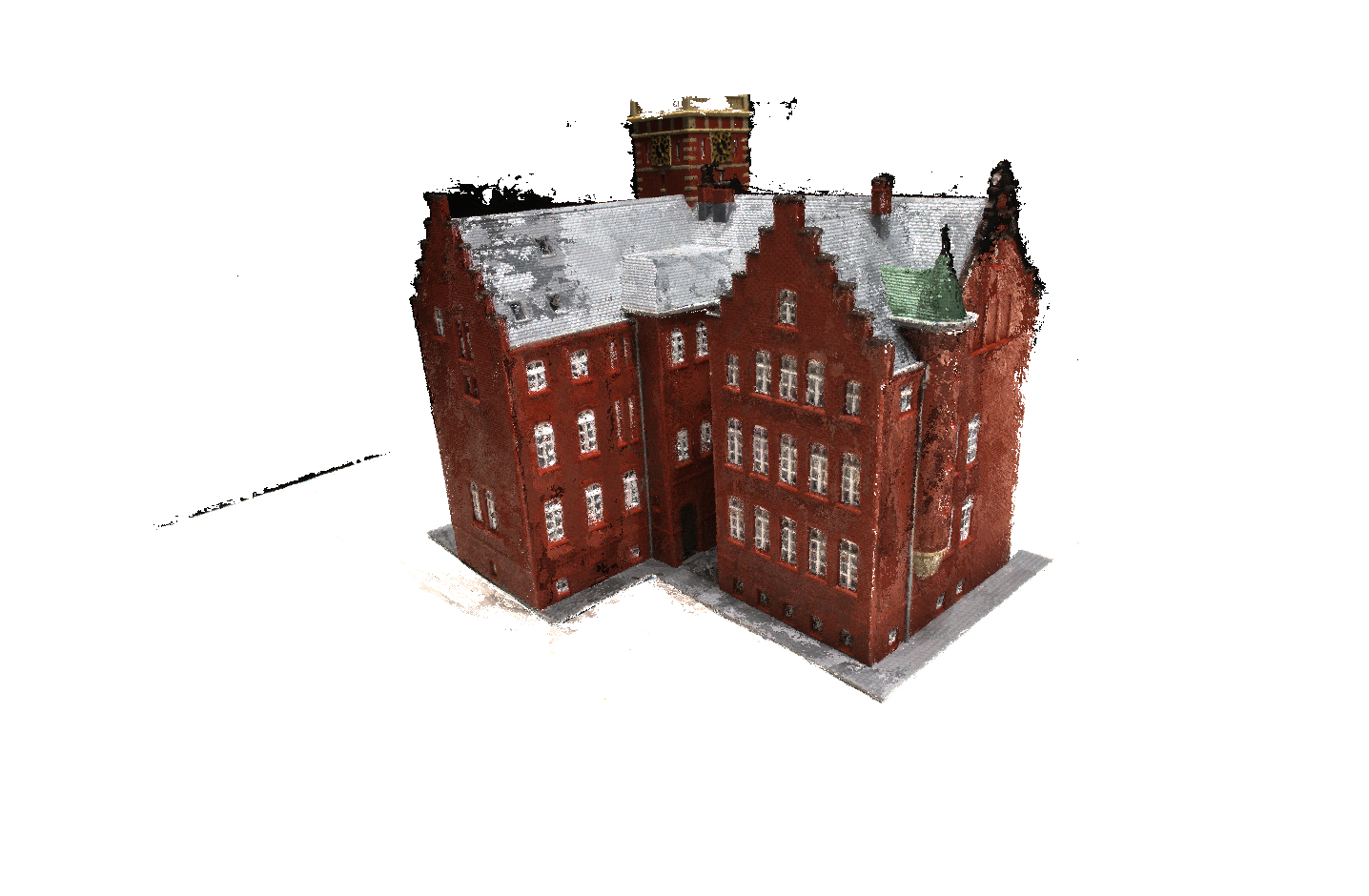}  &
    \includegraphics[width=0.22\linewidth,trim={8cm  5cm 8cm  3cm},clip]{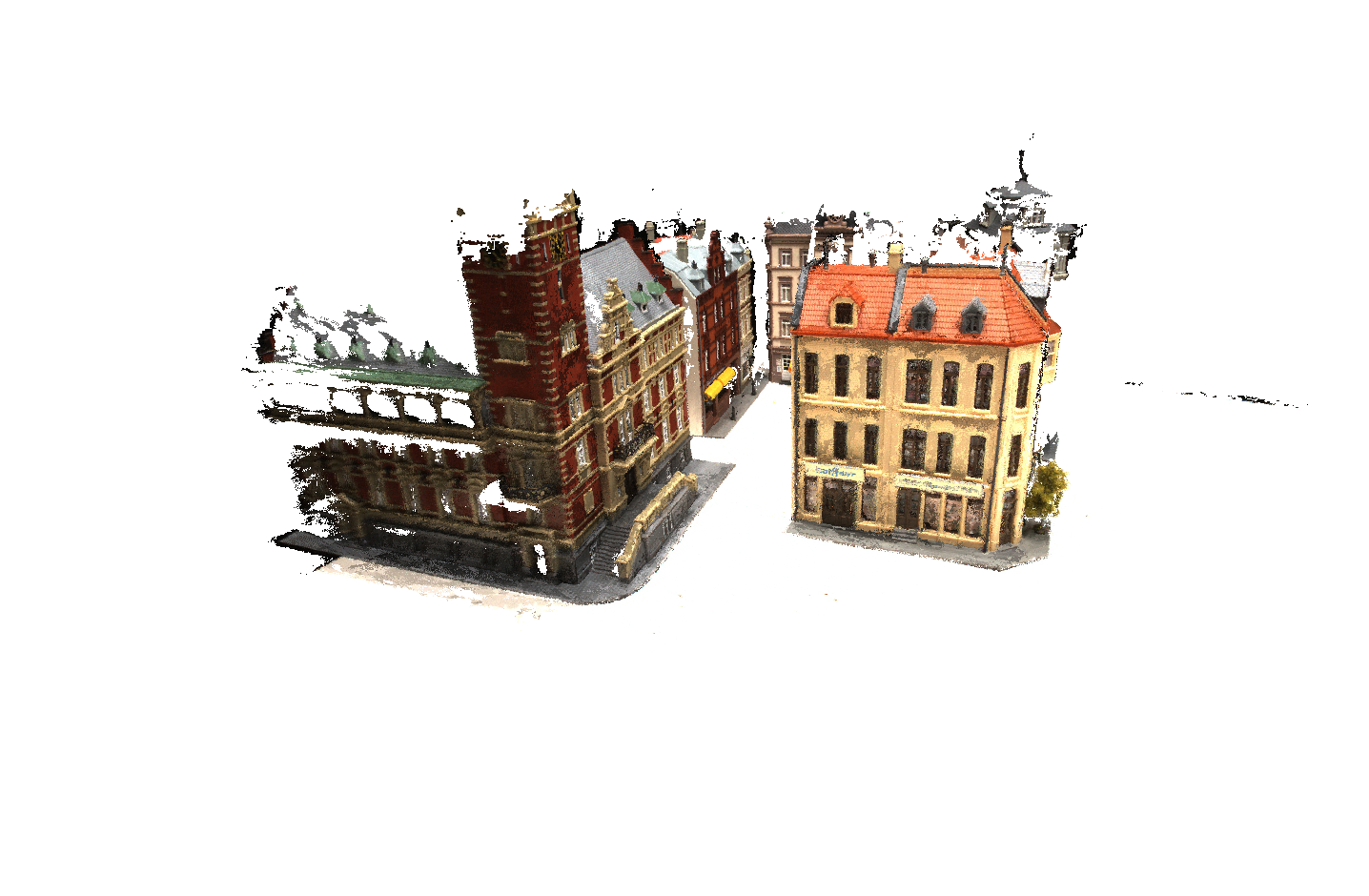}
    \end{tabular}
    \vspace{0.1em}
    \caption{\textbf{Qualitative results on DTU.} We show 3D reconstruction of different objects and scenes provided by DTU to assess the capability of our approach to generate accurate point cloud reconstructions even though we focus on highly detailed depth maps estimation.}
    \label{fig:dtu-qualitatives}
\end{figure*}

In Figure \ref{fig:dtu-qualitatives}, we report qualitative results about 3D reconstruction on DTU. We generate a depth map for each view available for a single scene, leveraging a total of 5 views for each prediction, and assemble such depth maps by applying geometric and photometric filtering common in the literature \cite{jensen2014large}.

\section{Keyframes Ranking Qualitative Results}

\begin{figure*}[h]
    \centering
    \renewcommand{\tabcolsep}{3pt}
    \begin{tabular}{@{}ccccccc@{}}
    \rotatebox[origin=l]{90}{\tiny \quad \quad \quad \textbf{Random}} &
    \linethickness{1.5pt}
    \begin{overpic}[width=0.14\linewidth]{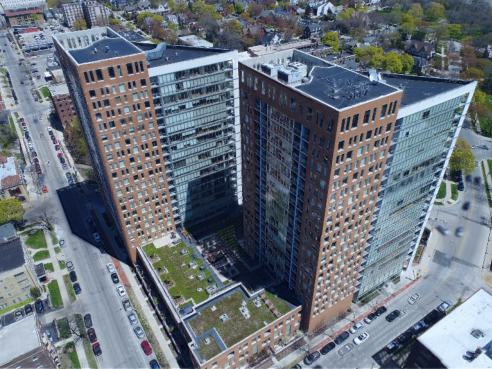}
    \put(10,22){\color{red}\line(0,2){45}}
     \put(10,67){\color{red}\line(2,0){30}}
     \put(40,22){\color{red}\line(0,2){45}}
     \put(10,22){\color{red}\line(2,0){30}}
    \end{overpic} &
    \includegraphics[width=0.14\linewidth]{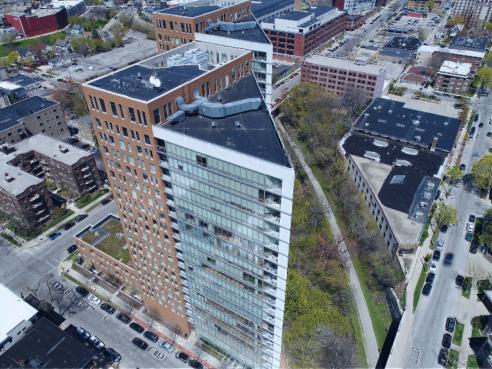} &
    \includegraphics[width=0.14\linewidth]{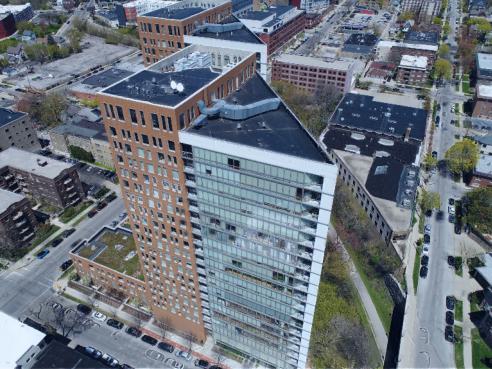} &
    \linethickness{1.5pt}
    \begin{overpic}[width=0.14\linewidth]{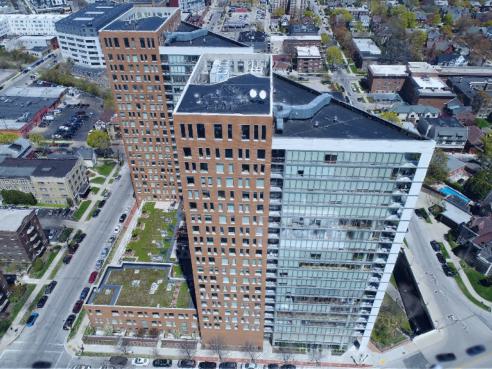}
    \put(18,32){\color{red}\line(0,2){42}}
     \put(18,74){\color{red}\line(2,0){22}}
     \put(40,32){\color{red}\line(0,2){42}}
     \put(18,32){\color{red}\line(2,0){22}}
     \end{overpic} &
    \includegraphics[width=0.14\linewidth]{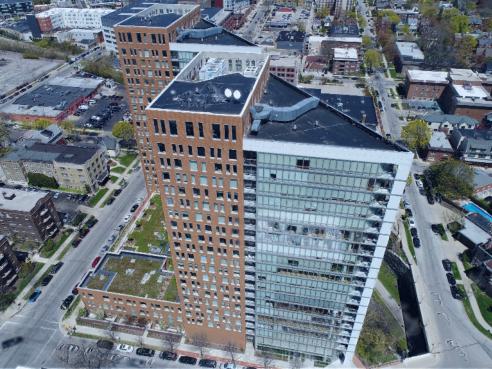} &
    \includegraphics[width=0.14\linewidth]{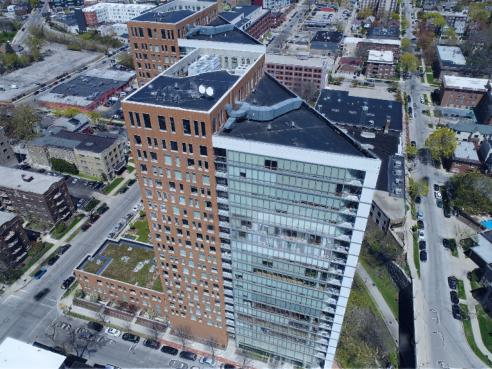} \\
    \rotatebox[origin=l]{90}{\tiny \quad \quad \quad \textbf{Ordered}} &
    \linethickness{1.5pt}
    \begin{overpic}[width=0.14\linewidth]{suppl_imgs/views_pruning/sample_blur_2/random/target.jpg}
    \put(10,22){\color{red}\line(0,2){45}}
     \put(10,67){\color{red}\line(2,0){30}}
     \put(40,22){\color{red}\line(0,2){45}}
     \put(10,22){\color{red}\line(2,0){30}}
    \end{overpic} &
    \linethickness{1.5pt}
    \begin{overpic}[width=0.14\linewidth]{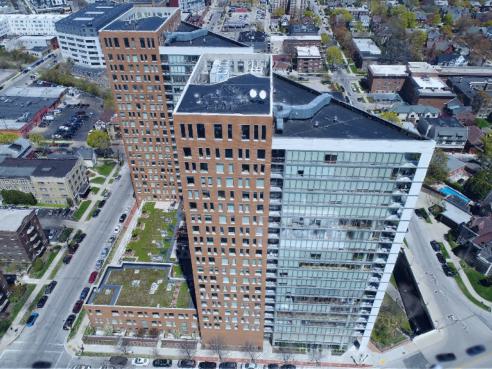}
    \put(18,32){\color{red}\line(0,2){42}}
     \put(18,74){\color{red}\line(2,0){22}}
     \put(40,32){\color{red}\line(0,2){42}}
     \put(18,32){\color{red}\line(2,0){22}}
     \end{overpic} &
    \includegraphics[width=0.14\linewidth]{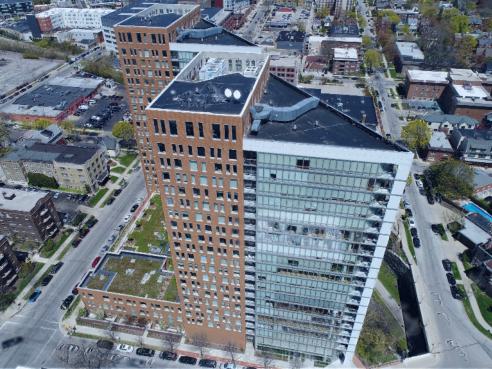} &
    \includegraphics[width=0.14\linewidth]{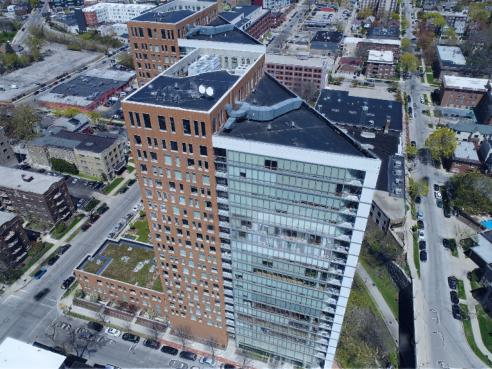} &
    \includegraphics[width=0.14\linewidth]{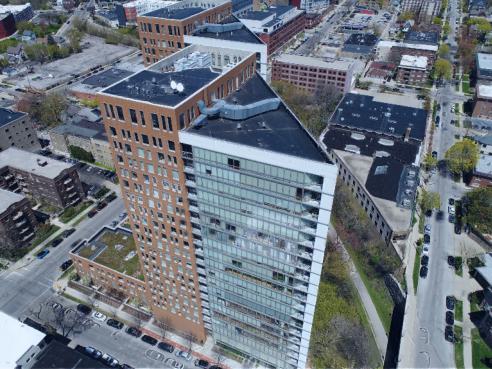} &
    \includegraphics[width=0.14\linewidth]{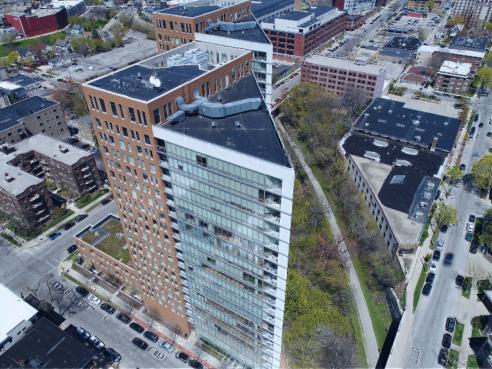} \\
    \rotatebox[origin=l]{90}{\tiny \quad \quad \quad \textbf{Ordered (Blur)}} &
    \linethickness{1.5pt}
    \begin{overpic}[width=0.14\linewidth]{suppl_imgs/views_pruning/sample_blur_2/random/target.jpg}
    \put(10,22){\color{red}\line(0,2){45}}
     \put(10,67){\color{red}\line(2,0){30}}
     \put(40,22){\color{red}\line(0,2){45}}
     \put(10,22){\color{red}\line(2,0){30}}
    \end{overpic} &
    \includegraphics[width=0.14\linewidth]{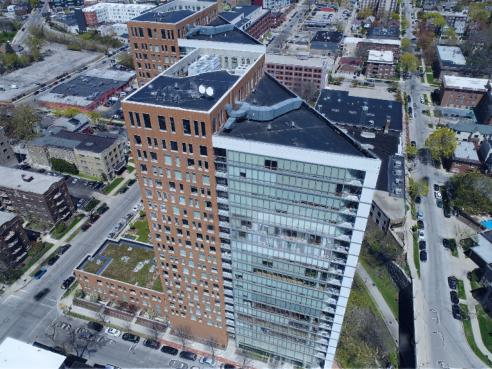} &
    \includegraphics[width=0.14\linewidth]{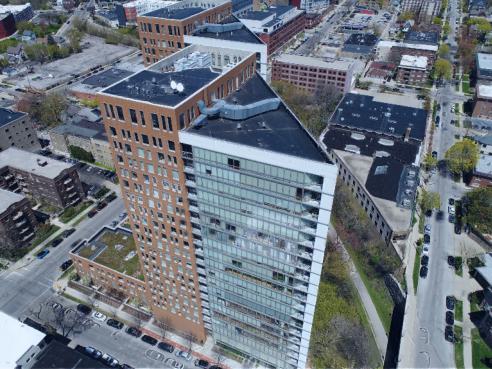} &
    \includegraphics[width=0.14\linewidth]{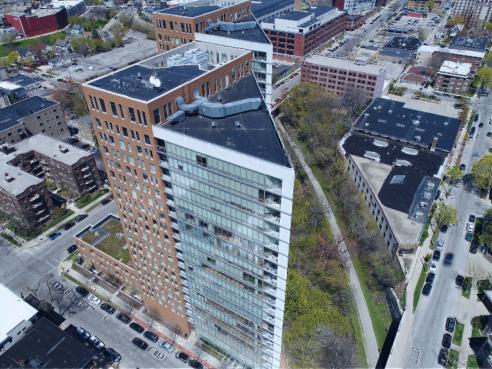} &
    \linethickness{1.5pt}
    \begin{overpic}[width=0.14\linewidth]{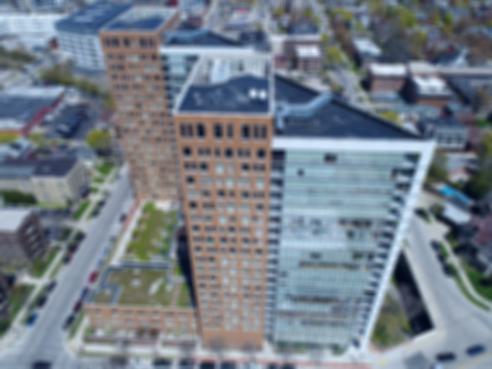}
    \put(18,32){\color{red}\line(0,2){42}}
     \put(18,74){\color{red}\line(2,0){22}}
     \put(40,32){\color{red}\line(0,2){42}}
     \put(18,32){\color{red}\line(2,0){22}}
    \end{overpic}&
    \includegraphics[width=0.14\linewidth]{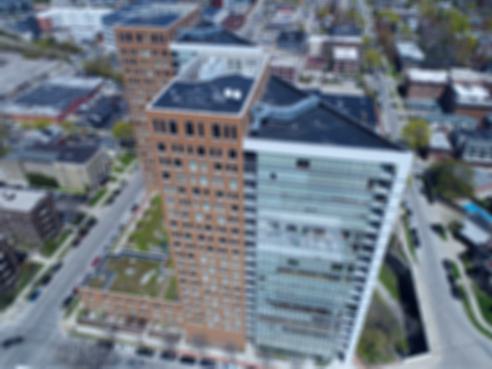} \\
    &
    \scriptsize \textbf{Reference} &
    \scriptsize \textbf{Source 1} &
    \scriptsize \textbf{Source 2} &
    \scriptsize \textbf{Source 3} &
    \scriptsize \textbf{Source 4} &
    \scriptsize \textbf{Source 5} \\
    \end{tabular}
    \vspace{0.1em}
    \caption{\textbf{Keyframes Ranking Example.} In the first row, we show a scene from Blended containing 5 source views in random order. In the second row, we show our framework reordering. If we apply Gaussian blur to the images with the best score and apply again reordering (last row), our framework assigns to them the worst score this time.}
    \label{fig:view-pruning}
\end{figure*}

We provide a qualitative example of the effectiveness of keyframes ranking enabled by our framework. In Figure \ref{fig:view-pruning}, we show a scene from Blended composed of 6 frames: the first one is the reference view, then source views follow in random order. We extract correlation scores with the procedure described in the main paper and order views accordingly in the second row. We can notice that higher scores are assigned to views with a higher visual overlap, e.g. the first source view in the ordered row is the one that maximizes matches with the building highlighted in red, the street on its left, and the garden between buildings, which are largely occluded in the other views. Finally, in the last row, we take the first 2 most correlated views according to our framework output, we apply a simple Gaussian blur to simulate out-of-focus images and rank once more. We can observe that our framework now assigns the lowest score to the out-of-focus views, although these were the best before. These experiments qualitatively demonstrate that our approach takes deeply into account not only the relative position between views but also the 3D structure of the scene and the quality of matches it can recover from the available views. Thus, our framework provides a view-centric methodology to discard poorly correlated views (e.g. out-of-focus, blurred, with moving objects), which cannot be achieved by reasoning only about relative pose.

\section{Source Views Scheduling Analysis}

As already detailed in the main manuscript, we apply a simple round-robin schedule to sample the source view used to sample correlation matches at each network iteration. This approach does not cause any particular problem. Indeed, even though the source view is changed at each iteration the depth state is independent of the latter, thus enforcing consistency. To assess that our approach is not significantly affected by source views ordering, we perform a simple experiment: on the Blended test set, we compute metrics for each permutation of the source views and compute the standard deviation of the performance. In Table \ref{tab:source-order} are reported the results of such experiment.
The very low variance reported at the very bottom confirms how the ordering we use to iterate over the source views has negligible impact on the final quality of the predicted depth map. 

\clearpage

\begin{table}[h]
    \centering
    \begin{tabular}{c|ccccccc}
    \hline \hline
    Permutation & MAE         & RMSE        & $>$1 m      & $>$2 m      & $>$3 m      & $>$4 m      & $>$8 m      \\
    \hline
    N. 1 & 0.316181 & 1.186300 & 0.069861 & 0.031390 & 0.017675 & 0.011238 & 0.003503 \\
    N. 2 & 0.317703 & 1.188342 & 0.070145 & 0.031508 & 0.017682 & 0.011203 & 0.003491 \\
    N. 3 & 0.318048 & 1.187708 & 0.070530 & 0.031465 & 0.017649 & 0.011226 & 0.003501 \\
    N. 4 & 0.319271 & 1.187811 & 0.070630 & 0.031412 & 0.017647 & 0.011224 & 0.003536 \\
    N. 5 & 0.315665 & 1.186850 & 0.069935 & 0.031355 & 0.017527 & 0.011122 & 0.003460 \\
    N. 6 & 0.316765 & 1.187873 & 0.070035 & 0.031531 & 0.017691 & 0.011216 & 0.003522 \\
    \hline
    Std. & 0.001328  & 0.000755 & 0.000319 & 0.000069 & 0.000061 & 0.000042 & 0.000026 \\
    \hline \hline
    \end{tabular}
    \caption{\textbf{Source Views Scheduling Analysis.} For each sample in the test set of Blended we evaluate each permutation of the source views and compute the standard deviation of the metrics. We use 3 source views to limit the number of permutations.}
    \label{tab:source-order}
\end{table}

\section{Qualitative Results -- impact of the depth range}

\begin{figure}[h]
    \renewcommand{\tabcolsep}{2pt}
    \begin{tabular}{ccccccc}
    \rotatebox[origin=l]{90}{\tiny \quad \quad \quad \textbf{Views}} &
    \includegraphics[width=.12\textwidth]{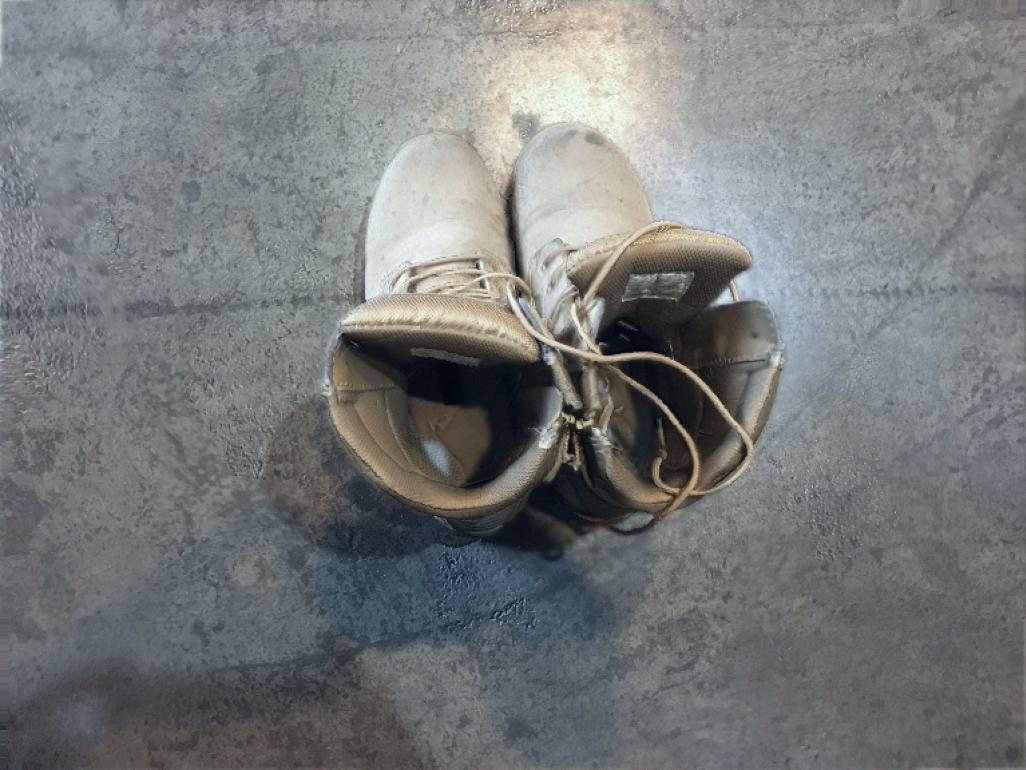} &
    \includegraphics[width=.12\textwidth]{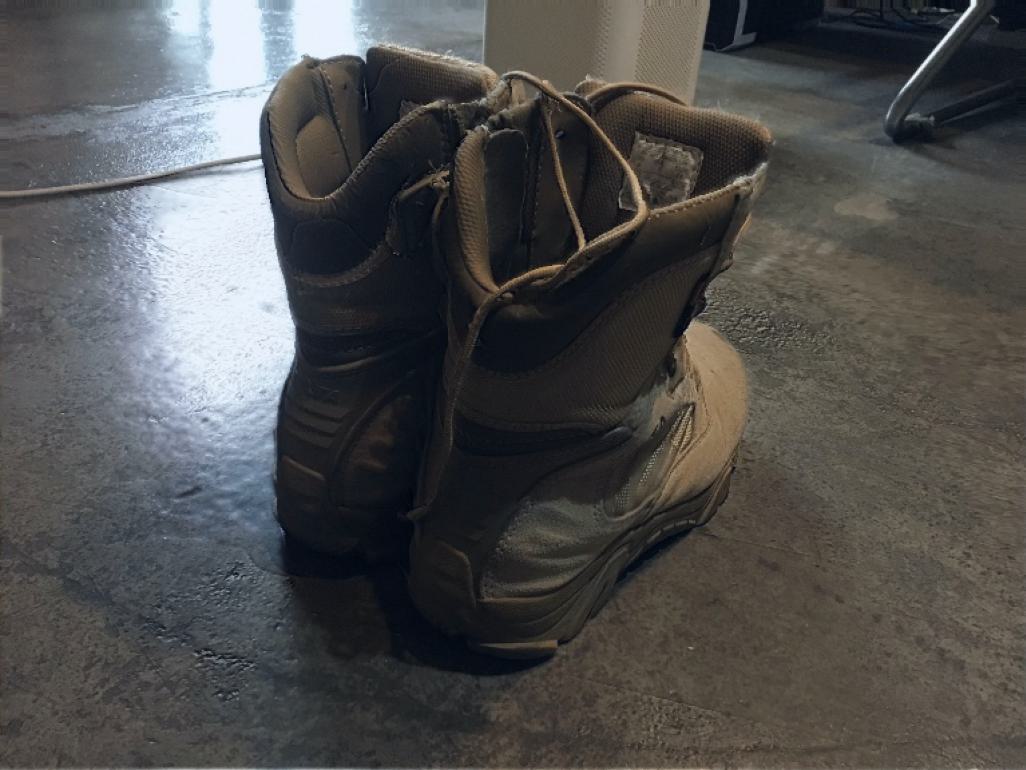} &
    \includegraphics[width=.12\textwidth]{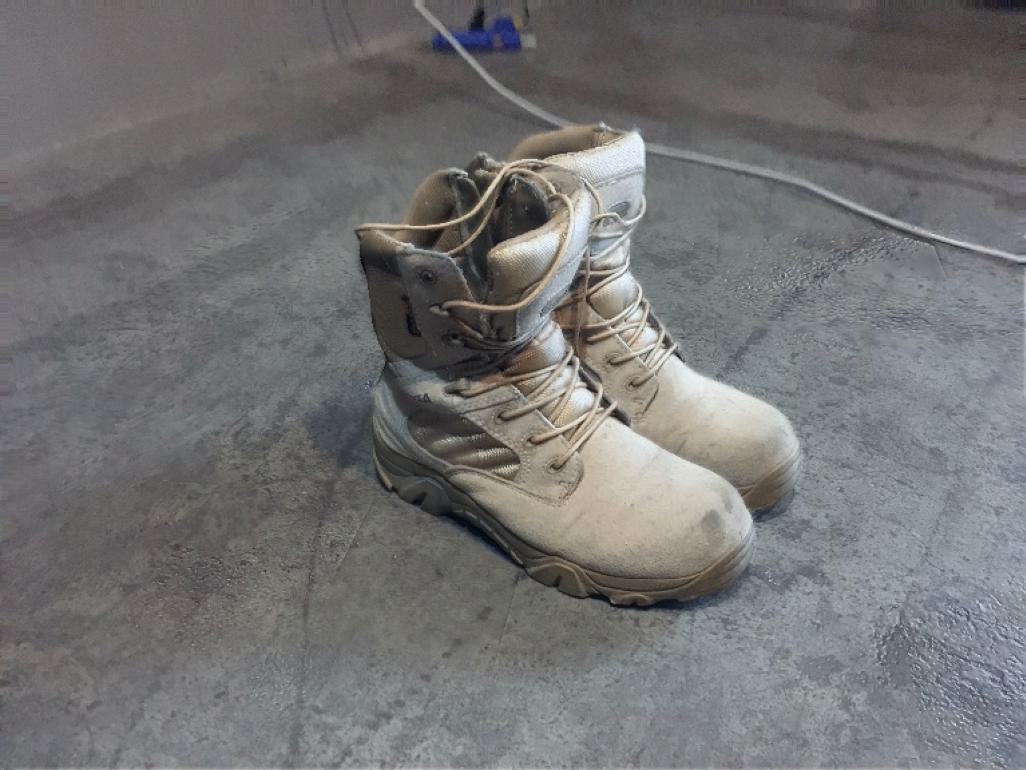} &
    \includegraphics[width=.12\textwidth]{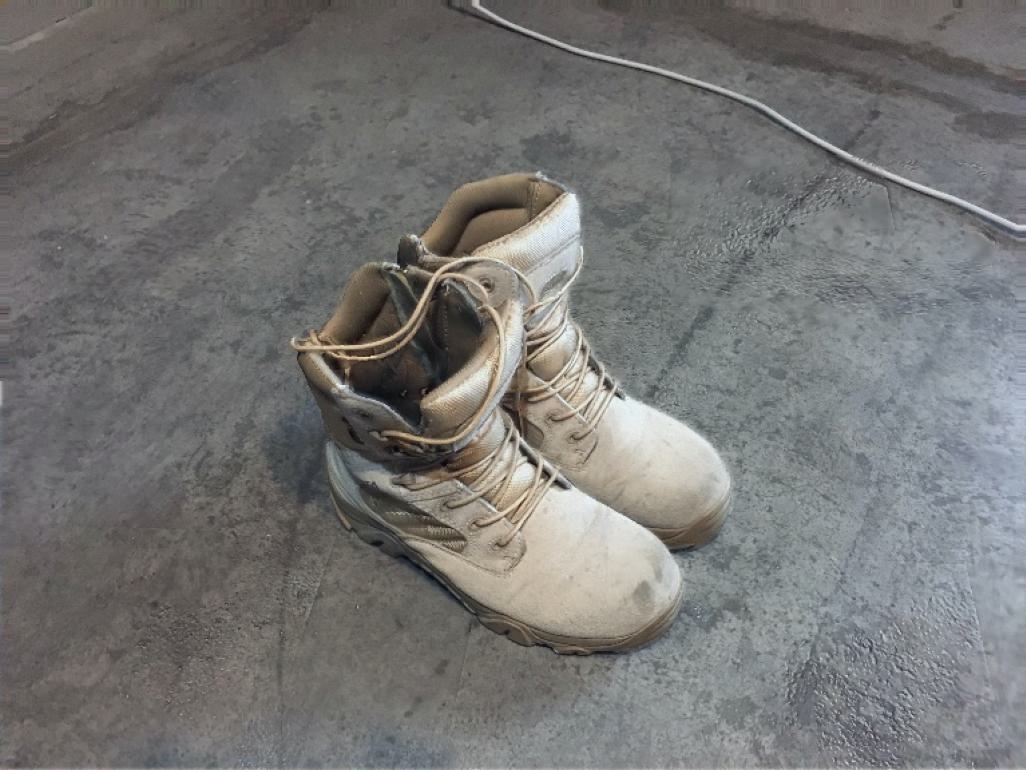} &
    \includegraphics[width=.12\textwidth]{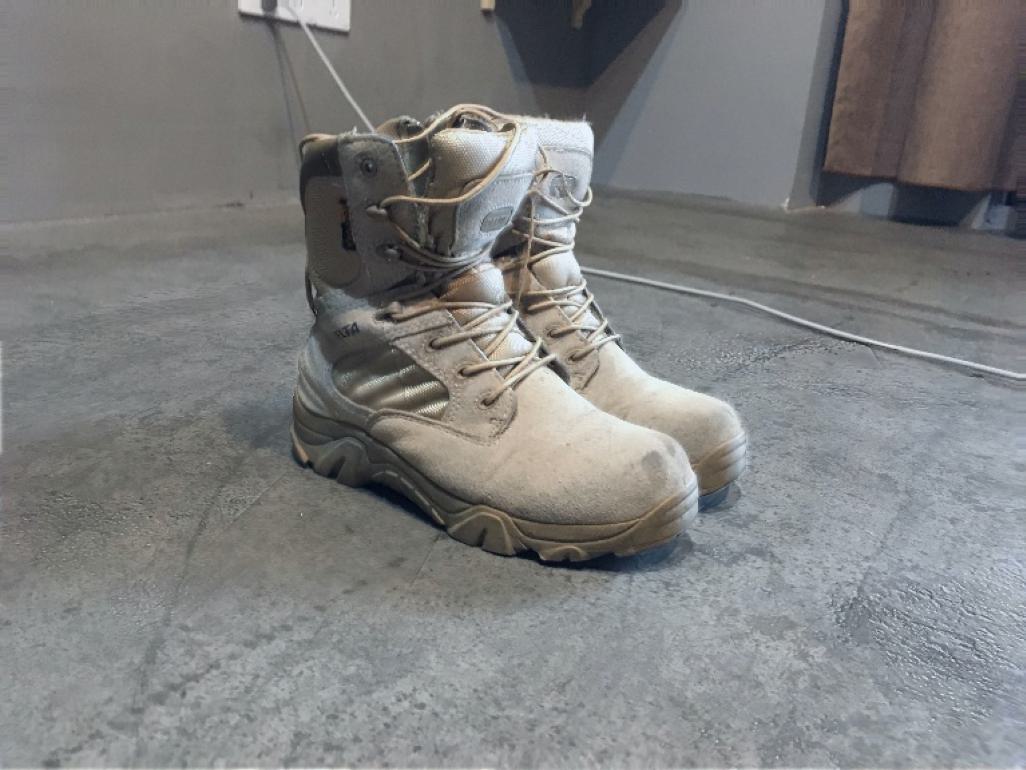}
    \\
    \rotatebox[origin=l]{90}{\tiny \quad \textbf{Ground Truth}} &
    \includegraphics[width=.12\textwidth]{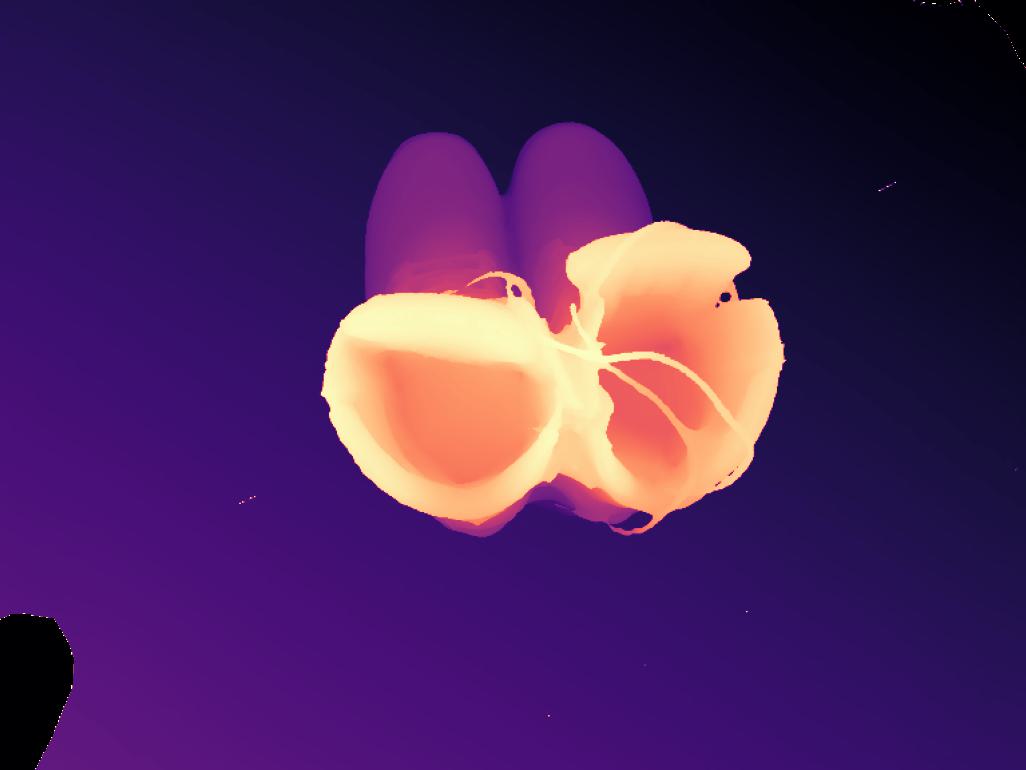} &
    \includegraphics[width=.12\textwidth]{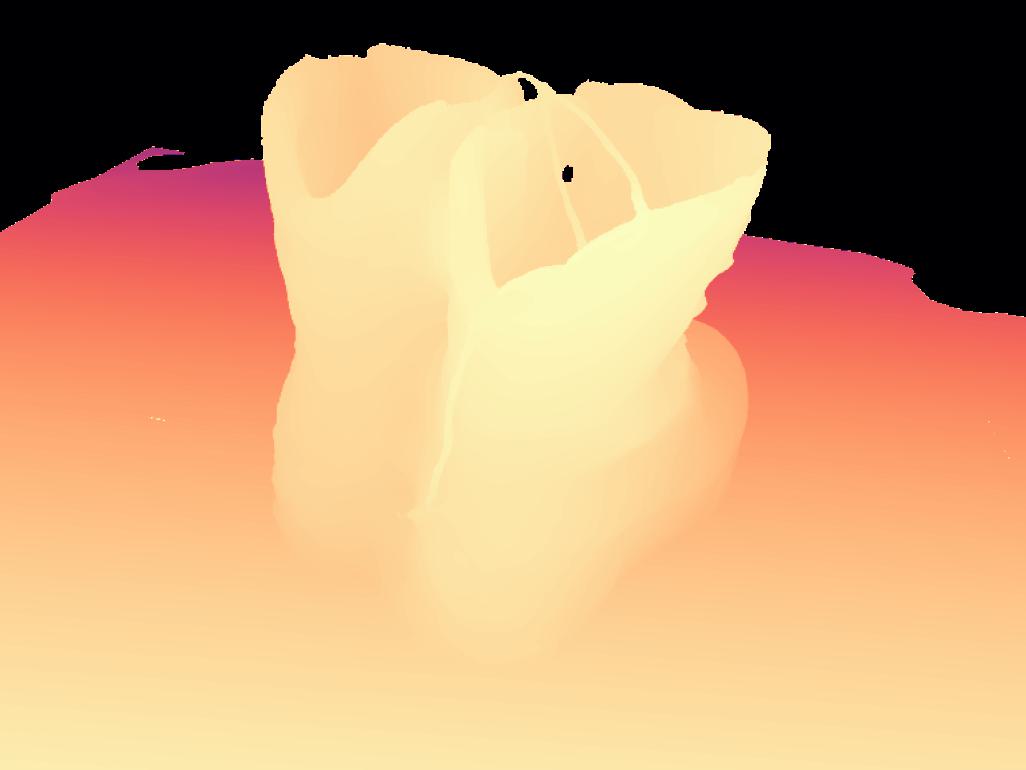} &
    \includegraphics[width=.12\textwidth]{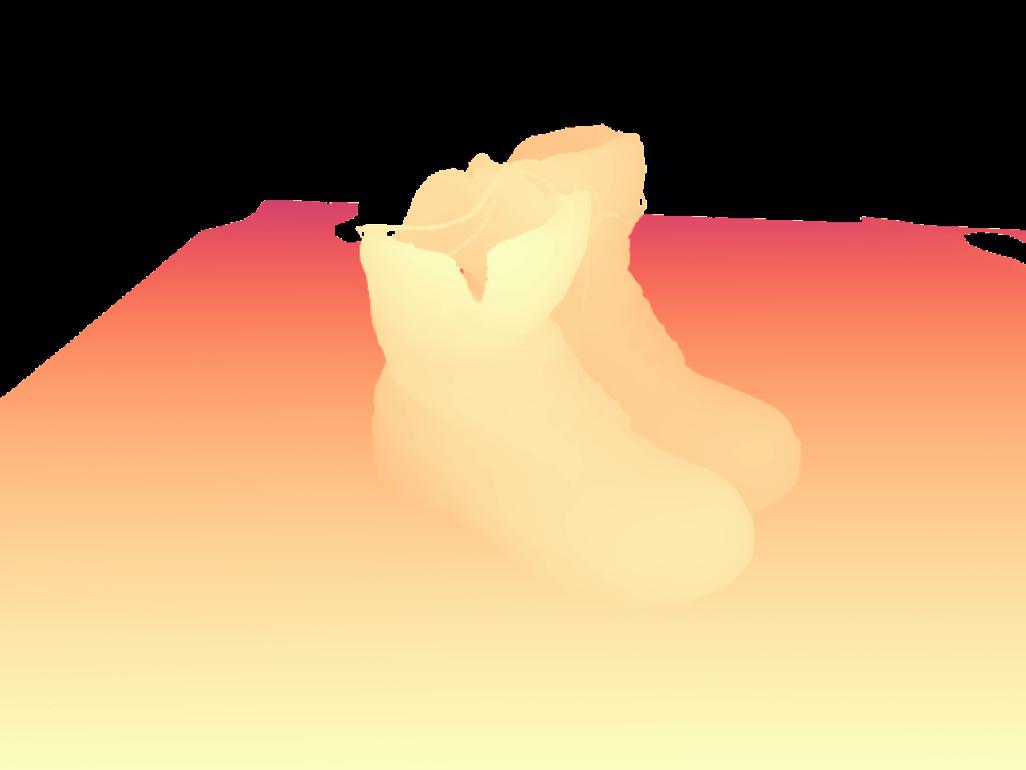} &
    \includegraphics[width=.12\textwidth]{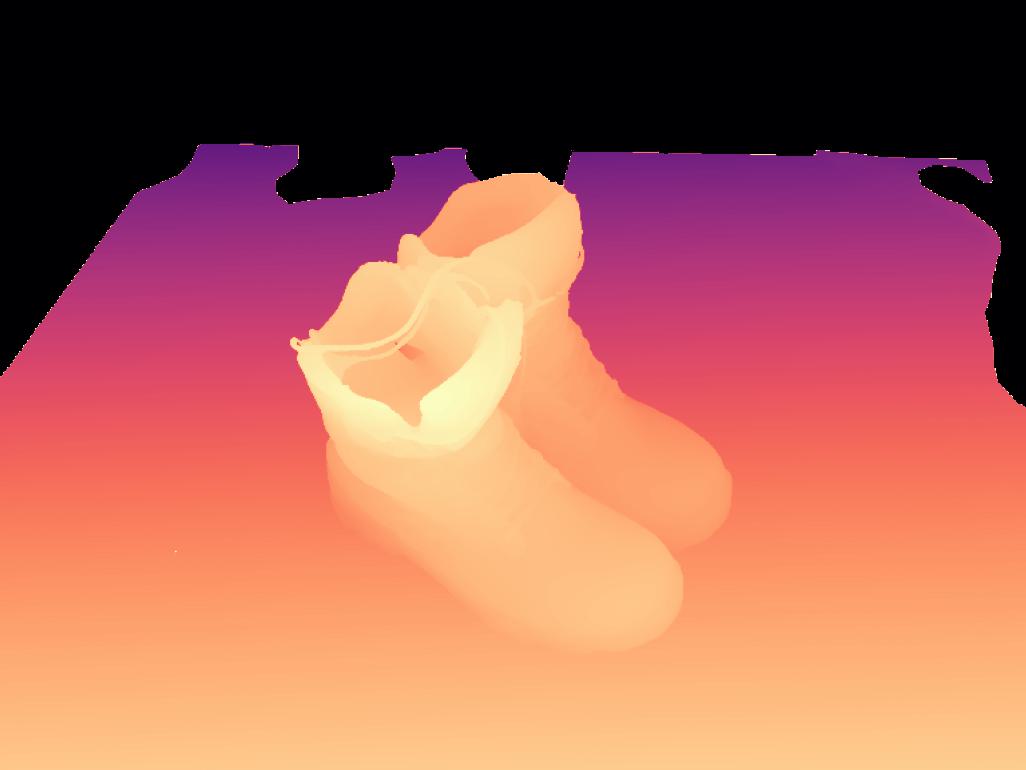} &
    \includegraphics[width=.12\textwidth]{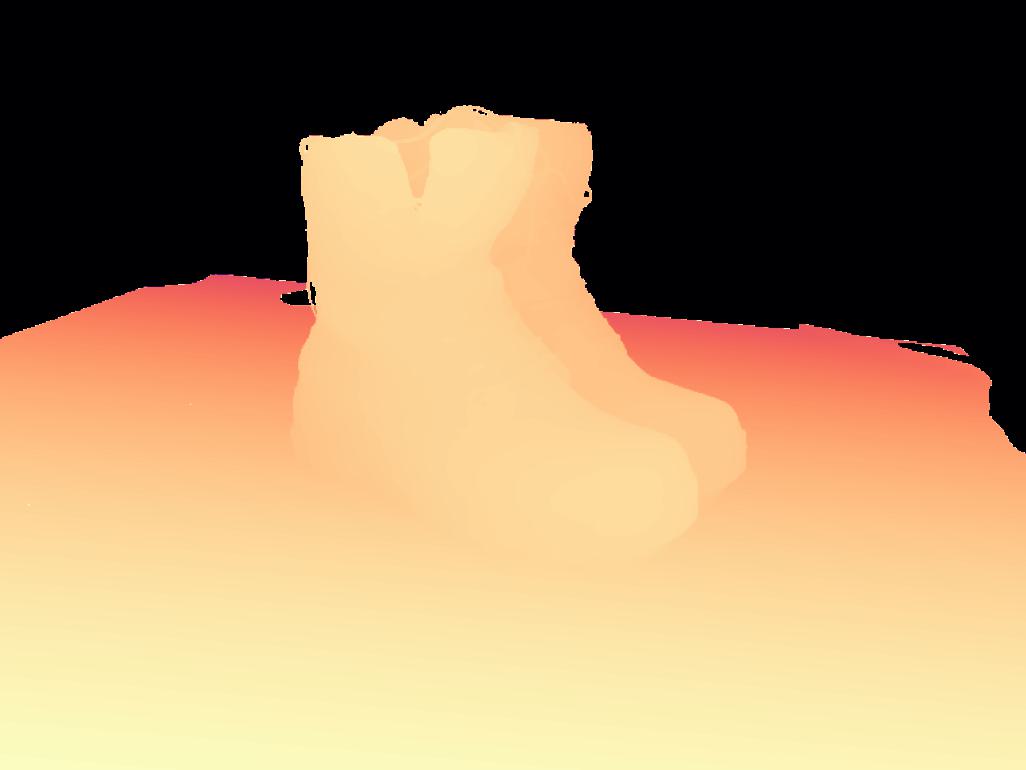}
    \\
    \rotatebox[origin=l]{90}{\tiny \quad \quad \quad \textbf{Ours}} &
    \includegraphics[width=.12\textwidth]{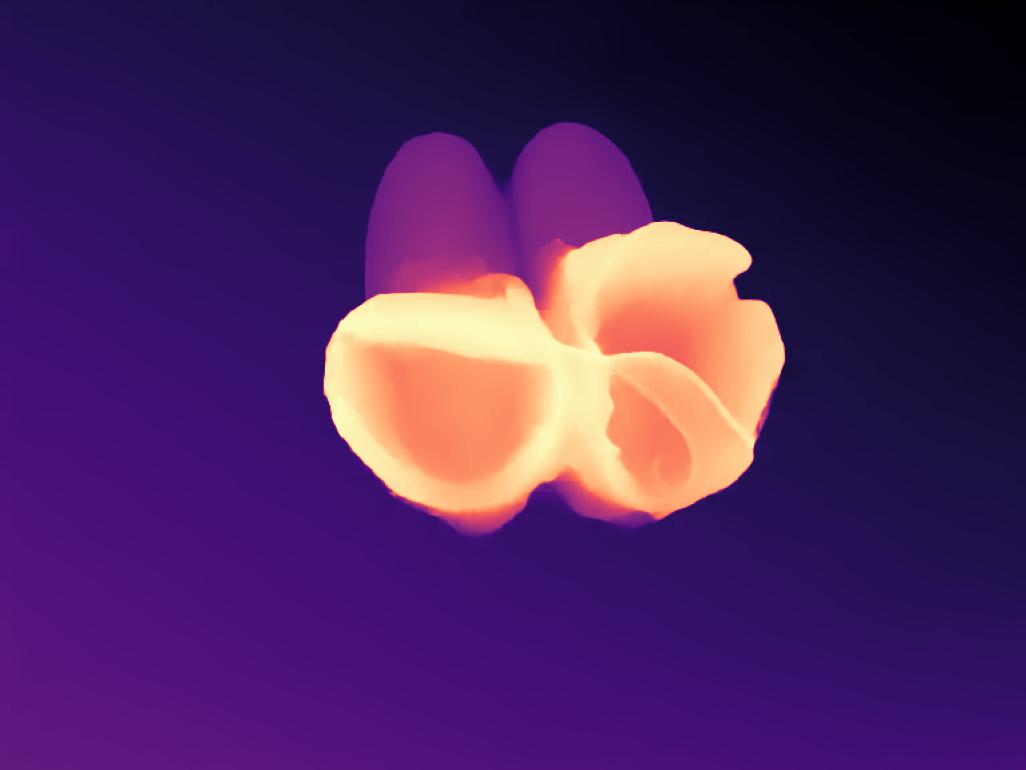} &
    \includegraphics[width=.12\textwidth]{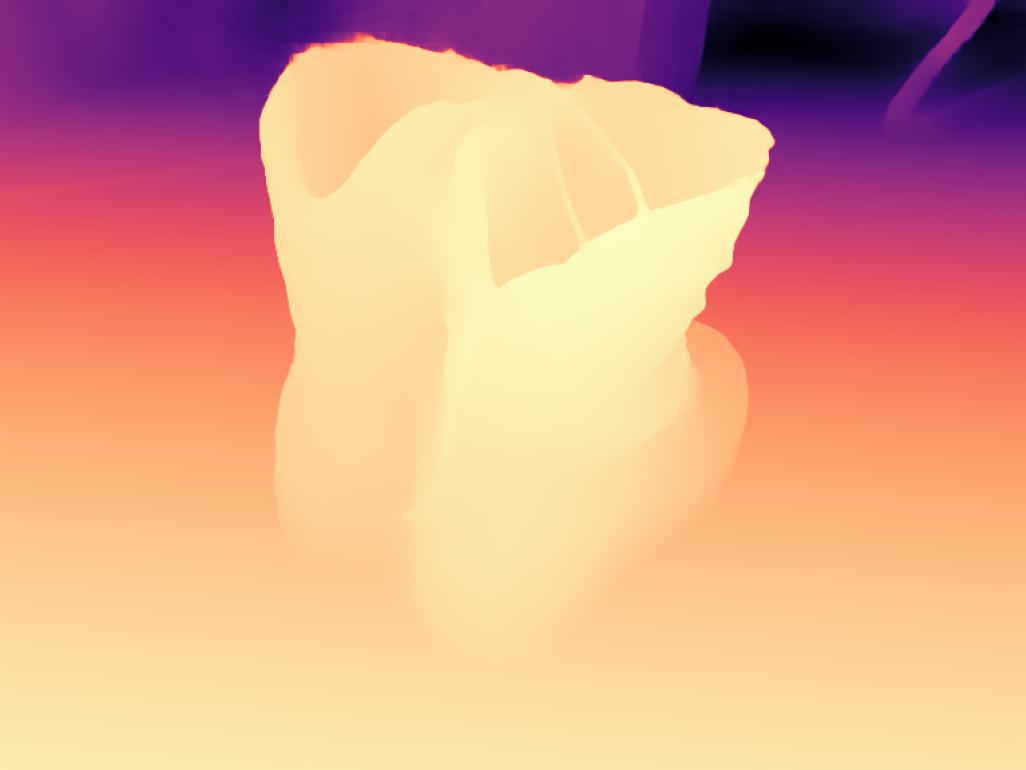} &
    \includegraphics[width=.12\textwidth]{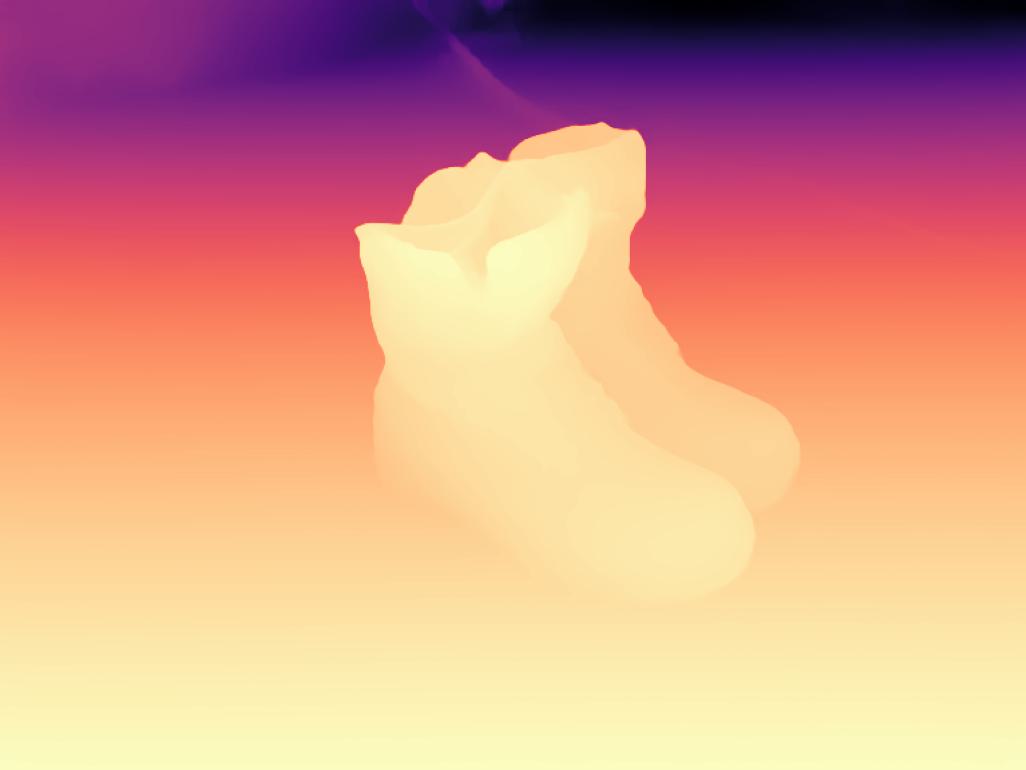} &
    \includegraphics[width=.12\textwidth]{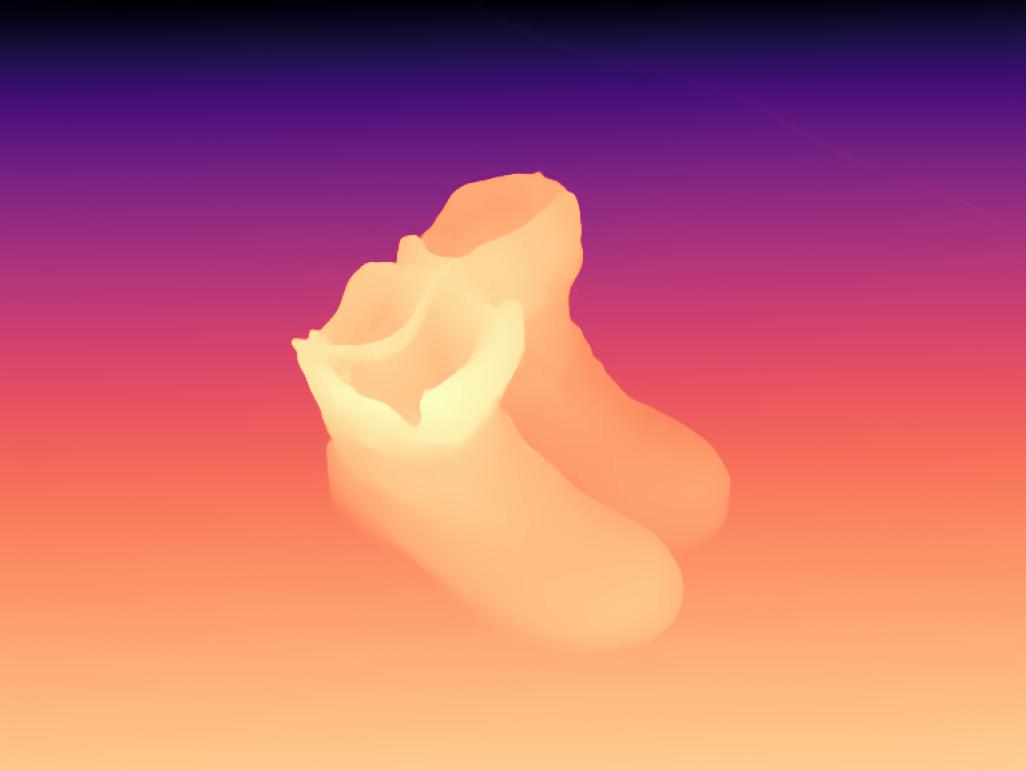} &
    \includegraphics[width=.12\textwidth]{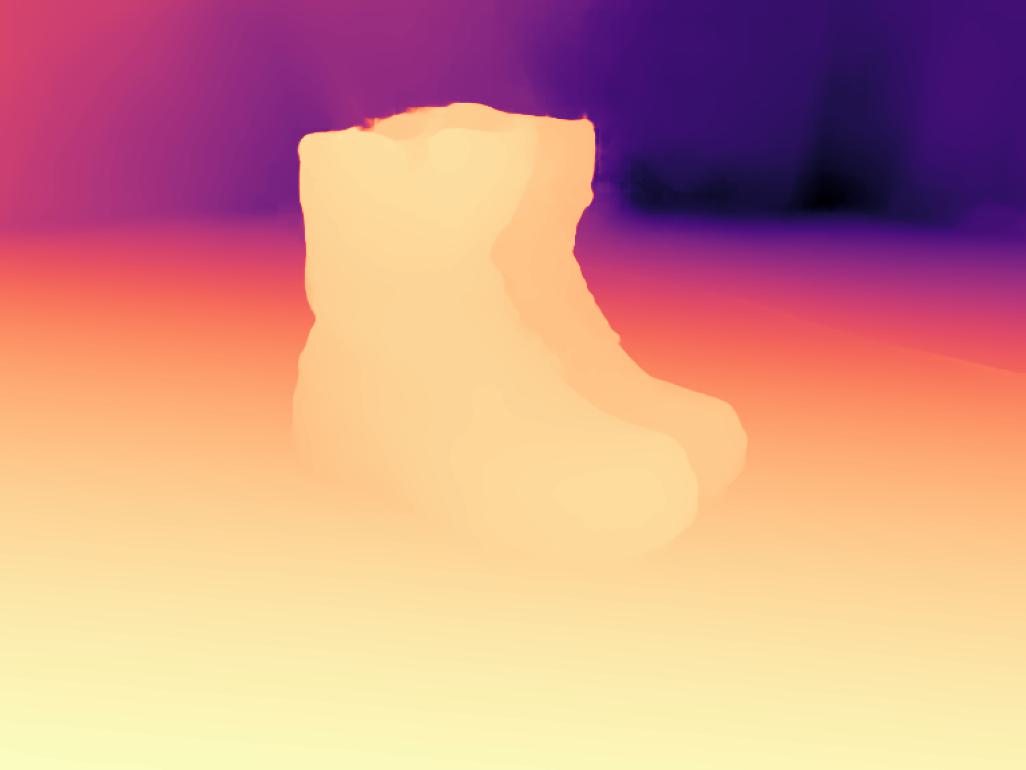}
    \\
    \rotatebox[origin=l]{90}{\tiny \quad \ \ \textbf{Gu et al. \cite{gu2020cascade}}} &
    \includegraphics[width=.12\textwidth]{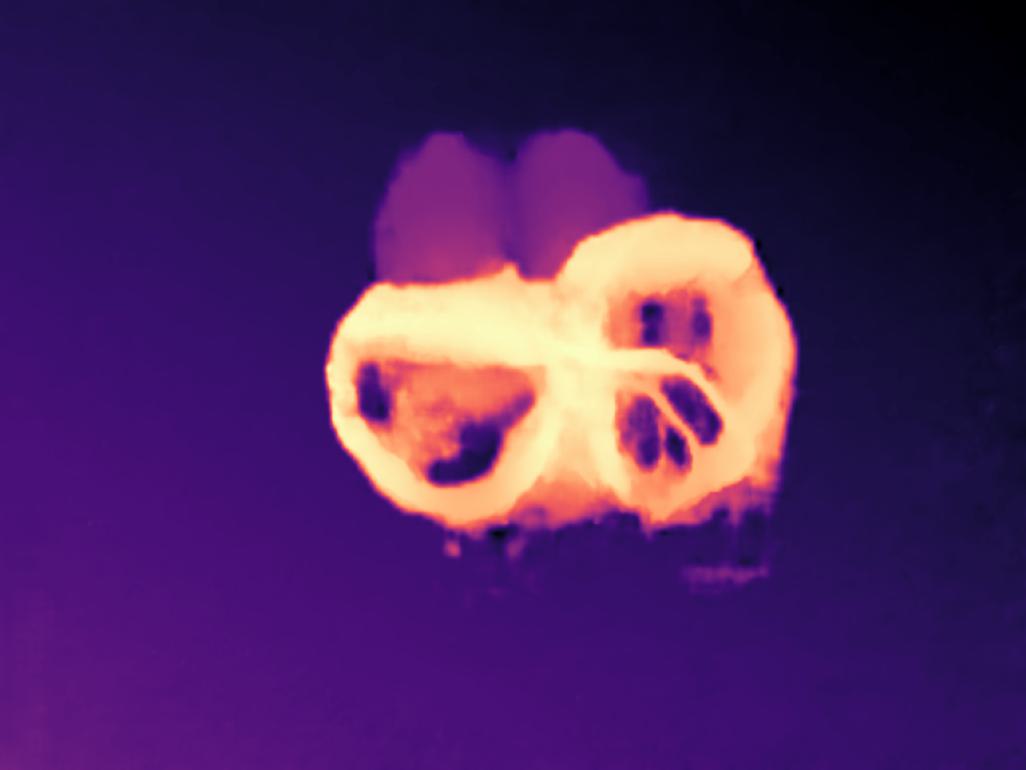} &
    \includegraphics[width=.12\textwidth]{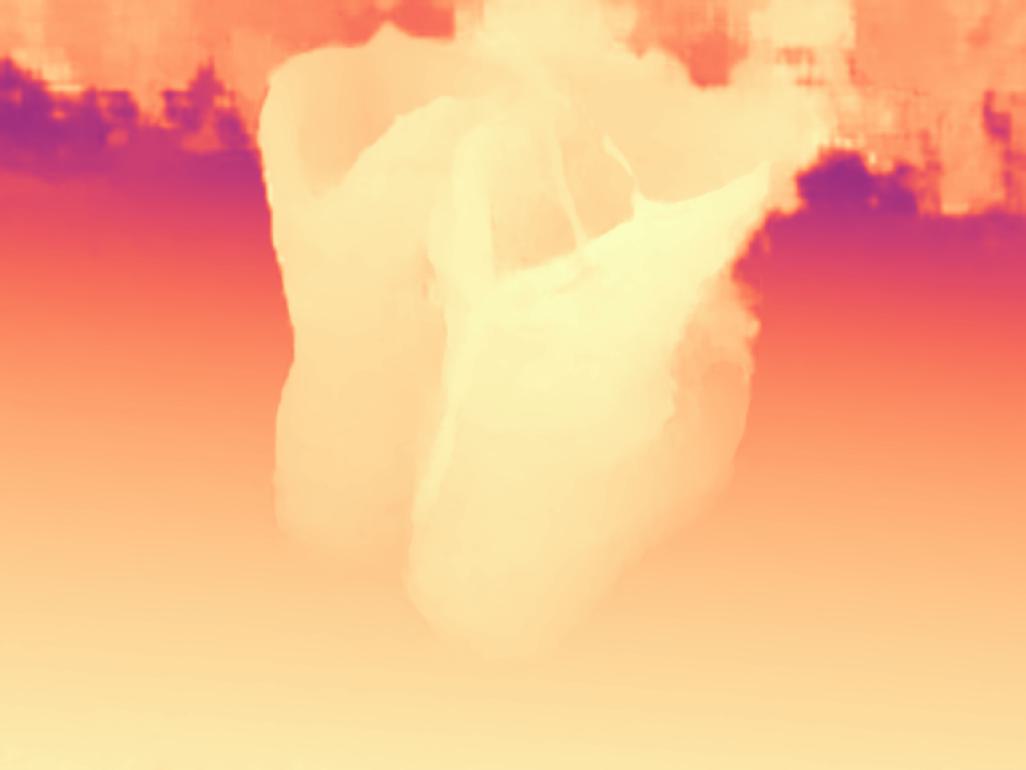} &
    \includegraphics[width=.12\textwidth]{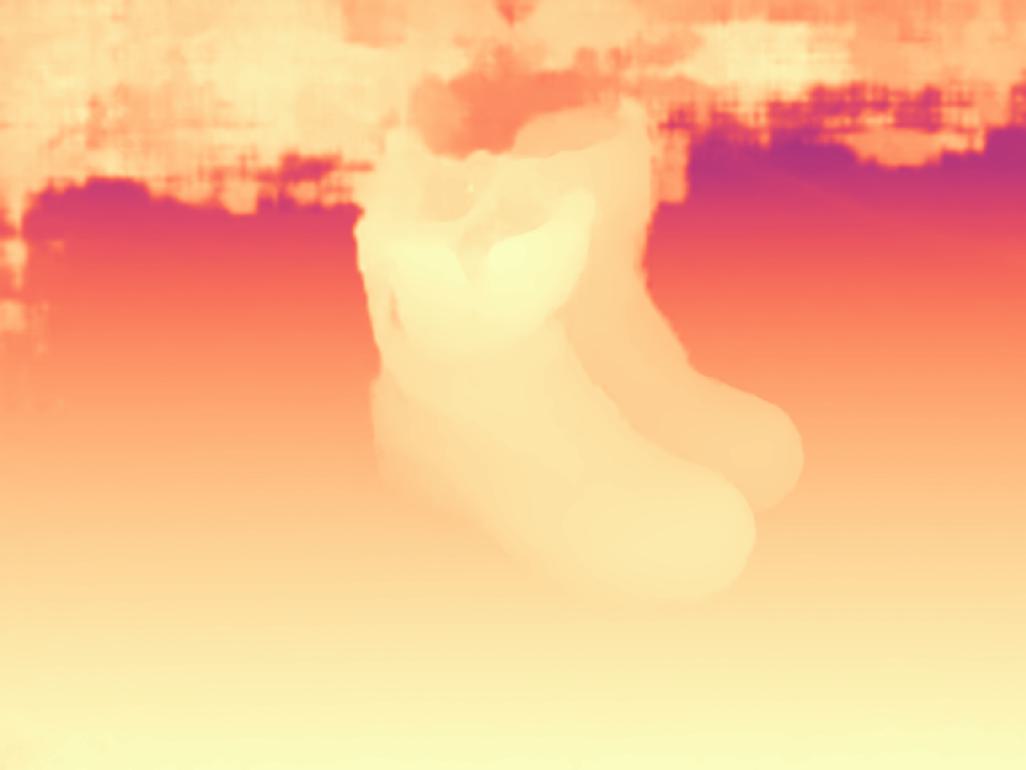} &
    \includegraphics[width=.12\textwidth]{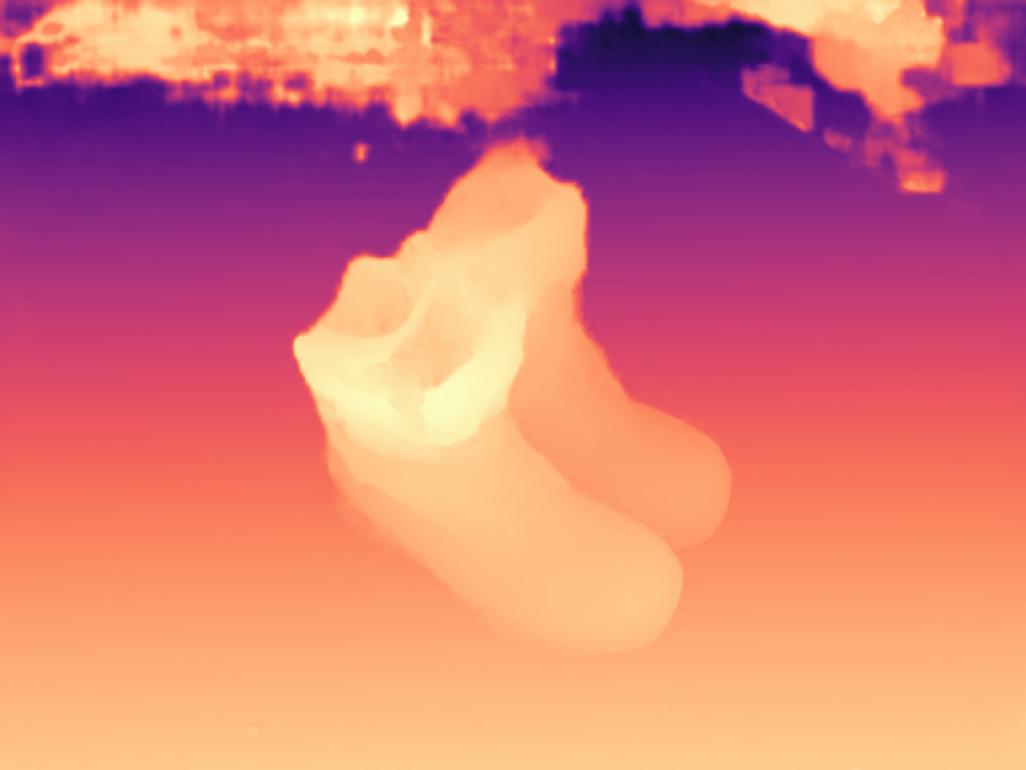} &
    \includegraphics[width=.12\textwidth]{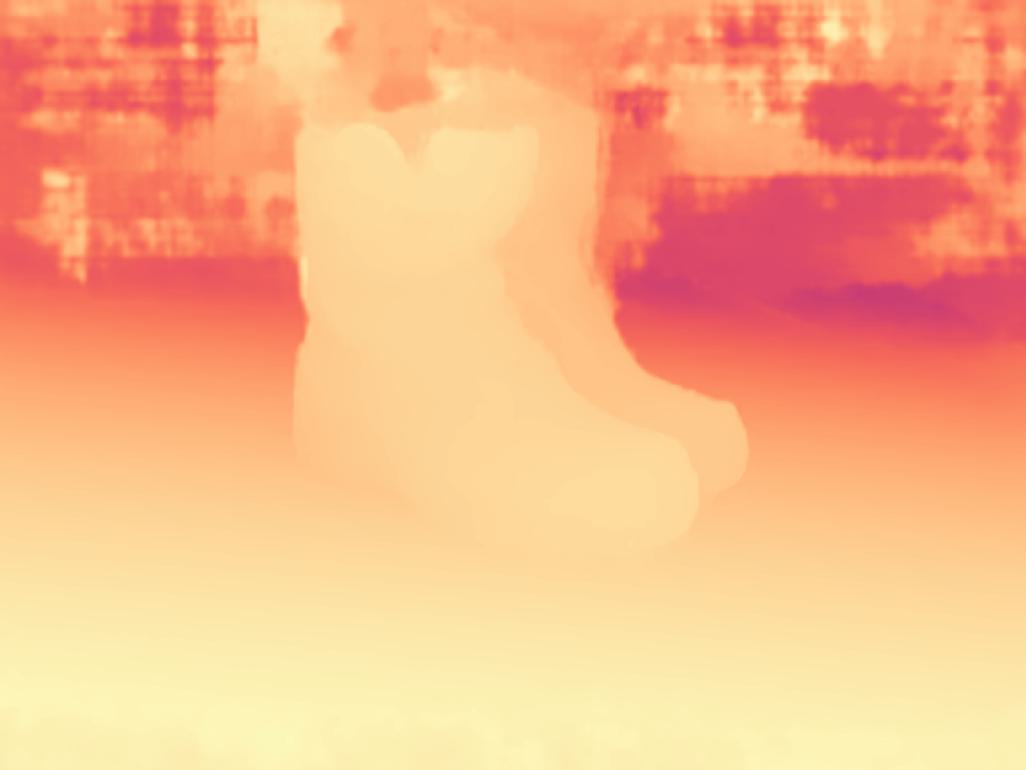}
    \\
    \rotatebox[origin=l]{90}{\tiny \quad \ \ \textbf{Wang et al. \cite{wang2021patchmatchnet}}} &
    \includegraphics[width=.12\textwidth]{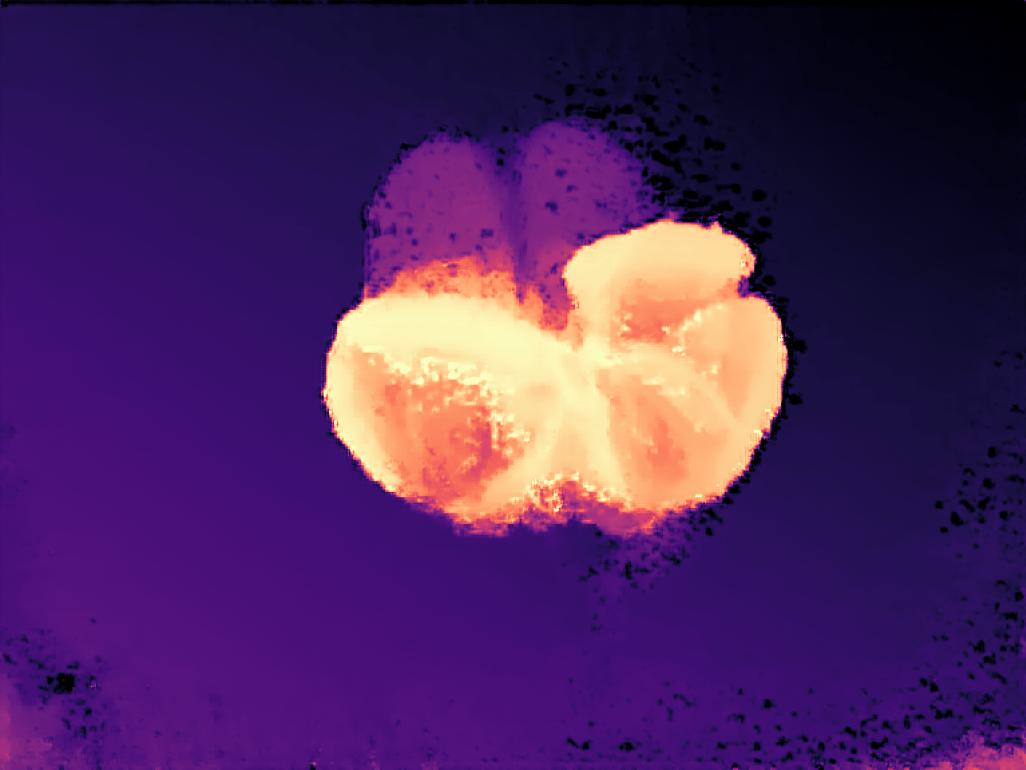} &
    \includegraphics[width=.12\textwidth]{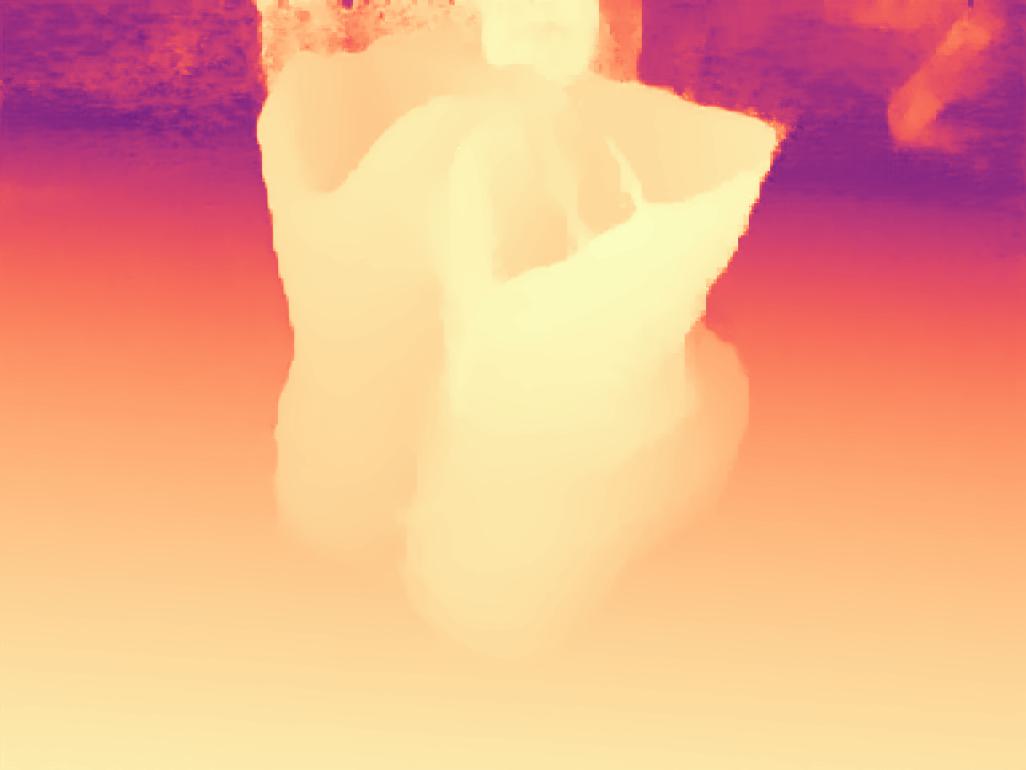} &
    \includegraphics[width=.12\textwidth]{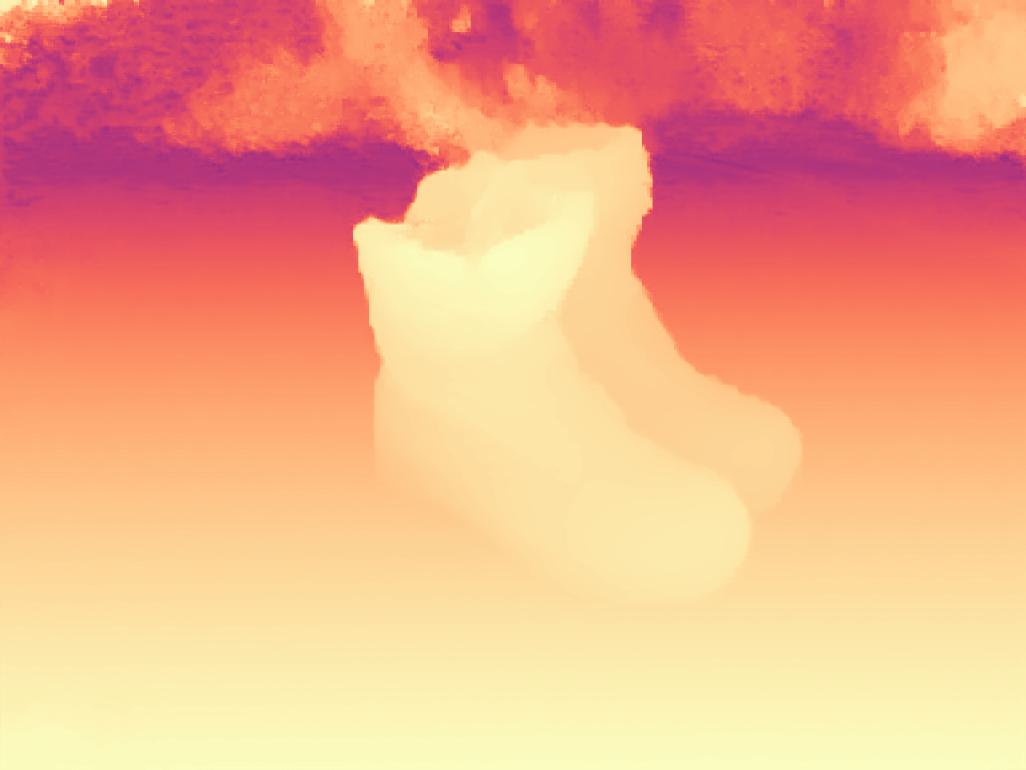} &
    \includegraphics[width=.12\textwidth]{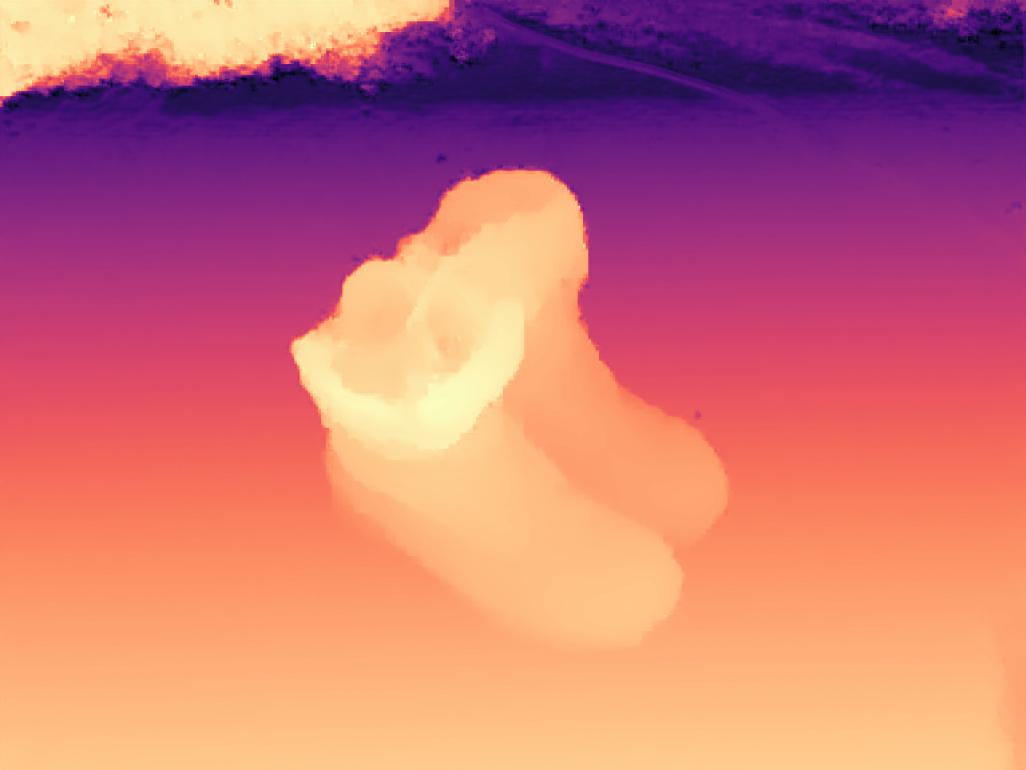} &
    \includegraphics[width=.12\textwidth]{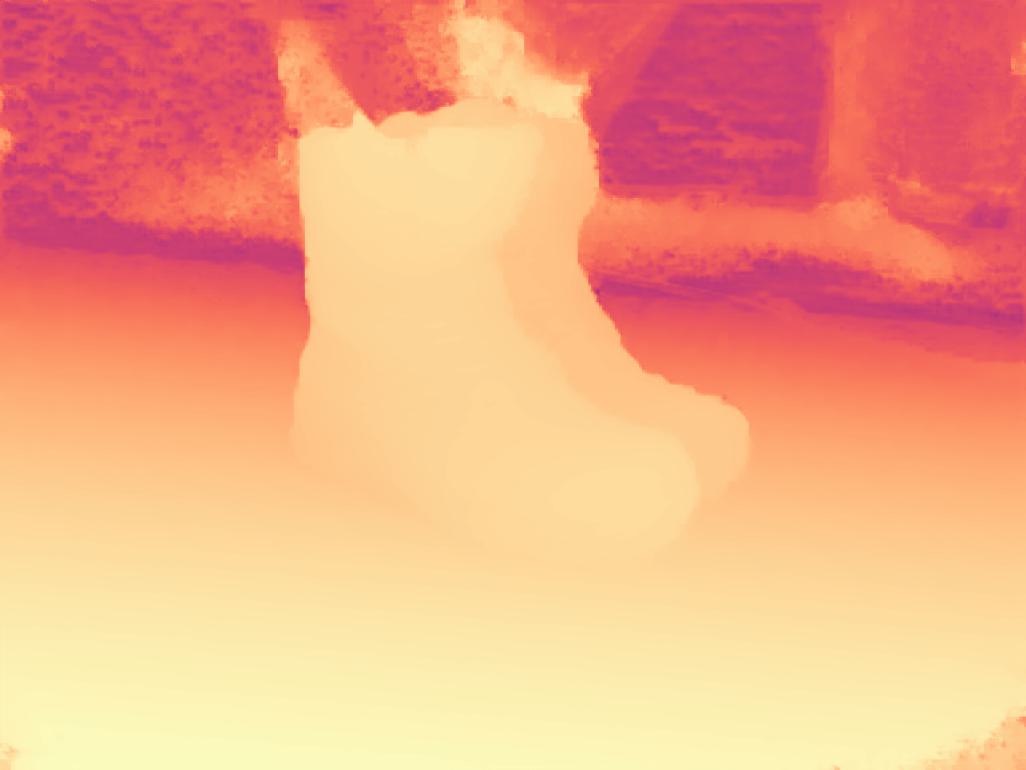}
    \\
    \end{tabular}
    \begin{tabular}{c}
    \includegraphics[width=0.29\textwidth]{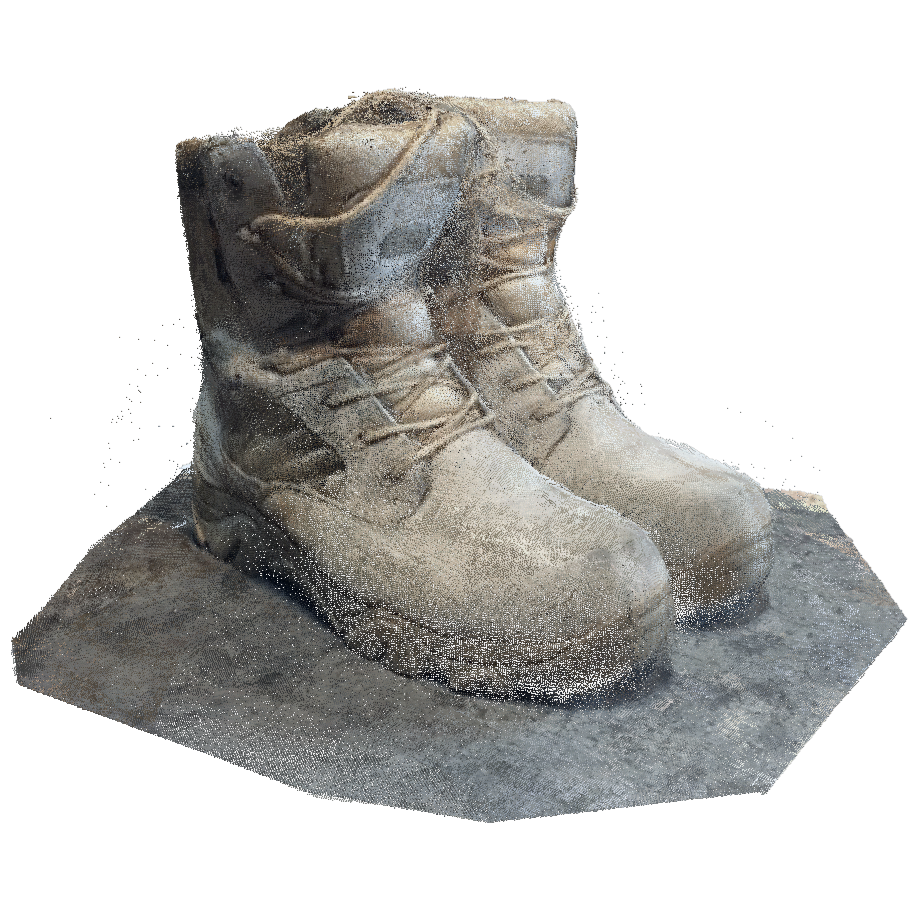}
    \end{tabular}
    \captionof{figure}{
    \textbf{Wrong Depth Range Effects Example on Blended} On left: five views from the Blended \cite{yao2020blendedmvs} scene and their ground-truth depth, followed by depth maps estimated by our framework, \cite{gu2020cascade} and \cite{wang2021patchmatchnet}. Our approach does not require any knowledge about the depth range and thus provides consistently smoother depth maps on the entire scene, even out of the pre-defined range where \cite{gu2020cascade} and \cite{wang2021patchmatchnet} struggle. On the right: 3D reconstruction obtained by merging our predictions, we limit the floor reconstruction to better highlight the object details, despite the fact that our framework is able to reconstruct the whole area.
    }
\label{fig:wrong-depth-range}
\end{figure}

We report an example showing the negative impact that an inaccurate depth range can produce on the predictions of existing frameworks.
Figure \ref{fig:wrong-depth-range} shows, from top to bottom, five images taken from Blended, their corresponding ground-truth depth, and the predictions by our framework and two prior works \cite{gu2020cascade,barnes2009patchmatch}.
These latter expose large artifacts in the farthest regions of the images, caused by the inaccurate depth range over which they operate. Indeed, as we can notice in the second row, ground-truth depth is not provided for those regions, and thus the depth range used for computing the depth map does not contain them -- since it is estimated directly from ground-truth.
On the contrary, our model produces clean and detailed depth maps even in these portions of the images.

\section{Network Structure and Training Details}

\textbf{Architecture Details.} In this section, we describe the core components of our framework in detail. In Table \ref{tab:framework-overview}, each module is detailed in terms of layers along with their parameters, inputs (in red), and outputs (in blue). Input source and target views are encoded through the Feature Encoder, then disentangled information is extracted from the reference view through the Reference Encoder (called context in Table \ref{tab:framework-overview} and accounting 128 channels).
Depth, hidden state, and reference features are used to predict sampling offsets, correlation scores are sampled according with the methodology described in the main paper and the recurrent block predicts a new hidden state and a $\Delta$depth update. Finally, a shallow module predicts upsampling weights from the hidden state and reference information and performs convex upsampling.

\begin{table}[h]
    \centering
    \resizebox{\linewidth}{!}{
    \begin{tabular}{c c}
    \begin{tabular}{l|l|l|l|l|l}
        \hline
        Name & Layer & K & S & In/Out & Input \\
        \hline
        \multicolumn{6}{c}{Residual Block Stride 2}                                  \\
        \hline
        conv0      & Conv2D + BatchNorm2D + ReLU & 3 & 2  & In/Out  & \inp{input}    \\
        conv1      & Conv2D + BatchNorm2D + ReLU & 3 & 1  & Out/Out & conv0          \\
        downs      & Conv2D + BatchNorm2D + ReLU & 1 & 2  & In/Out  & \inp{input}    \\
        \out{out}  & ReLU                        & - & -  & Out/Out & downs + conv1  \\
        \hline
        \multicolumn{6}{c}{Residual Block Stride 1}                                  \\
        \hline
        conv0      & Conv2D + BatchNorm2D + ReLU & 3 & 2  & In/Out  & \inp{input}    \\
        conv1      & Conv2D + BatchNorm2D + ReLU & 3 & 1  & Out/Out & conv0          \\
        \out{out}  & ReLU                        & - & -  & Out/Out & conv1 + input  \\
        \hline
        \multicolumn{6}{c}{Feature Encoder \& Reference Encoder}                       \\
        \hline
        conv0       & Conv2D + BatchNorm2D + ReLU & 7 & 2  & 3/64    & \inp{image}   \\
        conv1       & Residual Block Stride 2     & - & -  & 64/64   & conv0         \\
        conv2       & Residual Block Stride 1     & - & -  & 64/64   & conv1         \\
        conv3       & Residual Block Stride 2     & - & -  & 64/96   & conv2         \\
        conv4       & Residual Block Stride 1     & - & -  & 96/96   & conv3         \\
        conv5       & Residual Block Stride 2     & - & -  & 96/128  & conv4         \\
        conv6       & Residual Block Stride 1     & - & -  & 128/128 & conv5         \\
        conv7       & Residual Block Stride 2     & - & -  & 96/128  & conv6         \\
        conv8       & Residual Block Stride 1     & - & -  & 128/128 & conv7         \\
        \out{feats} & Conv2D                      & 1 & 1  & 128/256 & conv8         \\
        \hline
    \end{tabular}
    &
    \begin{tabular}{l|l|l|l|l|l}
        \hline
        Name & Layer & K & S & In/Out & Input                                                 \\
        \hline
        \multicolumn{6}{c}{Offsets Computation}                                               \\
        \hline
        conv0 & Conv2D + BatchNorm2D + ReLU & 3 & 1  & 128+128+1/256 & \begin{tabular}{@{}l@{}} \inp{context}, \inp{hidden$_{s-1}$},\\\inp{depth$_{s-1}$} \end{tabular} \\
        \out{offsets} & Conv2D              & 1 & 1  & 256/9$\times$9$\times$2 & conv0        \\
        \hline
        \multicolumn{6}{c}{Recurrent Block}                                                   \\
        \hline
        corr0   & Conv2D + ReLU               & 1 & 1  & 9$\times$9/256 & \inp{corrfeats}     \\
        corr1   & Conv2D + ReLU               & 3 & 1  & 256/192        & corr0               \\
        dfeats0 & Conv2D + ReLU               & 7 & 1  & 1/128          & \inp{depth$_{s-1}$} \\
        dfeats1 & Conv2D + ReLU               & 3 & 1  & 128/64         & dfeats0             \\
        conv0   & Conv2D + ReLU               & 3 & 1  & 192+64/128-1   & dfeats1,\ corr1     \\
        hidden0             & ConvGRU2D                   & (1, 5) & 1 & 128+1+128+128/128 & \begin{tabular}{@{}l@{}} \inp{context},\ conv0,\\ \inp{depth$_{s-1}$},\ \inp{hidden$_{s-1}$} \end{tabular}\\
        \out{hidden$_s$}    & ConvGRU2D                   & (5, 1) & 1 & 128/128         & hidden$_0$           \\
        conv1               & Conv2D + ReLU               & 3      & 1 & 128/64          & \out{hidden$_s$}     \\
        \out{$\Delta$depth} & Conv2D + ReLU               & 3      & 1 & 64/1            & conv1                \\
        \hline
        \multicolumn{6}{c}{Convex Upsampling}                                                                   \\
        \hline
        conv0        & Conv2D + ReLU                 & 3 & 1 & 128+256/128+256             & \inp{hidden$_s$}, \inp{context} \\
        \out{upmask} & Conv2D                        & 1 & 1 & 128+256/8$\times$8$\times$9 & conv0                           \\
        \hline
    \end{tabular}
    \end{tabular}
    }
    \vspace{0.1em}
    \caption{\textbf{Framework Modules Description.} We detail each learned component of our framework. Each module inputs and outputs are shown in red and blue, respectively.}
    \label{tab:framework-overview}
\end{table}

\textbf{Training Details.} We train our model on Blended, TartanAir and DTU with AdamW, learning rate $10^{-4}$ and weight decay $10^{-5}$. We always clip gradients with global norm 1 to stabilize the behavior of Gated Recurrent Units. On Blended, we normalize relative pose translation (between the reference and source views) to have a mean value of 1 for better numerical stability. On Blended, we train for 200K steps and then fine-tune for 100K steps with a learning rate of $10^{-5}$. On DTU, we fine-tune the 200K Blended checkpoint for 100K steps with a learning rate of $10^{-4}$. We always train with batch size 1 on 2 RTX 3090 in mixed precision. During training and evaluation, we always perform 10 cycles over the input source views, that is 40 total steps with 4 source views, except for the UnrealStereo4K stereo benchmark where we perform 40 updating steps on the unique source view available. In all the experiments, we compute dynamic offsets in a neighborhood of size $||\mathcal{N}|| = 9 \times 9$ for a total of 81 sampling coordinates. 

{\small
\bibliographystyle{ieeenat_fullname}
\bibliography{egbib}
}

%% file: preamble.tex
\usepackage[dvipsnames]{xcolor}

\DeclareCaptionFormat{teaser}{%
    \textbf{#1#2}#3
}

\newcommand{\bt}[1]{\textbf{#1}}